\pdfoutput=1
\documentclass{llncs}
\usepackage{amsmath} %<--pacchetto American Mathematician Society
\usepackage{amsfonts} 
\usepackage{amssymb}
\usepackage{color}
\usepackage{graphicx} 
\usepackage{booktabs}
\usepackage{multirow}
\DeclareMathOperator\erf{erf}
\begin{document}
\title{NODDI-SH: a computational efficient NODDI extension for fODF estimation in diffusion MRI}
\author{Mauro Zucchelli\inst{1} \and Maxime Descoteaux\inst{2} \and Gloria Menegaz\inst{1}}
\institute{Department  of Computer Science, University of Verona, Italy \and Sherbrooke Connectivity Imaging Lab, Computer Science, Universit\'e de Sherbrooke, Canada}
\maketitle
\setcounter{footnote}{0}

\begin{abstract}
Diffusion Magnetic Resonance Imaging (DMRI) is the only non-invasive imaging technique which is able to
detect the principal directions of water diffusion as well as neurites density in the human brain. Exploiting the ability of
Spherical Harmonics (SH) to model spherical functions, we propose a
new reconstruction model for DMRI data which is able to estimate
both the fiber Orientation Distribution Function (fODF) and the
relative volume fractions of the neurites in each voxel, which is
robust to multiple fiber crossings.  We consider a Neurite
Orientation Dispersion and Density Imaging (NODDI) inspired single fiber
diffusion signal to be derived from three compartments: intracellular,
extracellular, and cerebrospinal fluid. The model, called NODDI-SH, is derived by convolving the single fiber response
with the fODF in each voxel. NODDI-SH embeds the calculation of the fODF
and the neurite density in a unified mathematical model providing
efficient, robust and accurate results. 
Results were validated on simulated data and tested on
\textit{in-vivo} data of human brain, and compared to  and Constrained
Spherical Deconvolution (CSD) for benchmarking. Results revealed
competitive performance in all respects and inherent adaptivity to
local microstructure, while sensibly reducing the computational
cost. We also investigated NODDI-SH performance when only a limited
number of samples are available for the fitting, demonstrating that 60
samples are enough to obtain reliable results. The fast computational
time and the low number of signal samples required, make NODDI-SH
feasible for clinical application. 
%Moreover, the use of a single model
%to obtain both microstructural indexes and fODF can greatly benefit
%the processing pipeline for DMRI data. 
\end{abstract}

% \linenumbers

%% main text
\section{Introduction}
\label{sec:int}
The diffusion process of water molecules in brain tissues reflects the cytoarchitecture of the observed tissue. Neurites density and axon fibers topology are two of the main factors which influence the diffusion process in the white matter. With Diffusion Magnetic Resonance Imaging (DMRI) it is possible to indirectly observe the Ensemble Average Propagator (EAP) of the water molecules movements and reconstruct the white matter fiber Orientation Distribution Function (fODF) \cite{csd} as well as numerical indexes reflecting the neurites density in each voxel \cite{Alexander2012}. The proposed approach provides a novel method for the joint estimation of the fiber orientation and neurite densities also in presence of complex topologies (crossing, fanning) while keeping low computational complexity which ensures fast processing. In addition, robustness to the reduction in the number of samples enables clinical exploitability providing a powerful computational tool for microstructure characterization. 
%Numerous models have been proposed in the literature to obtain the
%fODF from the diffusion signal. 

So far, different models have been proposed to solve the two problems of fODF and volume fraction estimation individually. 
The most clinically used model is the Diffusion Tensor Imaging (DTI) \cite{Basser1994}, \cite{pierpaoli} 
which considered the EAP as a single multivariate Gaussian function. DTI enables the identification of only one single principal direction of diffusion and thus is not suitable to model complex WM fiber configurations such as fiber crossings,  kissing, and fanning \cite{desbook}.

Spherical Deconvolution (SD) methods \cite{sd}, \cite{acqua2007}, \cite{dell2010} consider the diffusion signal as the result of the convolution of the fODF times a single fiber response. By deconvolving the diffusion signal with a given response function (usually obtained from single fiber areas of the brain) it is possible to estimate the fODF. However, the deconvolution process is an ill-conditioned problem which can be affected by noise or error in the estimation of the response function. Constrained Spherical Deconvolution (CSD) \cite{csd} added non-negative constraints to the fODF estimation, increasing the resolution of the fODF which is extremely important for tractography application \cite{destmi}.

The EAP represents only an indirect measure of the different components of the tissue. Several models have been proposed to explicitly quantify the contribution of the different compartments present in each voxel in order to provide a map of the different tissues. A noteworthy example of multi-compartment model is the Composite Hindered and Restricted Model (CHARMED) \cite{Assaf2005}, which considers two different types of substrate in which the water molecules can diffuse:  intracellular and extracellular. The intracellular compartment is the one in which the water diffusion is restricted by the cell walls, e.g. axons and cell bodies, and it is modeled as an axially symmetric cylinder. The extracellular compartment captures the contributions of all the water molecules trapped in-between different neurites and in which the diffusion is only hindered, and it is modeled as a multivariate Gaussian. 
Two important modifications were added to the CHARMED by \cite{Alex2010} and \cite{Zhang2011}. \cite{Alex2010} extended CHARMED with the addition of a third compartment in order to fit the cerebrospinal fluid (CSF), while \cite{Zhang2011} replaced the coherently oriented cylinders of CHARMED with a Watson distribution of cylinders in order to model fiber dispersion. Such distribution is limited by the fact that it presents only one principal direction of diffusion and cannot be used to model crossing fibers. 
Histological evidence shows that the human axonal diameter is normally less than two micrometers \cite{aboitiz}, \cite{Axon_diameter}. Cylinders presenting such diameters exhibit almost no signal decay using gradient strength of $G=100$mT/m in Pulse Gradient Spin Echo (PGSE) acquisitions with a single diffusion time \cite{nilsson2010}, \cite{nilsson2013}. Therefore, in the Neurite Orientation Dispersion and Density Imaging (NODDI) \cite{Alexander2012} model the cylinders of \cite{Zhang2011} were replaced with a Watson distribution of ``sticks'' (cylinder of zero radius, equivalent to a Gaussian function with zero perpendicular diffusion). NODDI was able to provide good results on simulated data for retrieving the volume fractions of the different compartments. %note on computational complexity?
Improvements in computational efficiency were reached by the reformulation within a convex optimization framework proposed by 
\cite{daducci}, called Accelerated Microstructure Imaging via Convex Optimization (AMICO). Moving from the original non-linear optimization to a linear optimization greatly reduced the computational time of both techniques, without downgrading the performance of the models.  
The Watson distribution suffers from the fact that it can only model isotropically dispersed distributions, and not anisotropic fiber fanning and crossings. In order to overcome this limitation, \cite{noddibingham} replaced the Watson distributions with a Bingham distribution, allowing to model fiber fanning, but not crossings. A mixture of Bingham distributions was previously adopted also in \cite{Bingham1} and  \cite{Bingham2}, but the problem of finding the number of Bingham distributions to use in each voxel, and the hardness of the fitting make these techniques less exploitable for practical cases.
A general framework for calculating any kind of probability density function of fibers distribution on the sphere, given the model of a single fiber response, was proposed
in  Fiber ORientation Estimated using Continuous Axially Symmetric Tensors (FORECAST), \cite{forecast}. FORECAST exploits SH to model any distribution of axially symmetric tensors on the sphere. To our knowledge, FORECAST is also one of the first models exploiting the mean value of the signal to estimate rotation invariant fiber specific microstructural parameters (in this case parallel and perpendicular diffusivities). Jespersen et al. (2007) \cite{sune2007} advanced the FORECAST model considering instead of a single axially symmetric tensor, a two-compartmental model composed of an axially symmetric tensor and an isotropic tensor for modeling the single fiber response. Neurite density estimation obtained using this model was compared with light microscopy and electron microscopy images in \cite{sune2010}. FORECAST parameters estimation which uses the mean of the signal on each b-value for estimating microstructural parameters has been recently called Spherical Mean Technique (SMT), by \cite{kaden2} and \cite{kaden}.
  In particular in \cite{kaden} employ the same strategy as FORECAST in order to estimate the parallel and perpendicular diffusivity in each voxel. \cite{reisert} extended the SMT to other rotational invariant features, combining them with machine learning to estimate several microstructural parameters from the diffusion signal. 

In this work, we expanded the FORECAST single tensor model to a
NODDI-like multi-compartment model, gathering the advantages of both
techniques in a unified model called NODDI-SH. NODDI-SH is able to obtain accurate microstructural
information, such as the intracellular, extracellular, and CSF volume
fractions, in voxels presenting any kind of fiber distribution, from
crossing to fanning. Unlike the classical NODDI, NODDI-SH
permits to reconstruct at the same time also the fODF in each voxel,
making it suitable for tractography and connectomic studies. Thanks to the SMT,
NODDI-SH elegant mathematical framework enables an extremely fast
computation of the volume fractions and fast fODF reconstruction. 
NODDI-SH results were compared to those provided by NODDI and CSD,
respectively, for the volume fractions and the fODF, demonstrating
competitive results in both respects. We also tested NODDI-SH
performance when only a limited number of samples are provided for
the fitting. In addition, robustness to the reduction in the number of
samples was determined in order to establish the clinical
exploitability of the proposed model. 
%Simulated data were used in order to compare the results of NODDI-SH with respect to NODDI for what concerns the volume fractions in the case of isotropic fanning, anisotropic fanning, and crossing of fibers. We also compare the precision of NODDI-SH fODF estimation with respect to the popular CSD algorithm. 

The manuscript is organized as follows. Section \ref{sec:met} revisits
the FORECAST model and introduces the NODDI-SH model, as well as the
simulated and \textit{in-vivo} data used for the experiments. Section \ref{sec:res} shows the comparison of NODDI-SH with NODDI in the simulated and \textit{in-vivo} experiments, as
well as the robustness of the NODDI-SH with respect to the number of
samples. Section \ref{sec:dis} discusses our findings, and Section
\ref{sec:con} summarizes the conclusions of our work.

\section{Methods}
\label{sec:met}
Assuming that in the considered voxel the fiber population is homogeneous, the diffusion signal $E(\textbf{q})$ can be modeled as
\begin{equation}
\label{eq:eq_initial}
    E(q,{\textbf{u}}) = \int_{{\textbf{v}} \in \mathcal{S}^2} \rho({\textbf{v}}) F(q, {\textbf{u}}, {\textbf{v}}) d {\textbf{v}}
\end{equation}
where $F$ is the diffusion signal of a fiber aligned along ${\textbf{v}}$ and $\rho({{\textbf{v}}})$ is the fODF.
Since $\rho$ is probability density function it must satisfy the following conditions
\begin{eqnarray}
\label{eq:property}
    & \int_{{\textbf{v}}} \rho({\textbf{v}}) \ d {\textbf{v}}  = 1\\
    \label{eq:property2}
    & \rho({\textbf{v}}) \ge 0\  \forall \ {\textbf{v}} \in \mathcal{S}^2 
\end{eqnarray}
In diffusion MRI, the fODF is generally modeled using Spherical Harmonics (see Appendix \ref{ap_sh}) \cite{csd}, \cite{forecast}, \cite{kaden}. 
Replacing the SH equation into equation \eqref{eq:eq_initial} we obtain
\begin{equation}
\begin{split}
\label{eq:general}
    E(q,{\textbf{u}}) =& \int_{{\textbf{v}} \in \mathcal{S}^2}  \sum_{l=0, even}^{\infty} \sum_{m=-l}^{l} c_{lm} Y_l^m({\textbf{v}}) F(q, {\textbf{u}} , {\textbf{v}} ) d {\textbf{v}}\\
    =& \sum_{l=0, even}^{\infty} \sum_{m=-l}^{l} c_{lm} \int_{{\textbf{v}} \in \mathcal{S}^2}   Y_l^m({\textbf{v}}) F(q, {\textbf{u}} , {\textbf{v}} ) d {\textbf{v}}\\
\end{split}
\end{equation}
The choice of the single fiber signal $F$ will define the resulting signal basis. In this section we will consider two cases. In the first case $F$ is modeled as an axially symmetric Gaussian function (FORECAST), while in the second it results from the linear combination of the contribution arising from a three compartments model (NODDI-SH).
\subsection{FORECAST model}
FORECAST considers $F$ to be an axially symmetric tensor, with parallel diffusivity $\lambda_{\parallel}$ and perpendicular diffusivity $\lambda_{\perp}$
\begin{equation}
\label{eq:tensor}
 F(b, {\textbf{u}} , {\textbf{v}}) = \exp\bigl(-b {\textbf{u}}^T \textbf{D}({\textbf{v}}) {\textbf{u}}\bigr)
\end{equation}
where $b = 4\pi^2 \tau q^2$ is the \textit{b}-value at the given diffusion time $\tau$,
and  $\textbf{D}({\textbf{v}})$ is the diffusion tensor aligned along ${\textbf{v}}$
\begin{equation}
 \textbf{D}({\textbf{v}}) = (\lambda_{\parallel}- \lambda_{\perp}) ({\textbf{v}} {\textbf{v}}^T) + \lambda_{\perp}\textbf{I}
\end{equation}
Replacing equation \eqref{eq:tensor} in equation \eqref{eq:general}, we can calculate $E(\textbf{q})$ as 
\begin{equation}
\label{eq:forecast_base}
    E(b,{\textbf{u}}) = \sum_{l=0, even}^{\infty} \sum_{m=-l}^{l} c_{lm} \int_{{\textbf{v}} \in \mathcal{S}^2}    \exp\bigl(-b {\textbf{u}}^T \textbf{D}({\textbf{v}}) {\textbf{u}}\bigr)  Y_l^m({\textbf{v}}) d {\textbf{v}}
\end{equation}
Using the \textit{Funk-Hecke Theorem} \cite{Descoteaux2007} it is possible to solve the integral in equation \eqref{eq:forecast_base} (see Appendix \ref{sec:appendix_a} for the complete derivation) obtaining the final FORECAST model
\begin{equation}
\label{eq:signal_forecast}
\begin{split}
    E(b,{\textbf{u}}) =& \sum_{l=0, even}^{N} \sum_{m=-l}^{l} c_{lm} 2 \pi \exp\bigl(-b \lambda_{\perp} \bigr)    \Psi_{l}(b(\lambda_{\parallel}- \lambda_{\perp})) Y_l^m({\textbf{u}}) \\    
\end{split}
\end{equation}
where $\Psi_{l}(\xi)$ is the solution of $\int_{-1}^{1} P_l(\xi) \exp\bigl(-\xi t^2  \bigr)  d t$, with $P_l$ the Legendre polynomial of order $l$. 

The estimation of the fODF coefficients $c_{lm}$ requires the estimation of $\lambda_{\parallel}$ and $\lambda_{\perp}$ in each voxel. In order to estimate these diffusivities, \cite{forecast} proposed to exploit the average of the diffusion signal for a given \textit{b}-value, which is given by the $l=0, m=0$ element of the FORECAST basis
\begin{equation}
\label{eq:mean_forecast}
\begin{split}
    \overline{E}(b)& =   \dfrac{1}{2} \exp\bigl(-b \lambda_{\perp} \bigr)    \Psi_{0}(b (\lambda_{\parallel}- \lambda_{\perp}))\\     
\end{split}
\end{equation}
Equation \eqref{eq:mean_forecast} exploits the fact that the coefficient $c_{00}$ is equal to $\frac{1}{\sqrt{4\pi}}$, as detailed in Appendix \ref{sec:special}, in order to 
fulfill the requirements of equation \eqref{eq:property}. If the signal is acquired in more than one shell (e.g. using more than one \textit{b}-value), it is possible to use equation \eqref{eq:mean_forecast} to estimate $\lambda_{\parallel}$ and $\lambda_{\perp}$ accurately.

\subsection{The NODDI-SH model}
\label{sec:shadow}
The NODDI-SH model considers the single fiber diffusion signal $F$ as the weighted sum of three tissue compartments: intracellular $F_{ic}$, extracellular $F_{ec}$, and cerebrospinal fluid $F_{csf}$
\begin{equation}
\label{eq:shadow_single}
 F(\textbf{b}) = \nu_{ic}F_{ic}(\textbf{b}) + \nu_{ec}F_{ec}(\textbf{b})+ \nu_{csf}F_{csf}(\textbf{b}) 
\end{equation}
where $\nu_{ic}$, $\nu_{ec}$, and $\nu_{csf}$ are the relative volume fractions in the tissue, with the constraint $\nu_{ic}+\nu_{ec}+\nu_{csf}=1$.
The compartment $F_{ic}$ is modeled as a tensor with $\lambda_\perp=0$, as in \cite{Alexander2012}, the extracellular compartment $F_{ec}$ as a tensor with a certain parallel diffusivity $\lambda_{\parallel}$ and $\lambda_\perp=\lambda_\parallel \frac{\nu_{ec}}{\nu_{ec} + \nu_{ic}}$ \cite{Alex2010}, and the CSF compartment $F_{csf}$ as an isotropic tensor with $\lambda_{csf}=3\cdot 10^{-3}$ mm$^2$/s
\begin{equation}
\begin{split}
& F_{ic}(b{\textbf{u}}) =\exp\bigl(-b \lambda_{\parallel}({\textbf{u}}^T {\textbf{v}})^2 \bigr)  \\
& F_{ec}(b{\textbf{u}}) = \exp\bigl(-b \bigl[ (\lambda_{\parallel}- \lambda_{\perp})({\textbf{u}}^T {\textbf{v}})^2 + \lambda_{\perp} \bigr] \bigr)  \\ 
& F_{csf}(b)  =\exp\bigl(-b\ \lambda_{csf} \bigr)  \\ 
\end{split}
\end{equation}

Replacing equation \eqref{eq:shadow_single} into equation \eqref{eq:general} (see Appendix \ref{sec:appendix_b}) leads to the final NODDI-SH model formulation
\begin{equation}
\begin{split}
    E(b,{\textbf{u}}) = &  c_{00}\sqrt{4 \pi} \nu_{csf}\exp\bigl(-b \lambda_{csf} \bigr)  +\sum_{l=0, even}^{N} \sum_{m=-l}^{l} c_{lm} 2 \pi \times \\ & \times \Bigl[\nu_{ic}  \Psi_{l}(b \lambda_{\parallel})  
+ \nu_{ec} \exp\bigl(-b \lambda_{\perp} \bigr) \Psi_{l}(b (\lambda_{\parallel}- \lambda_{\perp})) \Bigr] Y_l^m({\textbf{u}}) \\     
\end{split}
\end{equation}
in which the cerebrospinal fluid compartment volume fraction $\nu_{csf}$, influences only the first coefficient of the basis (the one referring to the isotropic SH), while intra and extra cellular compartments affect all the coefficients.

Using the SMT \cite{forecast}, \cite{kaden}, \cite{kaden2}, the first element of the basis at $l=0$ and $m=0$ represents the mean value of the diffusion signal at a given \textit{b}-value $\overline{E}(q)$
\begin{equation}
\label{eq:mean}
\begin{split}
    \overline{E}(b) =&    \nu_{csf}\exp\bigl(-b \lambda_{csf} \bigr)  +  \frac{1}{2}\Bigl[\nu_{ic}  \Psi_{0}(b \lambda_{\parallel})  
+ \\
& + \nu_{ec} \exp\bigl(-b \lambda_{\perp} \bigr) \Psi_{0}(b (\lambda_{\parallel}- \lambda_{\perp})) \Bigr]  \\     
\end{split}
\end{equation}
The estimation of the volume fractions is performed as follows: we initially create a dictionary of 383 combinations of possible volume fractions. The dictionary was created by sampling in a non-uniform manner the volume fractions space. In particular, we considered the number of combinations of $\nu_{ic}$ and $\nu_{ec}$ to be proportional to $1-\nu_{csf}$. Using this method the dictionary presents a higher variability of intracellular and extracellular volume fractions where the CSF volume fraction is low, and a lower variability where the CSF volume fraction is high.  
For each combinations of these volume fractions we calculated the estimated $\overline{E}$ using equation \eqref{eq:mean}, for each \textit{b}-value. 
In order to select the best combination, we evaluated the mean of the diffusion signal at each \textit{b}-value as $\frac{1}{N}\sum_{i=0}^N E(\vert \textbf{b}_i \vert)$ with N the number of samples on each shell and thus selected the set of volume fractions which better approximates the average diffusion signal. 
The evaluation of equation \eqref{eq:mean} is extremely fast, and even if we test all the 383 combinations of volume fractions, the total estimation time is of the order of fraction of second per voxel (see Section \ref{sec:times} for computational time details). 

\subsection{Basis coefficients estimation}
The estimation of the basis coefficients for both FORECAST and NODDI-SH is performed via convex optimization using CVXOPT\footnote{\url{http://cvxopt.org}} allowing to enforce the positivity constraint as in equation \eqref{eq:property2} on a spherical grid composed of 181 points.
In detail, our objective is to find
\begin{equation}
    \min_{\textbf{c}} \Vert \hat{E} - \textbf{M} \textbf{c} \Vert_2^2
\end{equation}
subject to
\begin{eqnarray}
    \textbf{Y} \textbf{c} &\geq 0 \\
    c_{00} &= \frac{1}{\sqrt{4 \pi}}
\end{eqnarray}
where $\hat{E}$ is the measured diffusion signal, $\textbf{c}$  is the coefficients vector, $\textbf{M}$ the NODDI-SH or FORECAST basis matrix, and $\textbf{Y}$ is the SH basis matrix.
We implemented both algorithms under the Diffusion Imaging in Python (DIPY) \cite{dipy1}, \cite{dipy2} software library\footnote{\url{http://dipy.org}} and the code is available on request. The CSD response function was fixed for all the brain and modeled using the average eigenvalues of the diffusion tensor in voxels presenting fractional anisotropy higher than 0.8. We adopted the same strategy to estimate NODDI-SH $\lambda_\parallel$ parameter. In simulated data we set $\lambda_\parallel=1.7\times10^{-03}$ mm$^2$/s and $\lambda_\perp= 0.1\times10^{-03}$ mm$^2$/s.  SH basis of order 8 was chosen for both techniques.

NODDI results were calculated using the standard NODDI toolbox\footnote{\url{http://mig.cs.ucl.ac.uk/index.php?n=Tutorial.NODDImatlab}}. 
In NODDI model \cite{Alexander2012} intracellular volume fraction is weighted by a factor $(1-\nu_{csf})$ in order to be compared to NODDI-SH $\nu_{ic}$ \cite{MIA}. In our simulations this normalization led to more accurate  $\nu_{ic}$ estimation for the NODDI model.

\subsection{Simulated dataset}
To generate a model of white matter, we simulated the diffusion signal as the weighted sum of an intracellular and an extracellular contribution. The intracellular part was modeled as a cylinder with diameter $1 \mu m$ using the Multiple Correlation Function framework \cite{Grebenkov}, \cite{mcf}, \cite {MIA}, while the extracellular part was modeled as a symmetric Gaussian. Both the cylinder and the Gaussian share the same $\lambda_\parallel=1.7\times10^{-3}$mm$^2$/s, while the extracellular $\lambda_\perp$ depends on the extracellular volume fraction as detailed in Section \ref{sec:shadow}.

\begin{figure}[!t]
\centering{
\begin{tabular}{ccc}
$\kappa=128$, $\beta=0$ & $\kappa=32$, $\beta=0$  & $\kappa=4$, $\beta=0$    \\
\includegraphics[width=0.25\columnwidth]{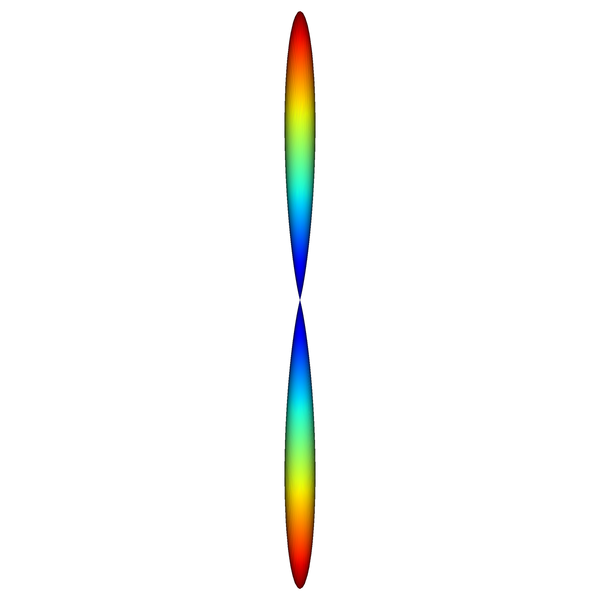} &
\includegraphics[width=0.25\columnwidth]{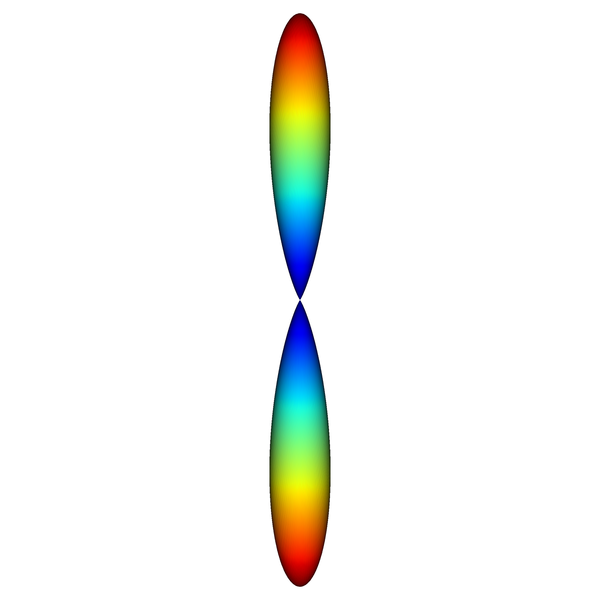} &
\includegraphics[width=0.25\columnwidth] {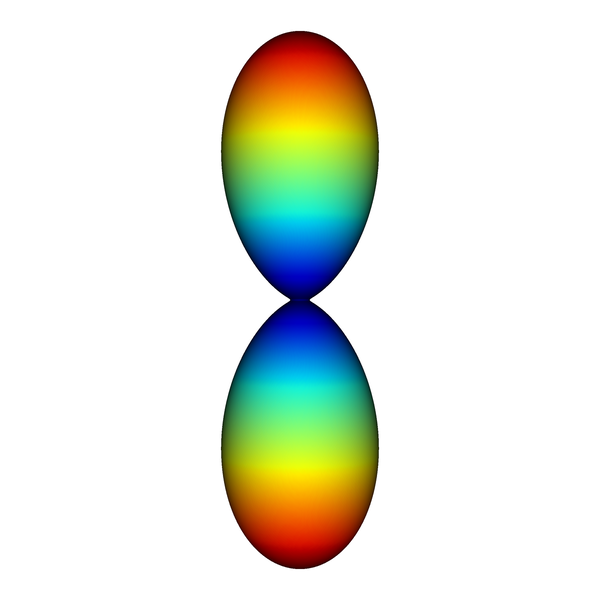}\\
$\kappa=128$, $\beta=64$  & $\kappa=32$, $\beta=16$ & $\kappa=4$, $\beta=2$    \\
\includegraphics[width=0.25\columnwidth]{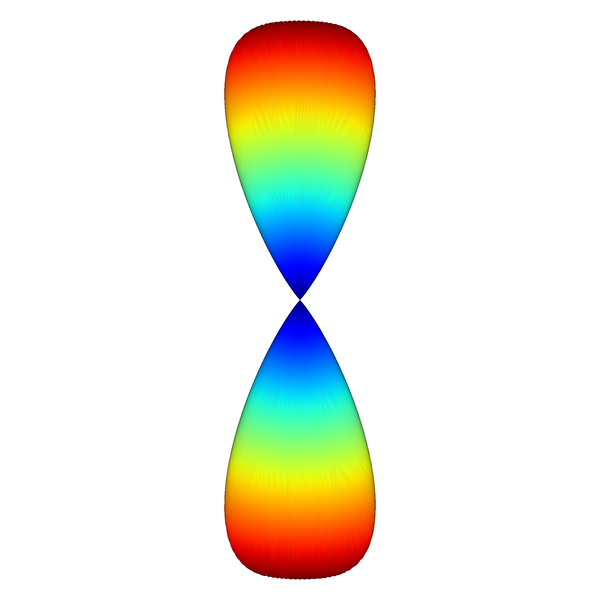} &
\includegraphics[width=0.25\columnwidth]{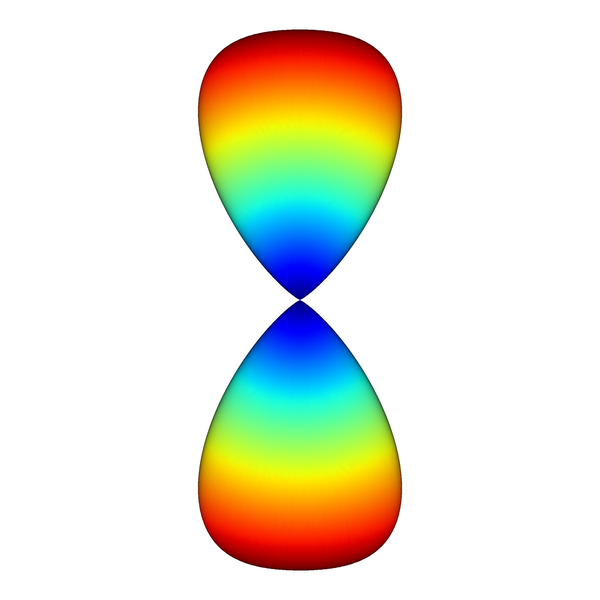} &
\includegraphics[width=0.25\columnwidth] {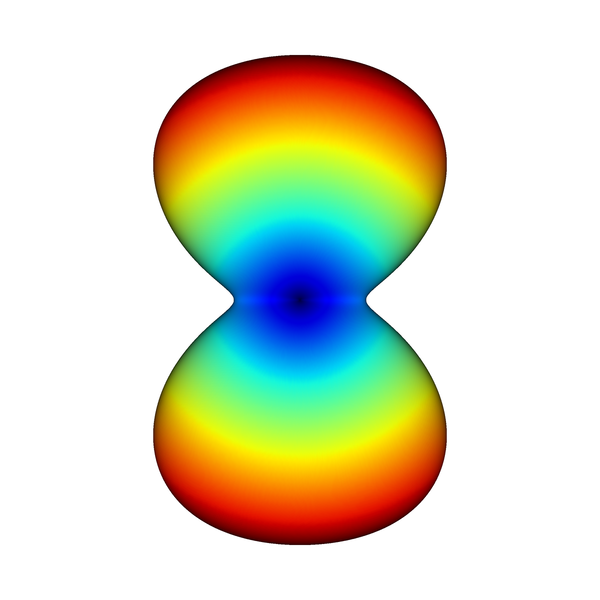}\\
  \end{tabular} }
\caption{Instances of Kent distribution given the parameters $\kappa$ and $\beta$.}
\label{fig:rho}
\end{figure}

To model different cases of dispersion and crossing, we initially extract $M$ random directions from the Kent distribution. Kent distribution \cite{Kent}, \cite{MIA} is able to model symmetric probability density functions on the sphere. In particular Kent distribution $\rho$ is modeled as 
\begin{equation}
 \rho(\textbf{u}, \beta, \kappa, \boldsymbol{\mu}) = \frac{1}{\textrm{c}(\kappa,\beta)}\exp\{\kappa\boldsymbol{\mu} \cdot\mathbf{u}+\beta[(\boldsymbol{\gamma}_{1}\cdot\mathbf{u})^2-(\boldsymbol{\gamma}_2\cdot\mathbf{u})^2]\} 
\end{equation}
where $\kappa$ is the concentration parameter which rules the degree of orientation dispersion and $\beta$ is the parameter which models the anisotropy of such a dispersion. The parameter $\beta$ is bounded between 0 and $\frac{\kappa}{2}$, with $\beta=0$ meaning perfectly isotropic dispersion and $\beta=\frac{\kappa}{2}$ completely anisotropic dispersion, respectively.
The variable $\boldsymbol{\mu}$ represents the central direction of the pdf, while the two orthogonal unit vectors $\boldsymbol{\gamma}_1$ and $\boldsymbol{\gamma}_2$ define the orientation of the anisotropic dispersion.
The function $c(\kappa, \beta)$ is defined as 
\begin{equation}
c(\kappa,\beta)=2\pi\sum_{j=0}^\infty\frac{\Gamma(j+\frac{1}{2})}{\Gamma(j+1)}\beta^{2j}\left(\frac{1}{2}\kappa\right)^{-2j-\frac{1}{2}} I_{2j+\frac{1}{2}}(\kappa)
\end{equation}
with $\Gamma$ the Gamma function and $I$ the modified Bessel function.
Figure \ref{fig:rho} shows some examples of the Kent distribution\footnote{Kent distribution has only one lobe. We added the antipodal symmetric lobe for visualization purposes.} $\rho$, given the parameters $\kappa$ and $\beta$.

To model dispersion and crossing we extract $M=100$ random directions $\boldsymbol{\mu}_i$ from the Kent distribution. For each of these directions, we simulate the white matter signal as the sum of a cylinder and a Gaussian aligned in the direction $\boldsymbol{\mu}_i$. The total signal is than calculated as 
\begin{equation}
E_{total} = \frac{1}{M}\sum_{i=1}^{M} \nu_{ic} E_{ic}(\boldsymbol{\mu}_i) + \nu_{ec} E_{ec}(\boldsymbol{\mu}_i)
\end{equation} 
where the directions $\boldsymbol{\mu}_i$, the intracellular volume fraction $\nu_{ic}$ and the extracellular volume fraction $\nu_{ec}$ are given.
In our simulations we consider the following parameters: $\kappa = [128,32,4]$, $\beta=[0, \frac{\kappa}{4}, \frac{\kappa}{2}]$, three different rotations of the Kent distribution on its axis, and 11 different orientations of the Kent distribution $\boldsymbol{\mu}$ for a total of 297 combinations of parameters only for the Kent distribution. We tested also different combinations of intracellular and extracellular volume fractions, with $\nu_{ic} = [0.6, 0.65, 0.70, \dots, 1.0]$ and $\nu_{ec} = 1.0 - \nu_{ic}$. We added Rician noise with a Signal to Noise Ratio (SNR) equal to 20 (10 different instances per combination of parameters). The total number of voxels generated with such a scheme is 26730.  

\begin{figure}[!t]
\centering{
\begin{tabular}{ccc}
$90^o$ & $60^o$  & $45^o$ \\
\includegraphics[width=0.28\columnwidth]{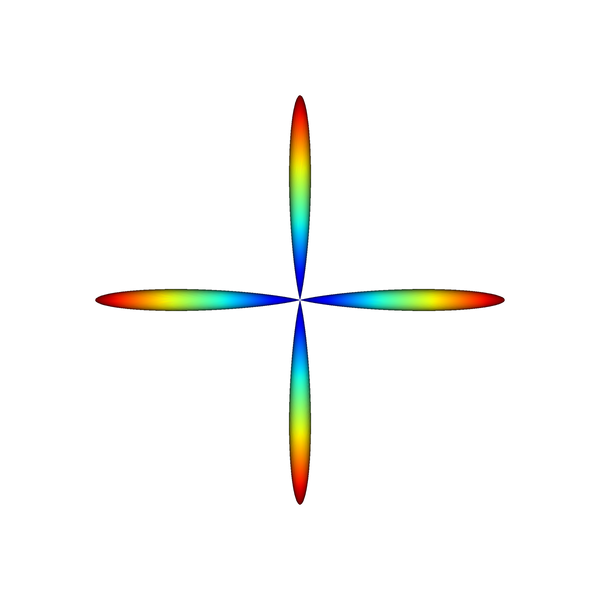} &
\includegraphics[width=0.28\columnwidth]{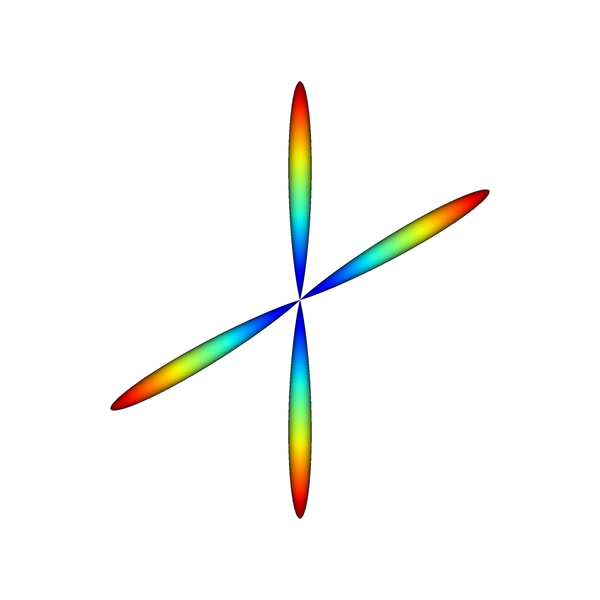} &
\includegraphics[width=0.28\columnwidth] {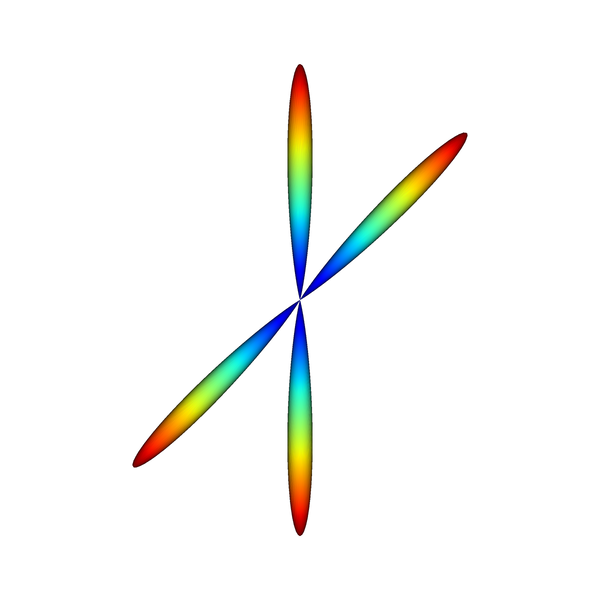}\\
  \end{tabular} }
\caption{Instances of crossings generated with two Kent distributions with $\kappa=128$ and $\beta=0$.}
\label{fig:crossing}
\end{figure}

With this model, we can simulate any kind of fiber fanning, but not fiber crossings. In order to overcome this limitation, we considered two Kent distributions, both with $\kappa=128$ and $\beta=0$ but centered on two different directions $\boldsymbol{\mu}$ and $\boldsymbol{\eta}$ (see Figure \ref{fig:crossing}).
The total signal equation in this case becomes
\begin{equation}
\begin{split}
E_{total} = & \frac{1}{M/2}\sum_{i=1}^{M/2} \nu_{ic} E_{ic}(\boldsymbol{\mu}_i) + \nu_{ec} E_{ec}(\boldsymbol{\mu}_i) + \\
& +\frac{1}{M/2}\sum_{i=1}^{M/2} \nu_{ic} E_{ic}(\boldsymbol{\eta}_i) + \nu_{ec} E_{ec}(\boldsymbol{\eta}_i)
\end{split}
\end{equation}
Crossing angles of 90, 60, and 45 degrees were considered, and, as in the previous case, the orientation of the crossing was changed in 11 different directions. The same volume fractions combination of the ``single fiber'' dataset was used for the crossing. Rician noise at $SNR=20$ was added to the signal in ten different instances. The total number of voxels, in this case, was 2970.

All the simulations were generated using the Human Connectome Project (HCP) \cite{sotiropoulos2013} sampling scheme, which will be presented in the next subsection.
%%% voxel crossings NODDI time 4515 s
%%% NODDI-SH 56s
\subsection{\textit{In-vivo} dataset}
In order to test the NODDI-SH model \textit{in-vivo} we considered one subject of the HCP diffusion MRI dataset.
HCP acquisition scheme is composed of 18 volumes sampled at \textit{b}-value 0 s/mm$^2$, plus 90 volumes at \textit{b}-value 1000,   2000, and 3000 s/mm$^2$, for a total of 288 volumes. Each of the 270 volumes which composed the three shell are acquired using an unique gradient direction. HCP pulse separation time $\Delta= 43.1$ms and pulse width $\delta= 10.6$ms corresponding to an effective diffusion time $\tau= \Delta- \delta/3 = 39.6$ms. Each volume consists of $145 \times 174 \times 145$ voxels with a resolution of 1.25mm$^3$.

HCP scheme was created using an incrementally optimal sampling \cite{caruyer}. This means that considering only the first $N_s$ samples, still provides an almost optimal angular coverage.  This property was exploited for reducing the number of q-samples as detailed in Section \ref{subsampling}. In particular, two subsets of $30$ and $60$ samples per shell were considered and the corresponding signal was compared to that obtained using the fully sampled data ($90$ samples).  In the same section, we also considered the case where only the first two shells with \textit{b}-value 1000 and 2000 s/mm$^2$ are provided to the model. The results of the NODDI-SH fitting were evaluated by calculating the mean square error (MSE) between the original HCP signal and the NODDI-SH signal for all the samples, also those which were not used for the fitting.

\section{Results}
\label{sec:res}
\subsection{Results of simulated data}
\begin{figure}[!ht]
\centering{
\setlength{\tabcolsep}{3pt}
\begin{tabular}{cc}
\hspace{0.8cm}{ISOTROPIC FANNING} & \hspace{0.8cm}{ANISTROPIC FANNING} \\
\includegraphics[width=0.45\columnwidth]{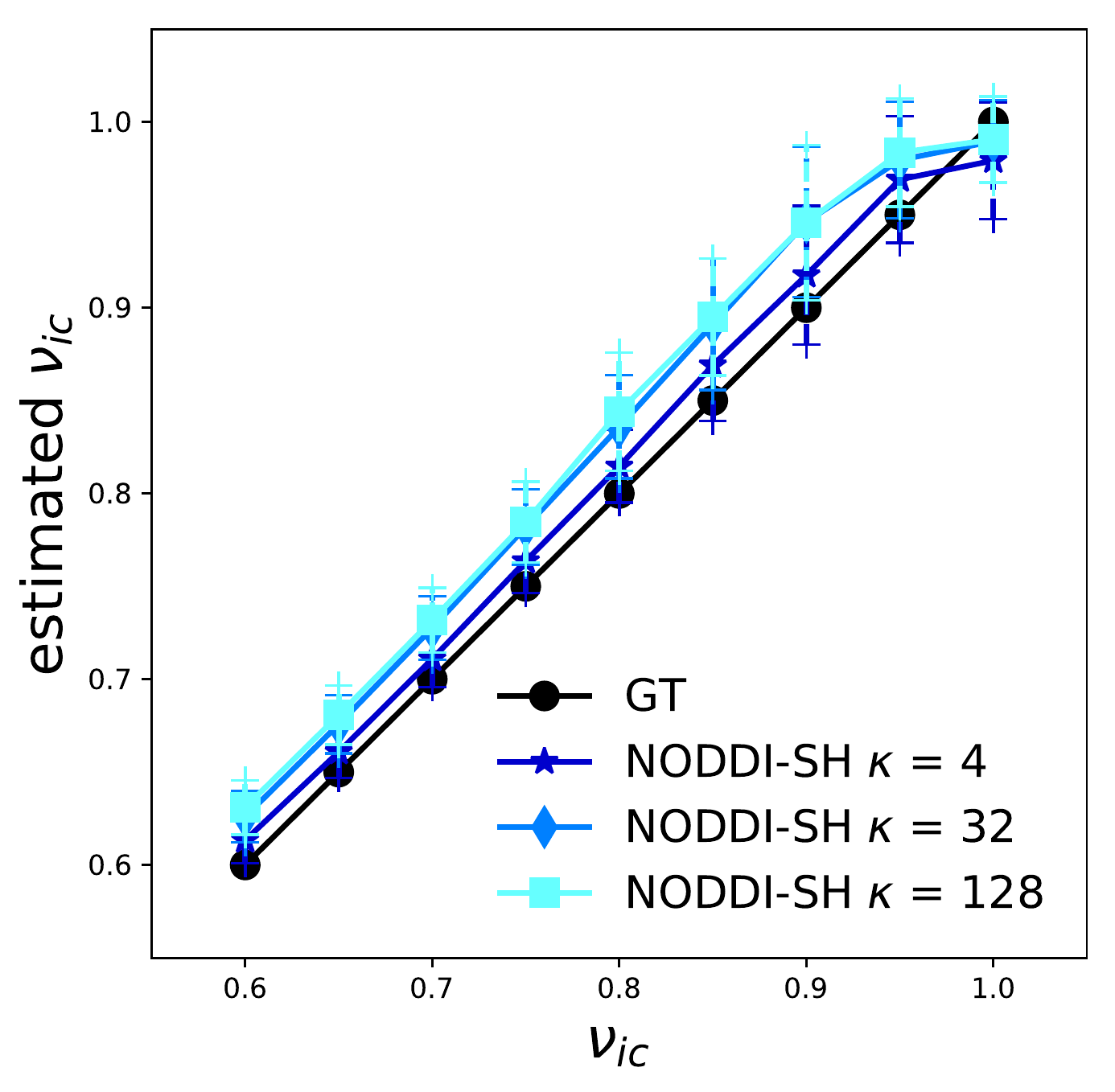} &
\includegraphics[width=0.45\columnwidth]{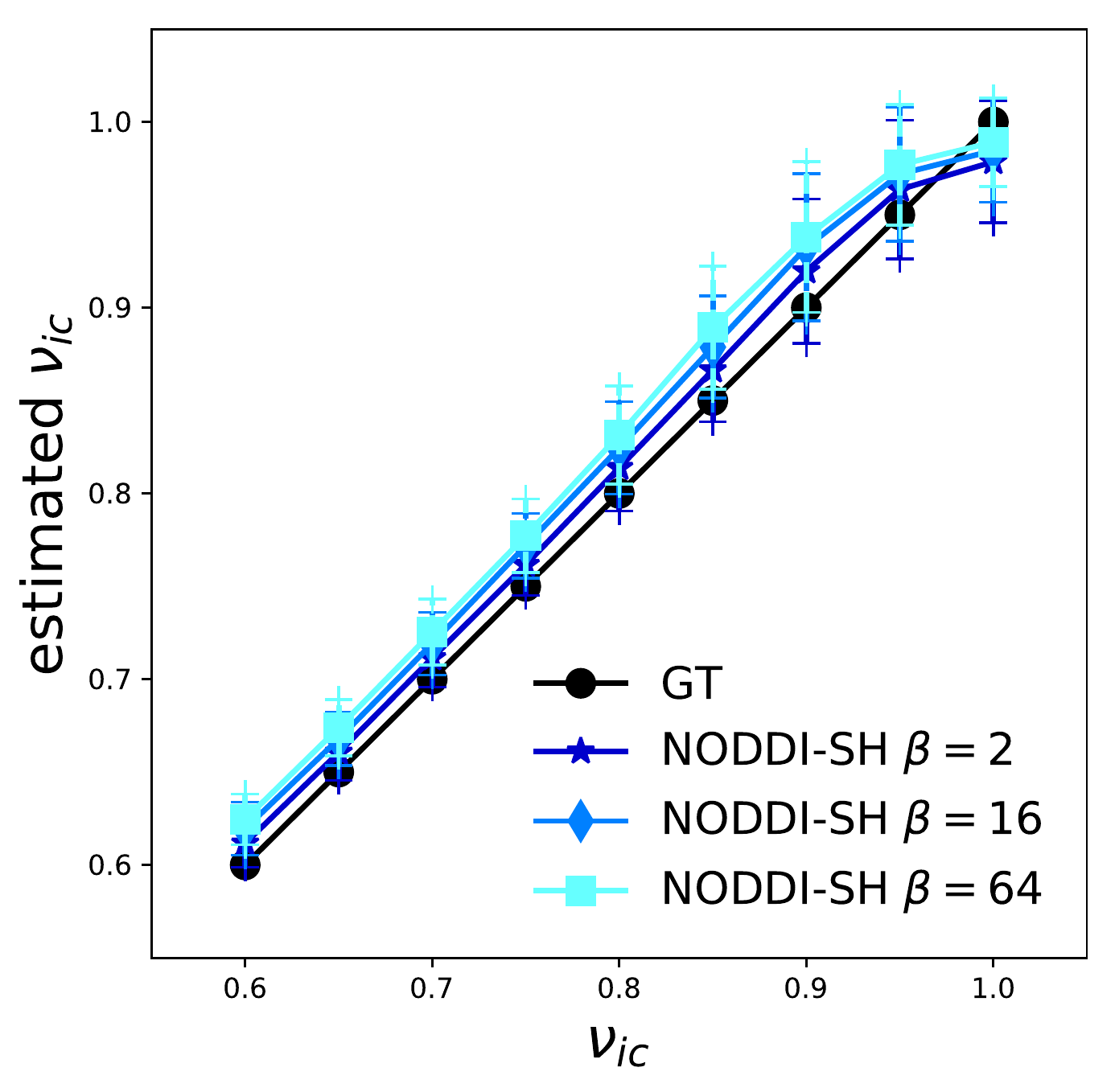} \\
\includegraphics[width=0.45\columnwidth]{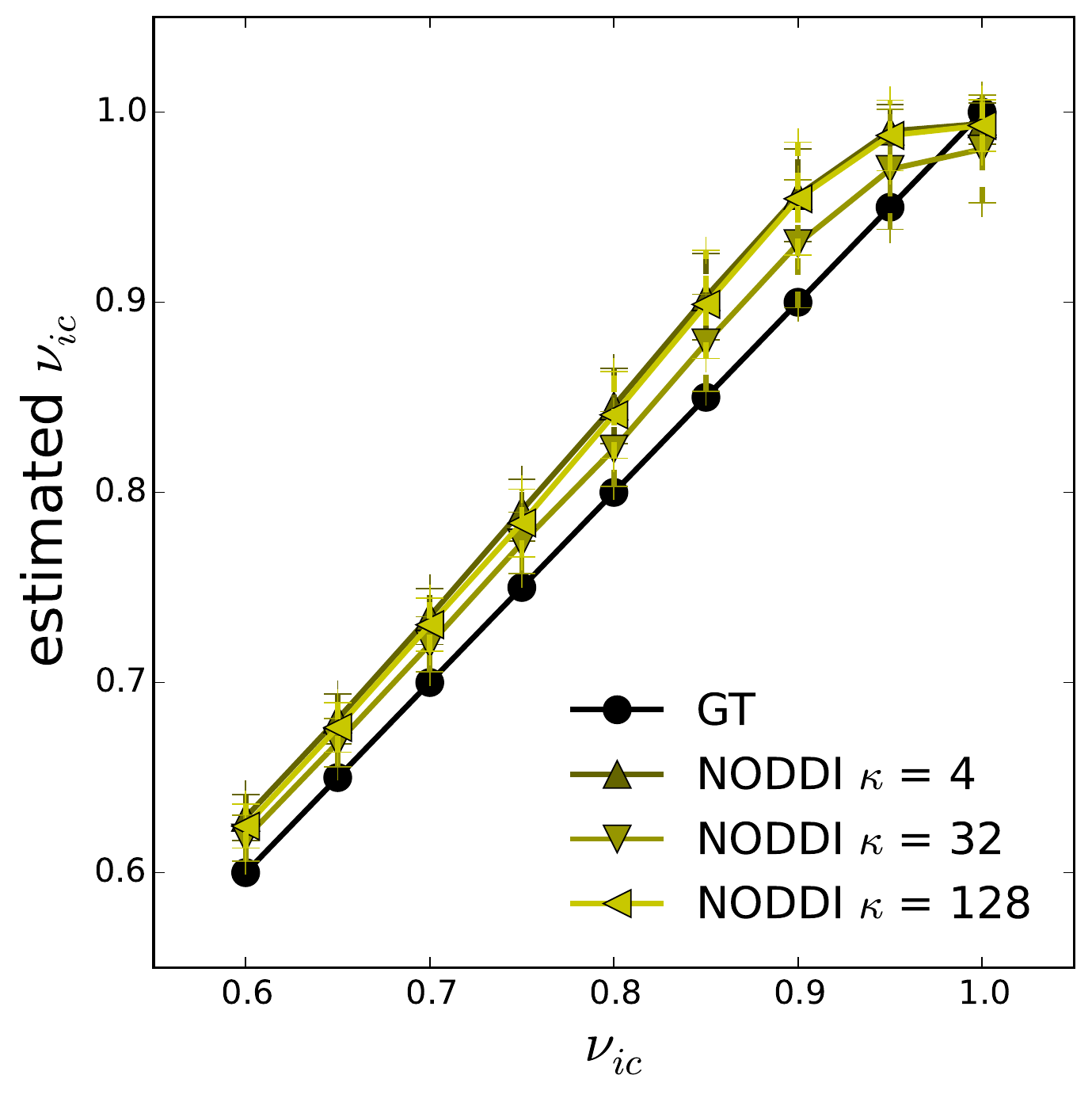} &
\includegraphics[width=0.45\columnwidth]{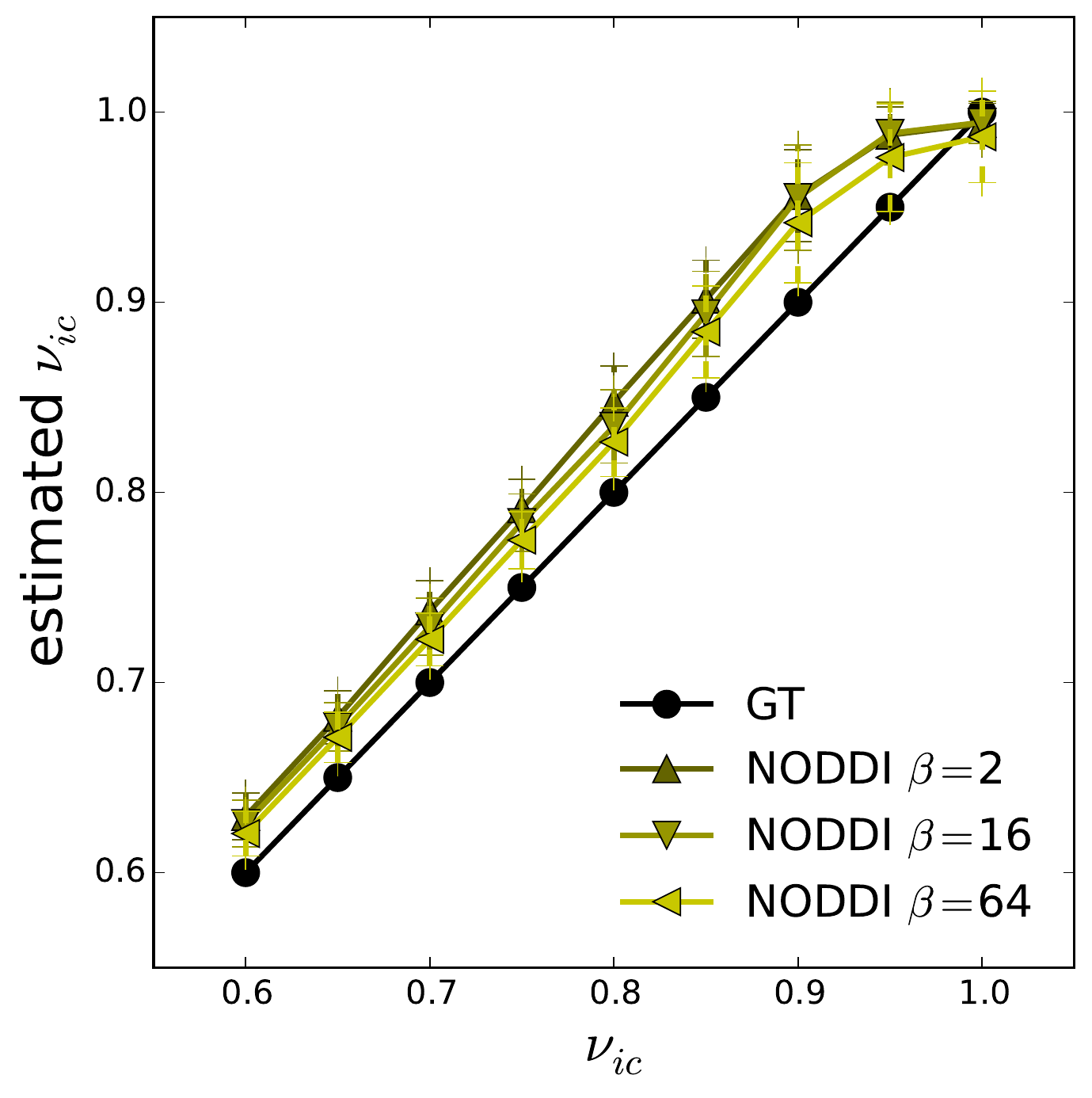} \\
  \end{tabular} }
\caption{Estimation of the intracellular volume fraction $\nu_{ic}$ of the simulated data for NODDI-SH (top row) and NODDI (bottom row) in the case of isotropic fanning ($\beta=0$) and anisotropic fanning.}
\label{fig:sim_vic}
\end{figure} 
Figure \ref{fig:sim_vic} (left column) shows the estimation of the intracellular volume fraction for NODDI-SH and NODDI in the case of isotropic dispersion ($\beta=0$), compared to the ground truth (GT) values (black circles). NODDI-SH is able to provide the most accurate estimation of $\nu_{ic}$ in case of highly dispersed fibers ($\kappa=4$), while NODDI in case of $\kappa=32$. For NODDI-SH, the estimation error increases while decreasing the dispersion ($\kappa=32$ and $\kappa=128$).
Both models tend to slightly overestimate $\nu_{ic}$ in all conditions. The average errors for NODDI-SH are 3.4$\%$, 3.1$\%$, and 1.6$\%$ for $\kappa=128$, $\kappa=32$, and $\kappa=4$ respectively. The same errors for NODDI are 3.4$\%$, 2.2$\%$, and 3.7$\%$. The standard deviation (plus sign in the graphs) of NODDI is slightly lower than NODDI-SH, with an average standard deviation of 2.1$\%$ and 2.7$\%$ for NODDI and SHADOW, respectively.
In case of anisotropic dispersion (Figure \ref{fig:sim_vic}, right column) NODDI-SH estimation is closer to the GT value with respect to the case of isotropic dispersion.
NODDI model provides better estimation of $\nu_{ic}$ for $\beta=64$, but progressively underestimates the intracellular volume fraction as the anisotropy of the dispersion decreases. NODDI-SH, on the contrary, presents an inverse pattern with its best $\nu_{ic}$ estimation for $\beta=2$ case, while the largest error is reached for $\beta=64$. The average errors for the $\beta=64$, $\beta=16$, and $\beta=2$ are respectively 2.8$\%$, 2.2$\%$, and 1.5$\%$ for NODDI-SH and 2.6$\%$, 3.3$\%$, and 3.8$\%$ for NODDI. The average standard deviations in this case are 1.8$\%$ and 2.5$\%$ for NODDI and NODDI-SH, respectively.
%{\renewcommand{\arraystretch}{1.2}
\begin{figure}[!b]
\centering{
\setlength{\tabcolsep}{3pt}    
\begin{tabular}{cc}
& \hspace{-6cm}CROSSING FIBERS \\
\includegraphics[width=0.45\columnwidth]{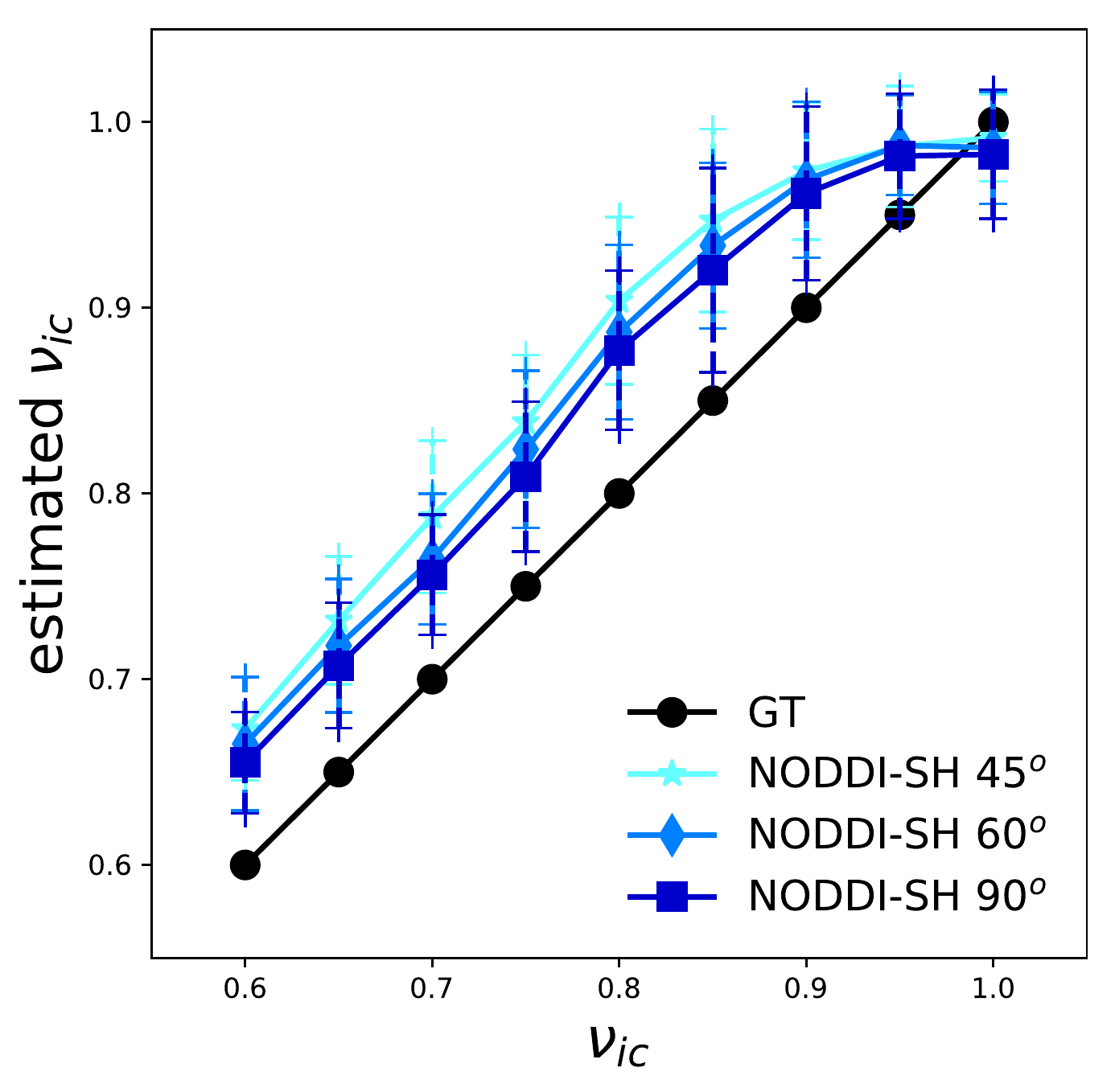} &
\includegraphics[width=0.45\columnwidth]{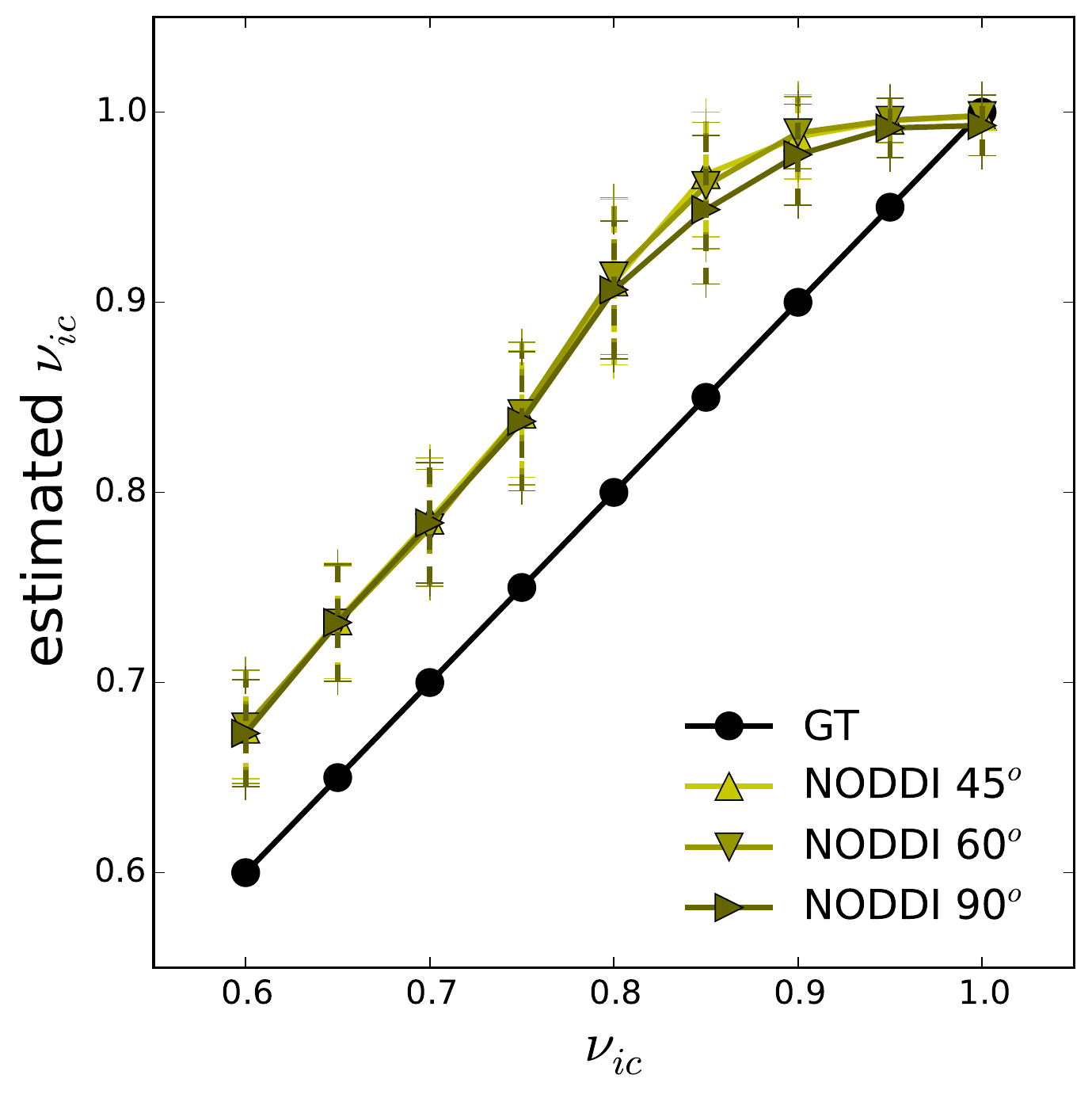} \\
  \end{tabular} }
\caption{Estimation of the intracellular volume fraction $\nu_{ic}$ of the simulated data for NODDI-SH (left) and NODDI (right) in the case of crossing fibers.}
\label{fig:sim_vic_cross}
\end{figure} 
Figure \ref{fig:sim_vic_cross} shows the estimation of $\nu_{ic}$ for the crossing fibers dataset. In this case, the estimation errors of both, NODDI-SH and NODDI, increase with respect to the fiber fanning dataset. Although estimating the intracellular volume fraction for crossing fibers represents a more difficult task with respect to the single fiber case, NODDI-SH estimation results slightly more accurate with respect to NODDI. In this dataset, average errors across crossing angles are 7.58$\%$ and 6.24$\%$ for NODDI and NODDI-SH, respectively. The amplitude of the crossing angle does not seem to have an impact on the estimation for both models. The average standard deviations in the crossing fiber case are 2.8$\%$ and 3.9$\%$ for NODDI and NODDI-SH, respectively.

\begin{figure}[!t]
\centering{
\setlength{\tabcolsep}{1pt}    
\begin{tabular}{cc}
%ISOTROPIC FANNING & ANISTROPIC FANNING \\
\hspace{0.8cm}{CSD} & \hspace{0.8cm}{FORECAST} \\
\includegraphics[width=0.48\columnwidth]{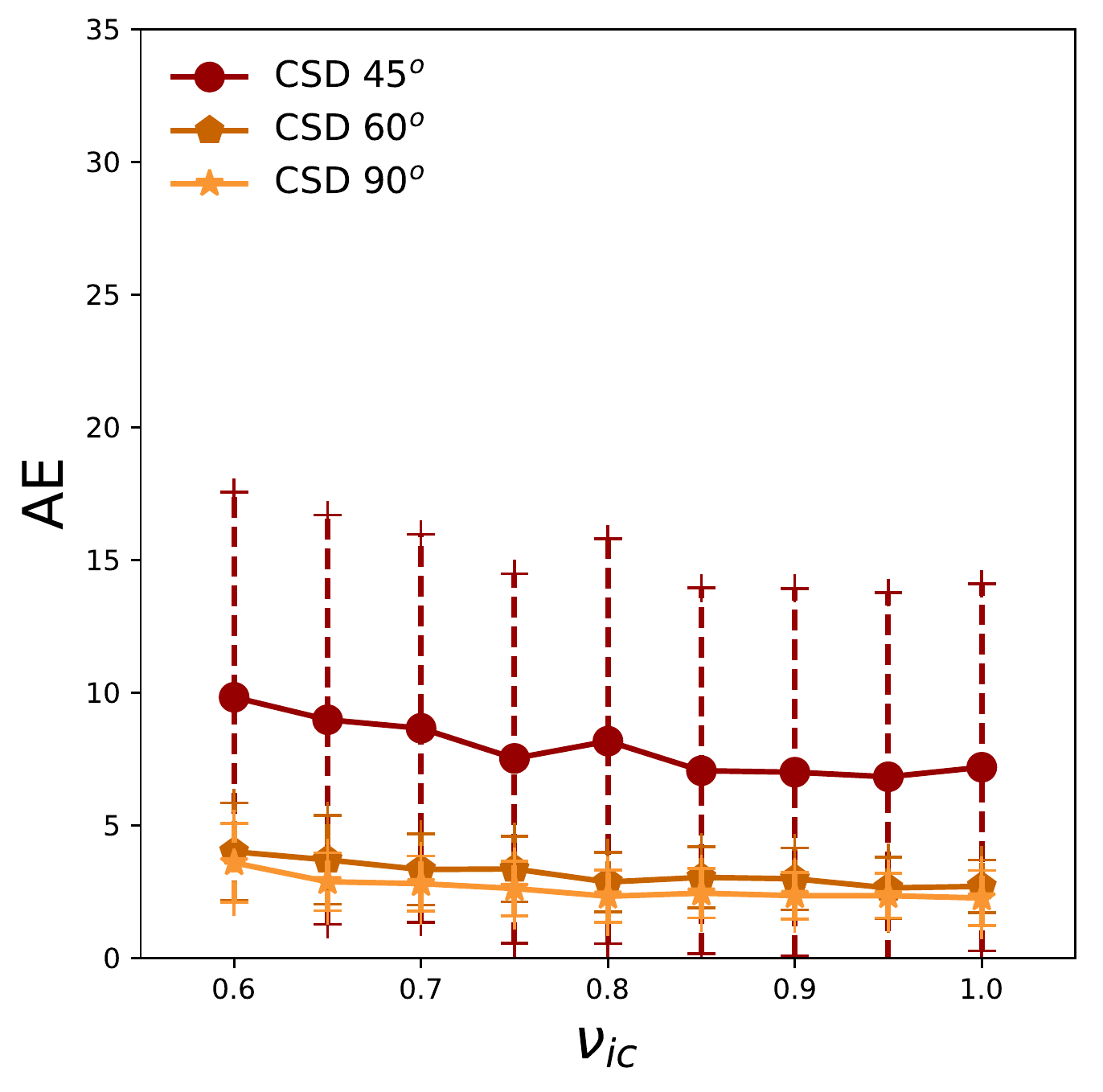} &
\includegraphics[width=0.48\columnwidth]{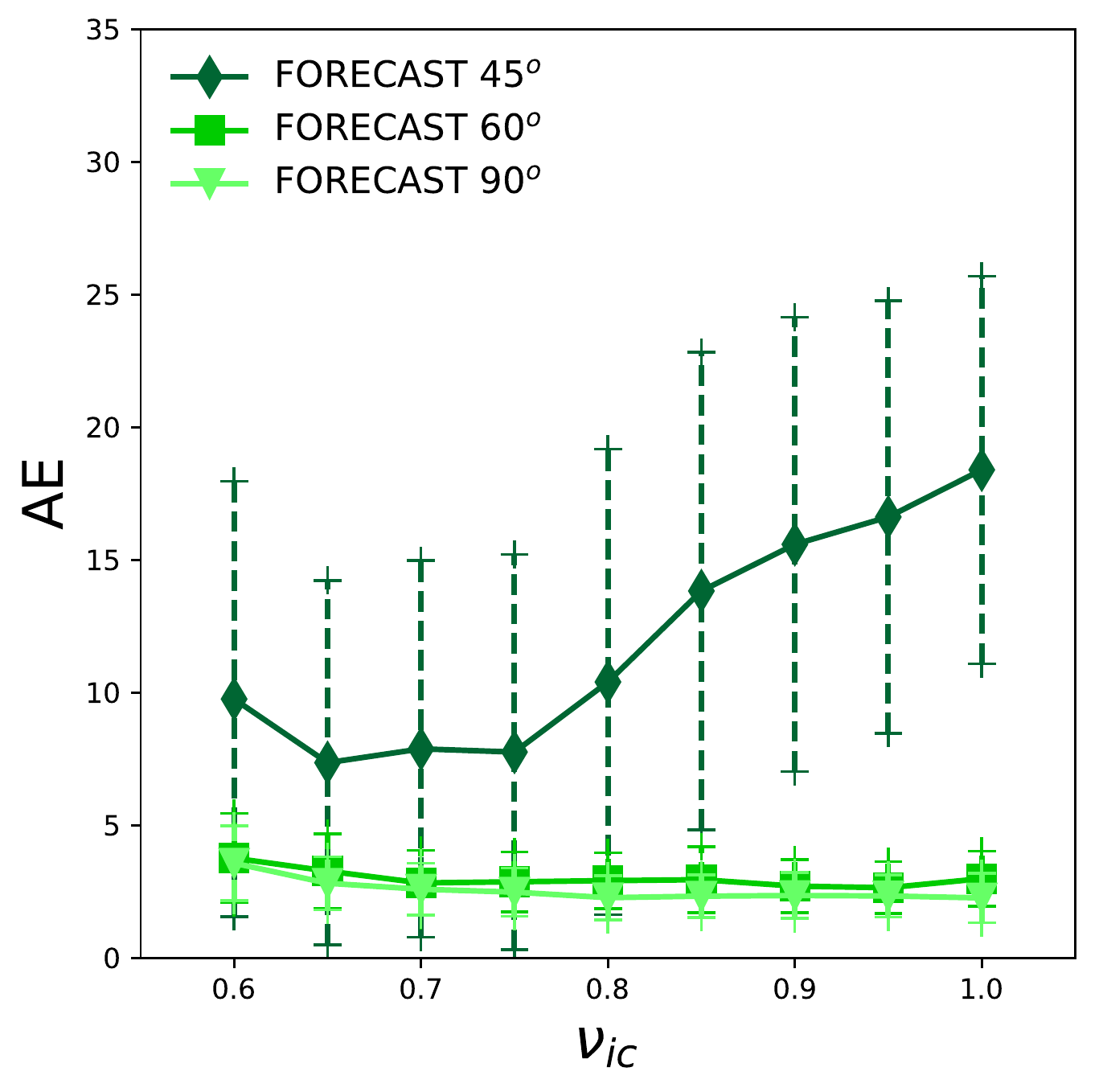} \\
\hspace{0.8cm}{NODDI-SH} & \hspace{0.8cm}{AVERAGE} \\
\includegraphics[width=0.48\columnwidth]{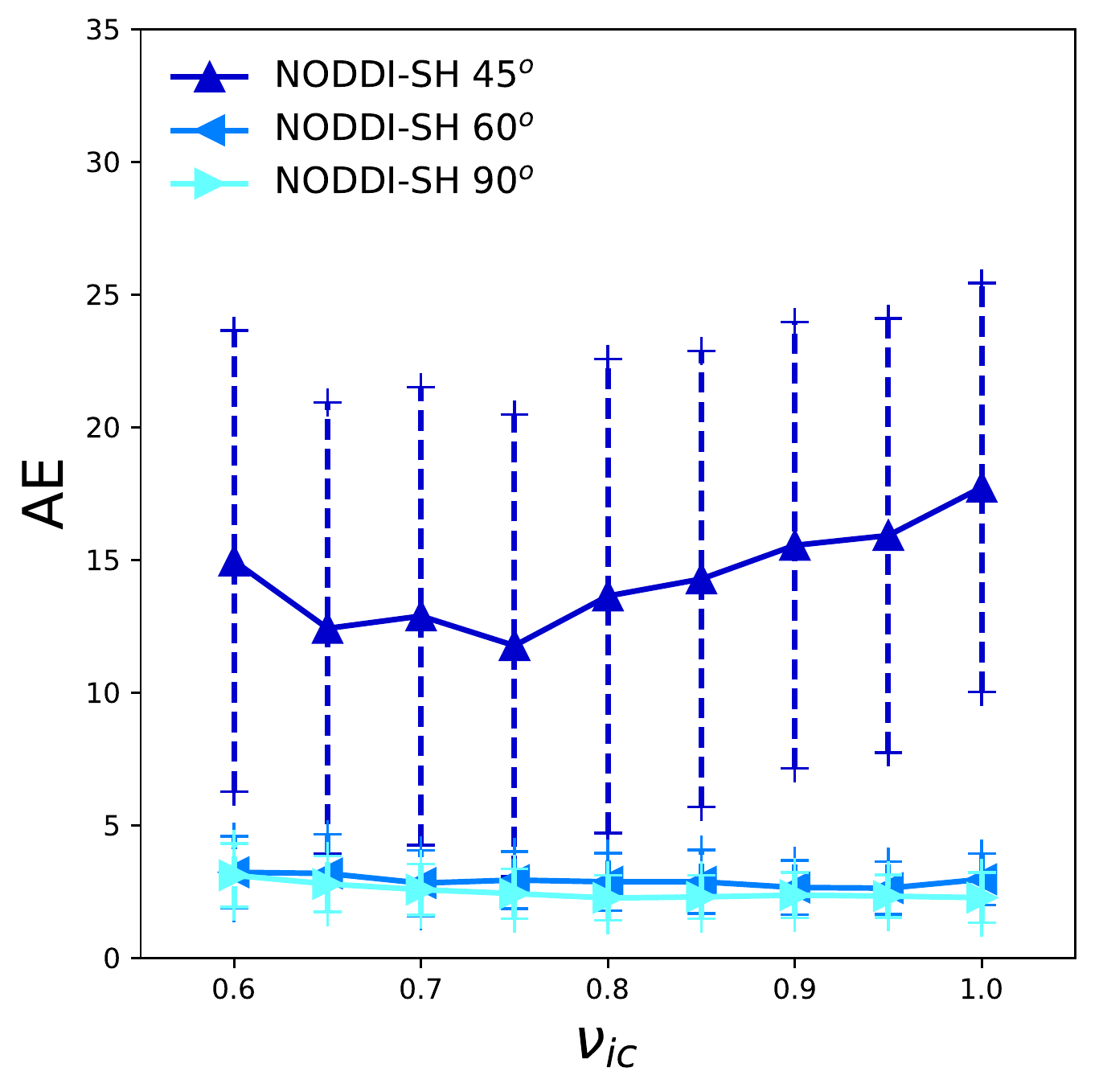} &
\includegraphics[width=0.48\columnwidth]{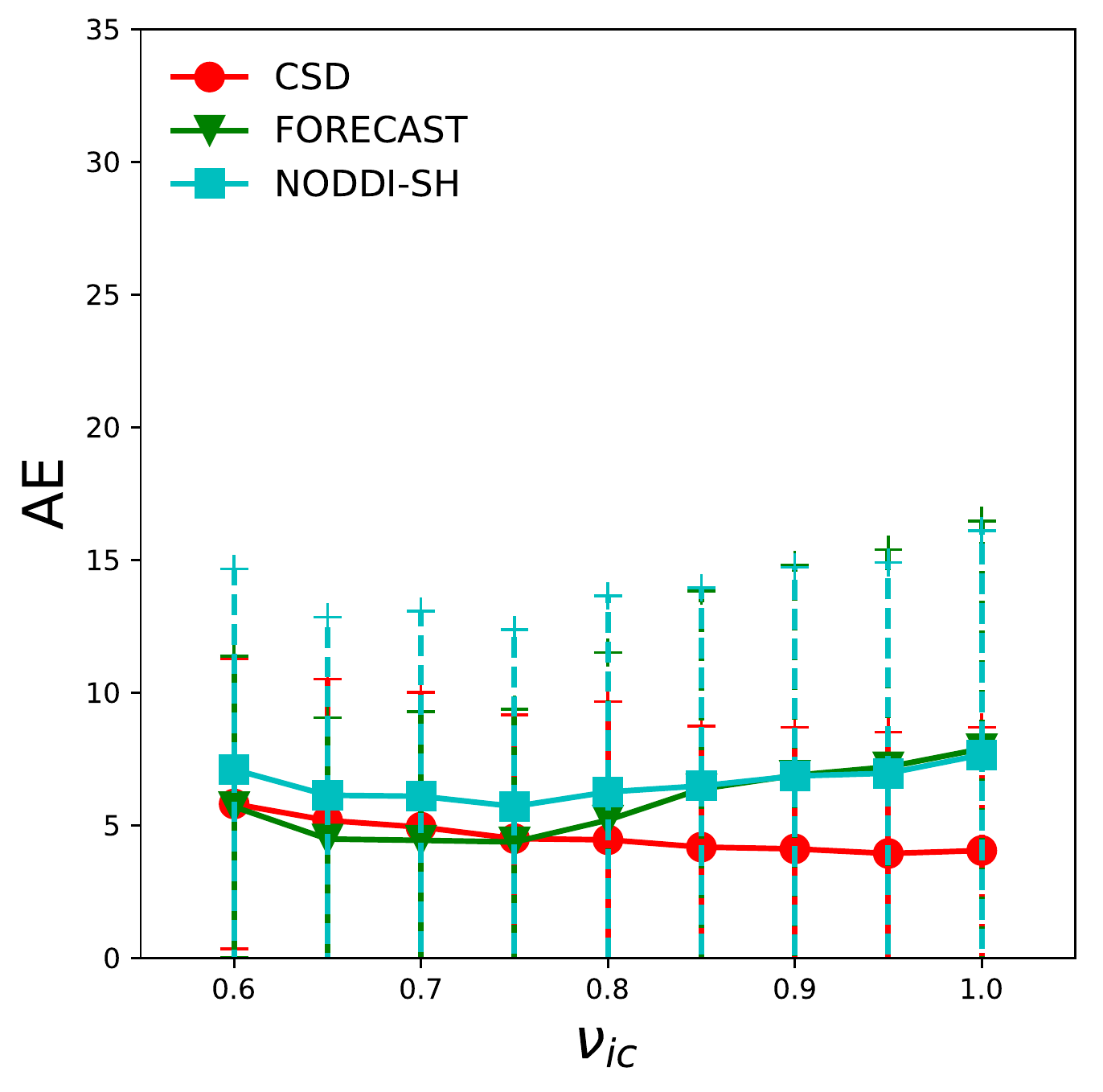} \\
  \end{tabular} }
\caption{Angular error (AE) corresponding to the fODF calculated with CSD, FORECAST, and NODDI-SH. The average represents the mean of the AE at 45, 60, and 90 degrees for each models.}
\label{fig:sim_cross}
\end{figure} 

In order to asses the precision of the fODF, the angular error (AE) between the fODF peaks and the ground truth mean directions of the simulated fibers was determined.
Figure \ref{fig:sim_cross} shows the AE estimated for CSD, FORECAST, and NODDI-SH fODFs. In particular, the AE was estimated as the average of the angular distance of the two ground truth directions with the closest fODF peak. CSD was able to retrieve the principal directions of diffusion almost perfectly in the 90 and 60 degrees cases ($AE < 4^o$) while its AE increased in the most difficult case, the 45 degrees crossing, with an average AE of 7.9$^o$ and the highest standard deviation. 
For both, FORECAST and NODDI-SH, the AE was less than 5$^o$ in the 90 and 60 degrees crossing cases. In the 45 degrees crossing the error was lower for low $\nu_{ic}$ values and increased progressively with the neurite density. CSD performed better than FORECAST and NODDI-SH in voxels featuring high restriction. However, it is important to stress that CSD results depend on the choice of the response function. In our experiments, we purposely selected the response function to be very close to the intracellular signal in order to obtain high-resolution fODFs. 
Both NODDI-SH and FORECAST are able to adapt its response function in every voxel, which is a clear advantage in the case of real tissues.

\subsection{Results on HCP data}
\label{res:real}
NODDI-SH three compartments model is designed to capture the heterogeneity of the brain tissues. However, there are no guarantees that NODDI-SH estimation of neurites density and volume fraction corresponds to the underlying tissue composition, as it is the case for all models. Therefore, results on \textit{in-vivo} data should be taken with care since we lack a histological ground truth to validate these data. Due to the lack of ground truth, the performance of NODDI-SH in \textit{in-vivo} data was assessed by comparison with NODDI that was used as a benchmark. %We can still compare NODDI-SH volume fraction estimation with the one obtained using NODDI, in order to qualitative evaluate \textit{in-vivo} results.
 
\begin{figure}[!t]
\centering{
\setlength{\tabcolsep}{5pt}    
\begin{tabular}{ccc}
%ISOTROPIC FANNING & ANISTROPIC FANNING \\
& INTRACELLULAR & CSF \\
\rotatebox{90}{\hspace{1.8cm}NODDI} &
\includegraphics[width=0.40\columnwidth, trim={2cm 2cm 1cm 2cm}, clip]{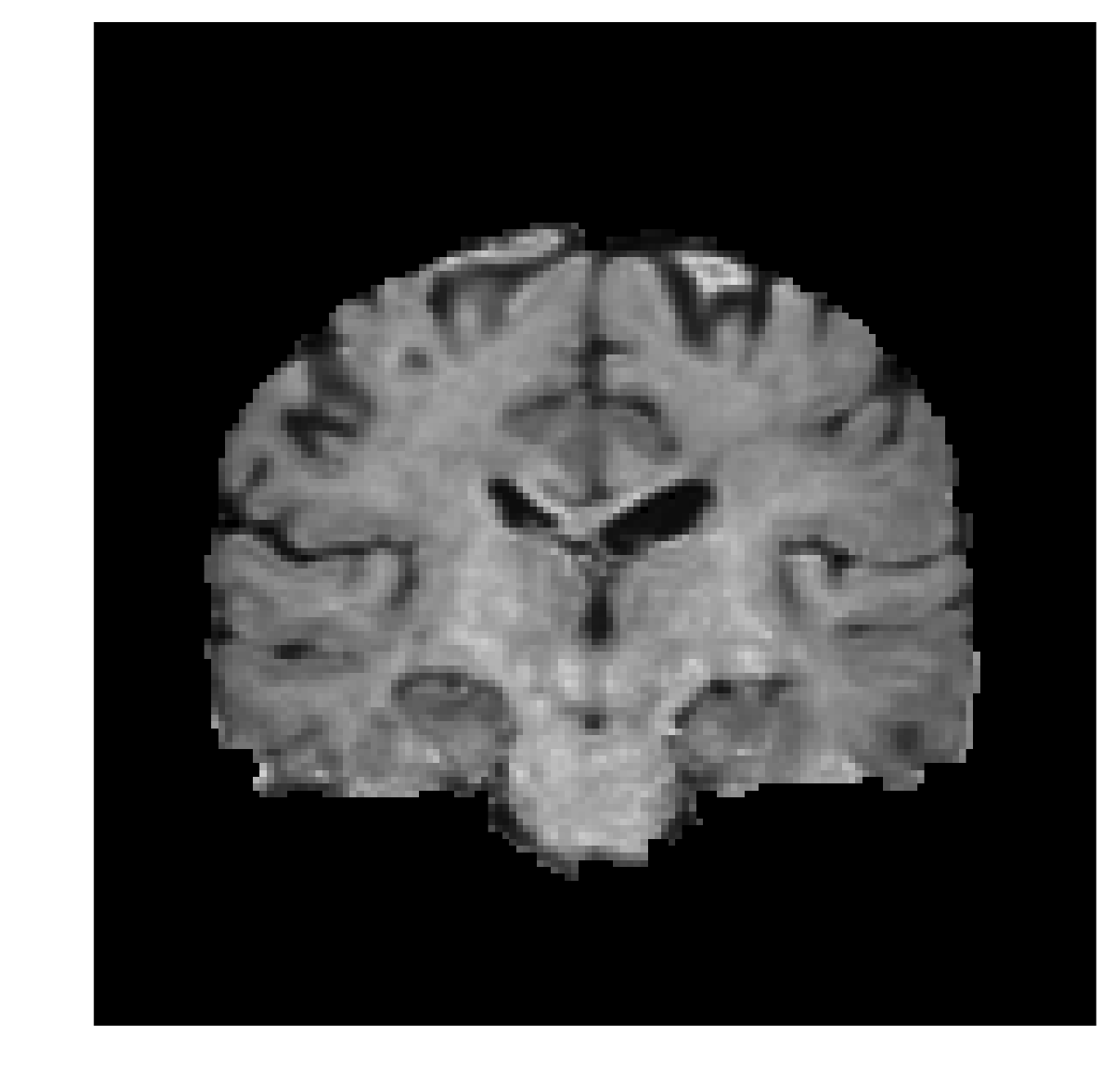} &
\includegraphics[width=0.40\columnwidth, trim={2cm 2cm 1cm 2cm}, clip]{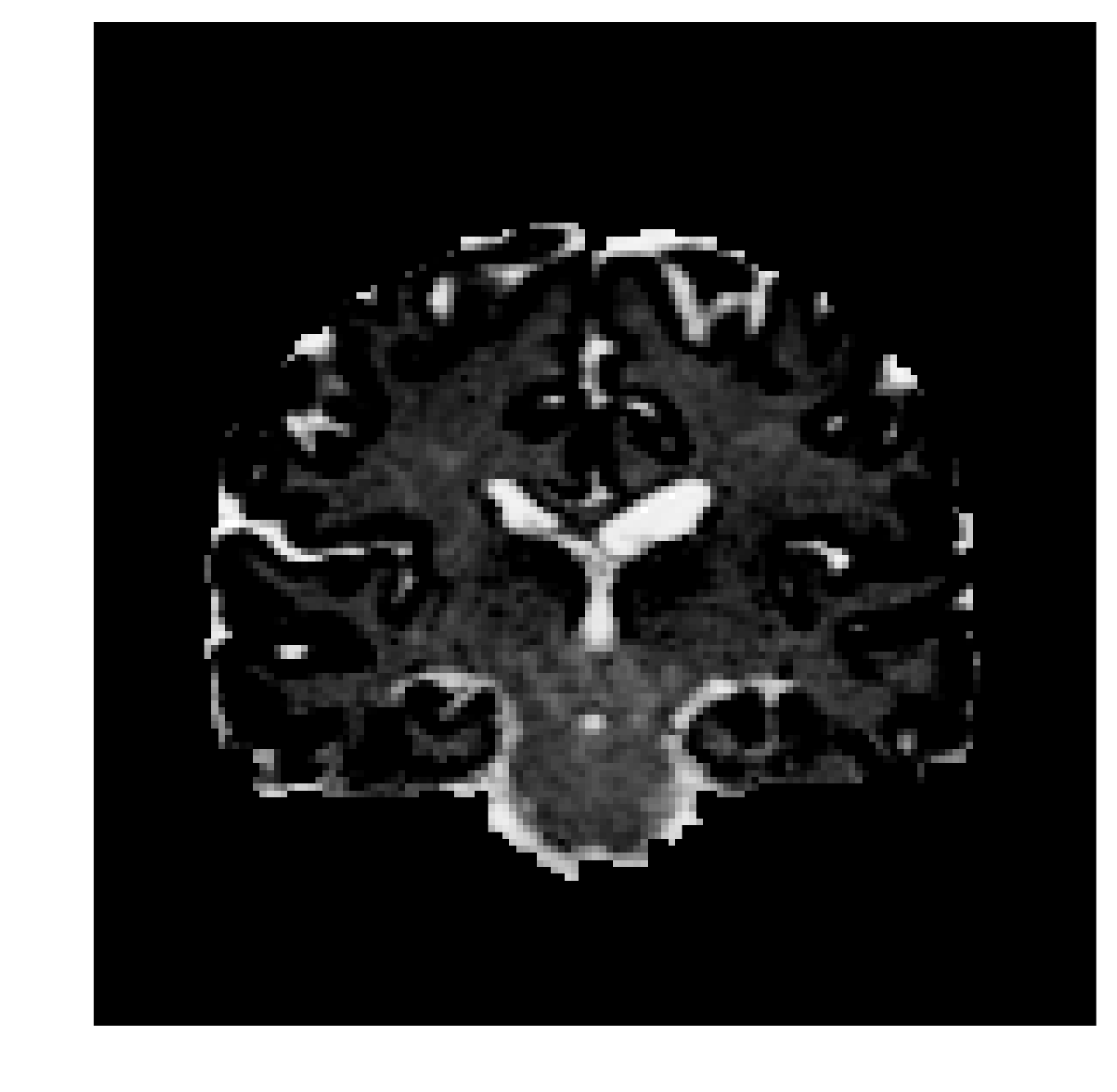} \\
\rotatebox{90}{\hspace{1.7cm}NODDI-SH} &
\includegraphics[width=0.40\columnwidth, trim={2cm 2cm 1cm 2cm}, clip]{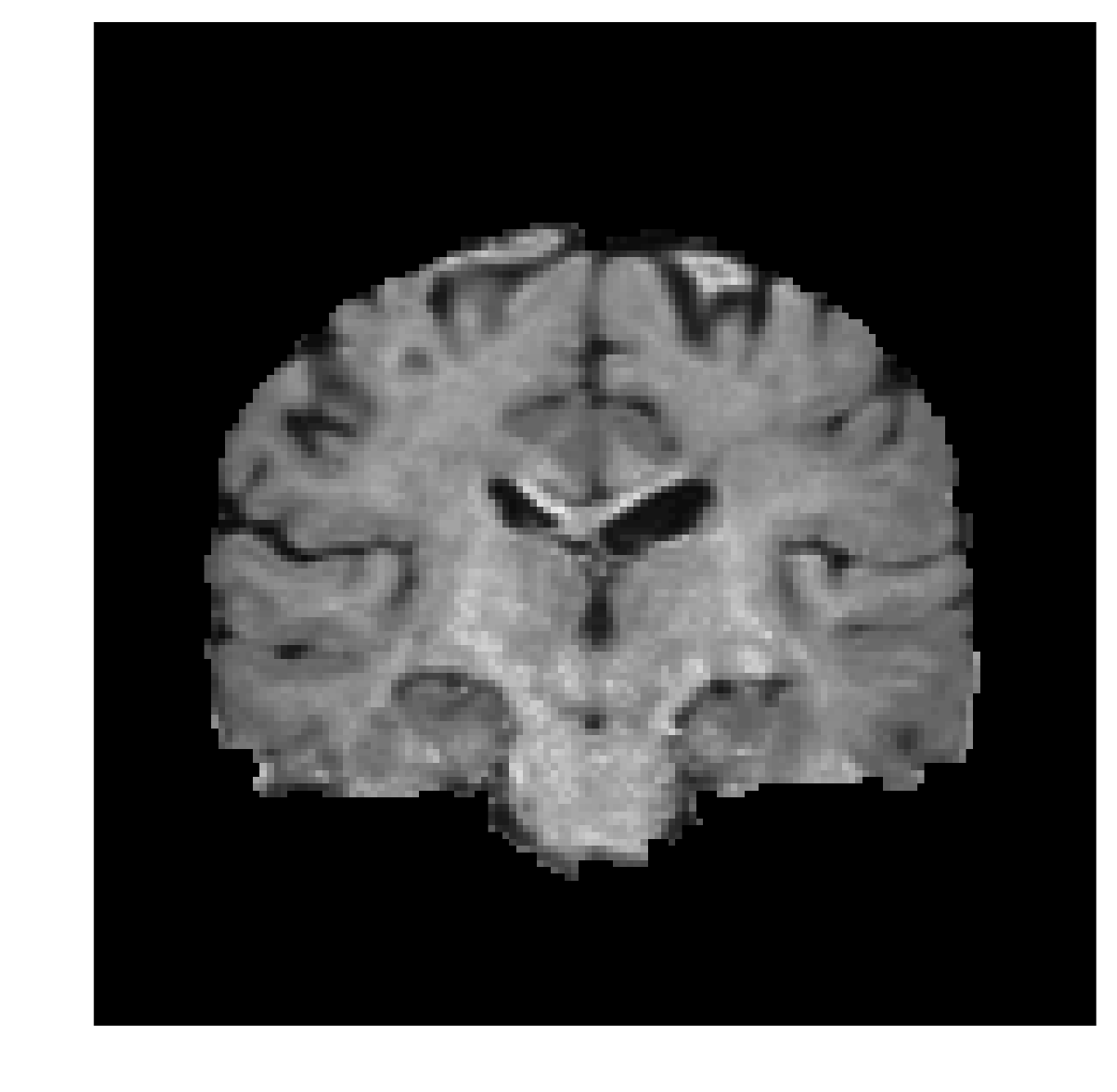} &
\includegraphics[width=0.40\columnwidth, trim={2cm 2cm 1cm 2cm}, clip]{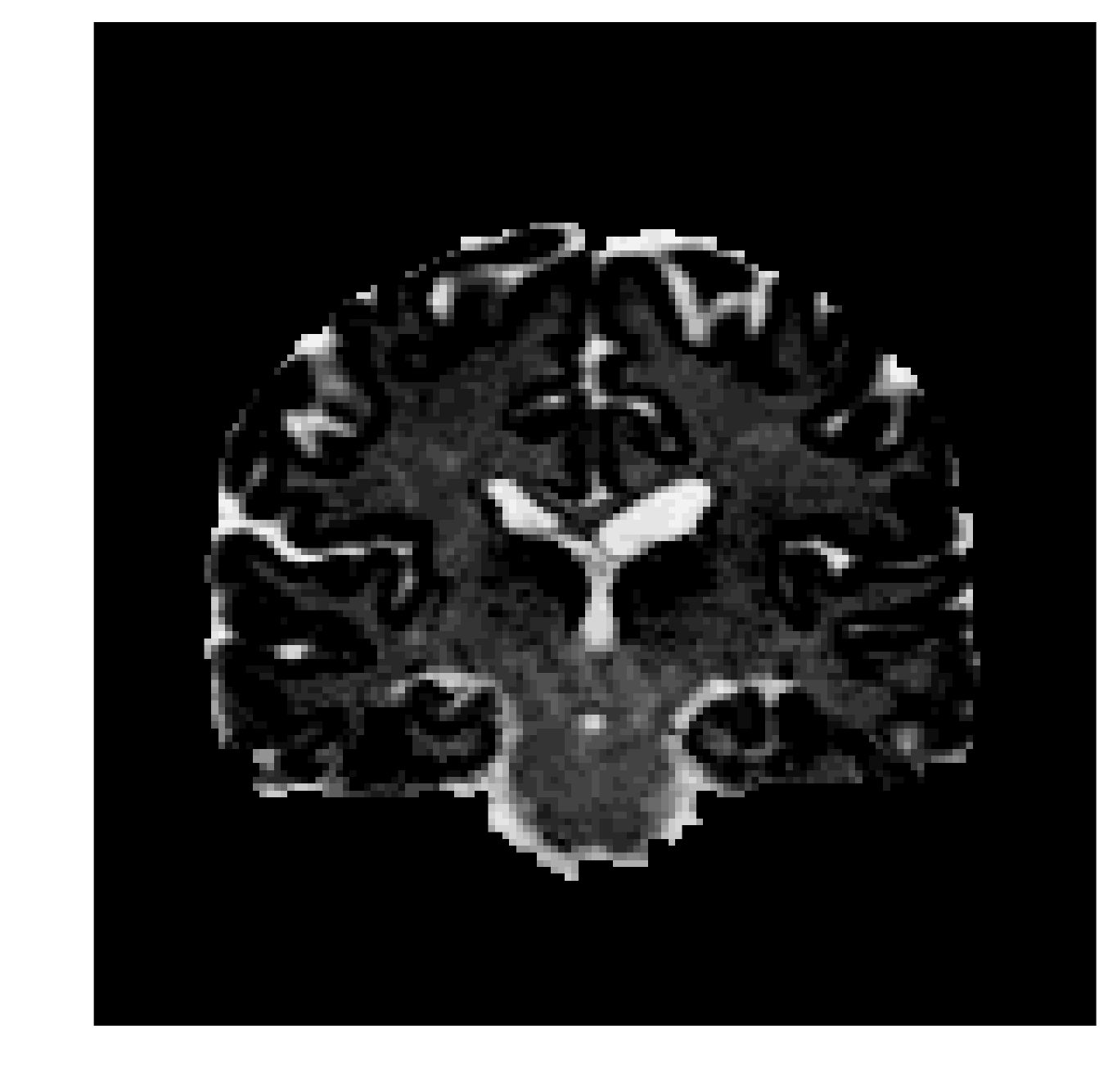} \\
& & \hspace{-6cm}\includegraphics[width=0.45\columnwidth, trim={0cm 0cm 0cm 11cm}, clip]{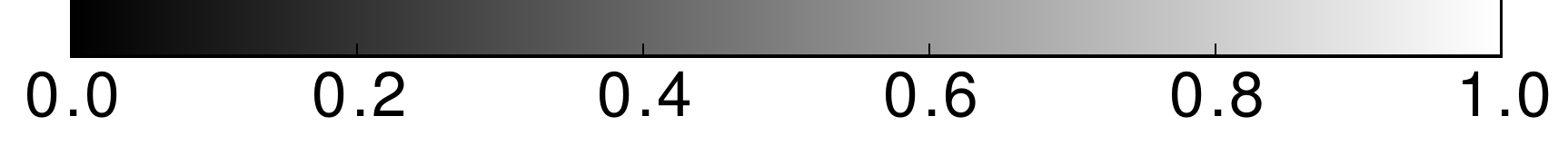} \\
  \end{tabular} }
\caption{Intracellular volume fraction (left) and CSF volume fraction (right) estimation of a coronal slice of one HCP brain calculated using NODDI (top row) and NODDI-SH (bottom row).}
\label{fig:noddi_shadow}
\end{figure}

Figure \ref{fig:noddi_shadow} shows the intracellular and CSF volume fractions estimated using NODDI and NODDI-SH, respectively.  The maps obtained with the two techniques are very similar. NODDI-SH intracellular volume fraction appears to be slightly more contrasted with respect to NODDI, in particular in single fiber areas such as the corpus callosum and the corticospinal tract. NODDI-SH CSF map presents more voxels with higher partial volume of CSF near the cortex, but in general, the agreement between the two maps is extremely high. For both models, CSF fraction maps present low, but non-zero, values in the white matter. Although the presence of free water between the axons is not likely, these values could be explained by an intrinsic error of the three compartment model used by both techniques. 

\begin{figure}[!t]
\centering{
\setlength{\tabcolsep}{5pt}    
\begin{tabular}{cc}
%ISOTROPIC FANNING & ANISTROPIC FANNING \\
NODDI MSE & NODDI-SH MSE \\
\includegraphics[width=0.40\columnwidth, trim={2cm 2cm 1cm 2cm}, clip]{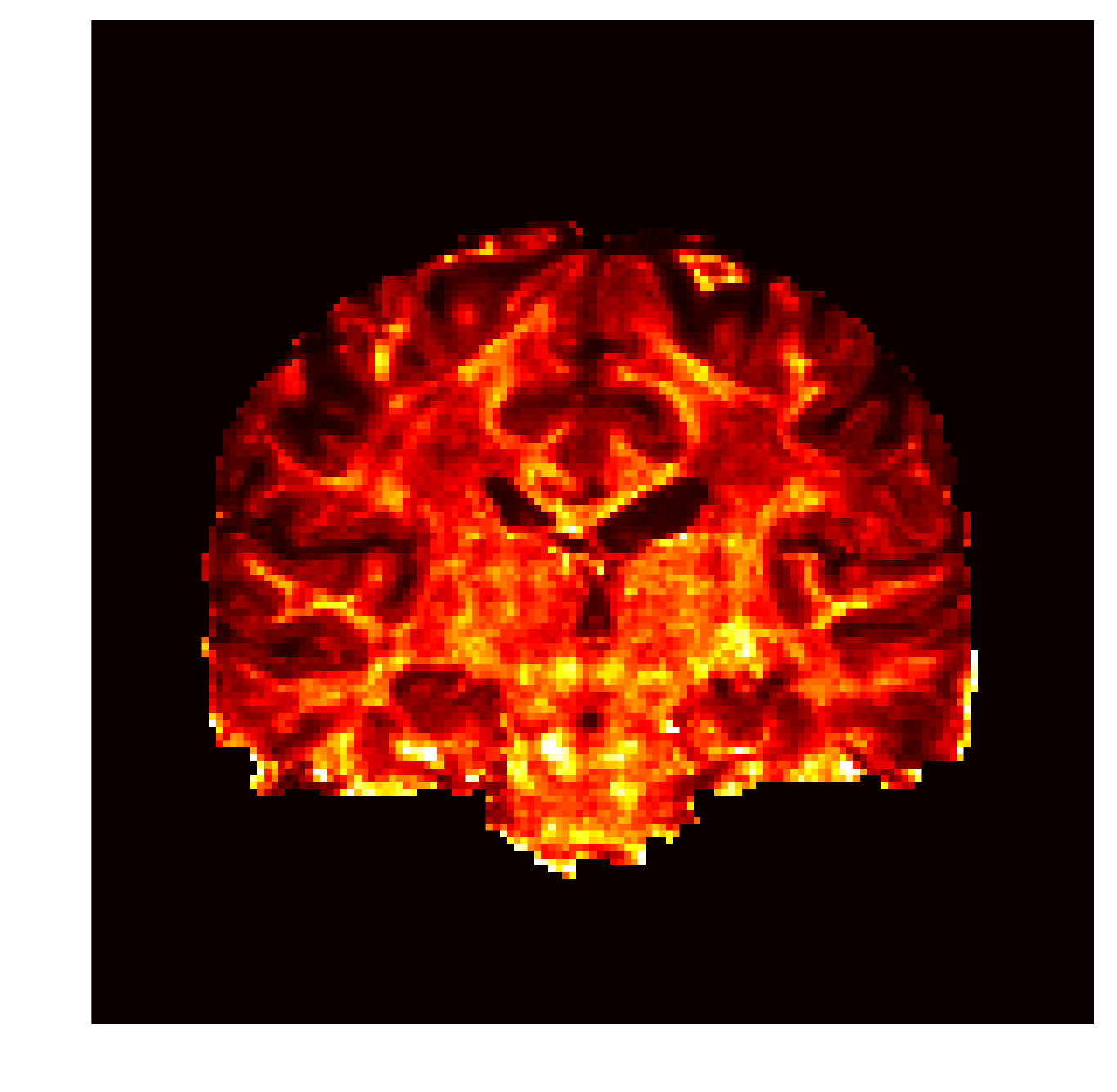} &
\includegraphics[width=0.40\columnwidth, trim={2cm 2cm 1cm 2cm}, clip]{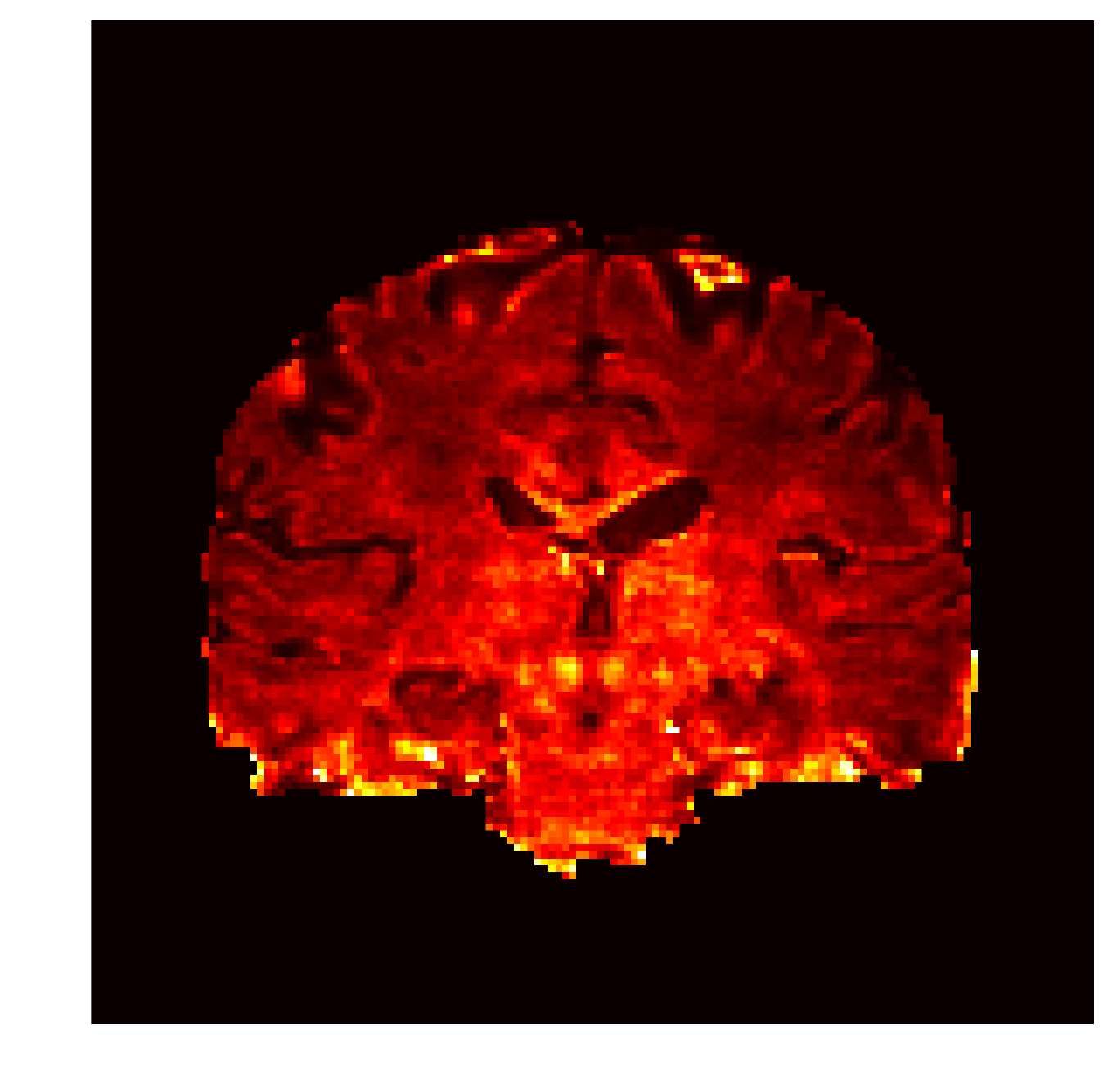} \\
& \hspace{-6cm}\includegraphics[width=0.45\columnwidth, trim={0cm 0cm 0cm 11cm}, clip]{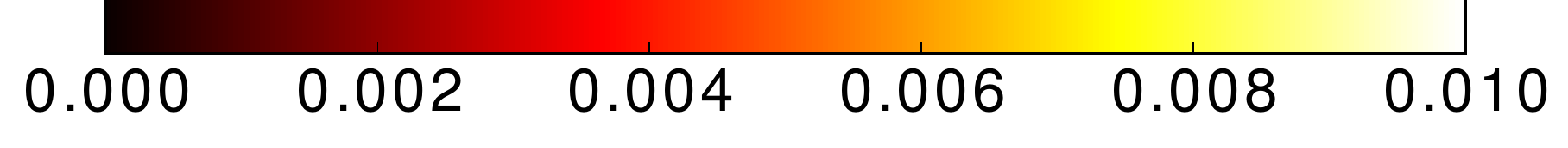} \\
  \end{tabular} }
\caption{NODDI MSE (left) and NODDI-SH MSE (right) of a coronal slice of one HCP brain.}
\label{fig:noddi_shadow_mse}
\end{figure}

Figure \ref{fig:noddi_shadow_mse} shows the MSE calculated between the normalized diffusion signal and the estimated diffusion signal of NODDI and NODDI-SH. It is worth to mention that in this picture we use a very narrow range of values in order to emphasize the contrast and that both techniques were able to estimate the diffusion signal accurately. 
NODDI MSE appears to be higher in white matter, and in particular in single fiber areas, with respect to gray matter or CSF where it is close to zero.
NODDI-SH MSE presents a more uniform pattern in the white matter, gray matter, and CSF. The highest MSE values for NODDI-SH were found in the corpus callosum and in the basal nuclei. 

\begin{figure}[!t]
\centering{
\setlength{\tabcolsep}{5pt}    
\begin{tabular}{cc}
%ISOTROPIC FANNING & ANISTROPIC FANNING \\
\small{NODDI MSE $>$ NODDI-SH MSE} & \small{$\vert$NODDI MSE - NODDI-SH MSE$\vert$} \\
\includegraphics[width=0.43\columnwidth, trim={0cm 0cm 0cm 0cm}, clip]{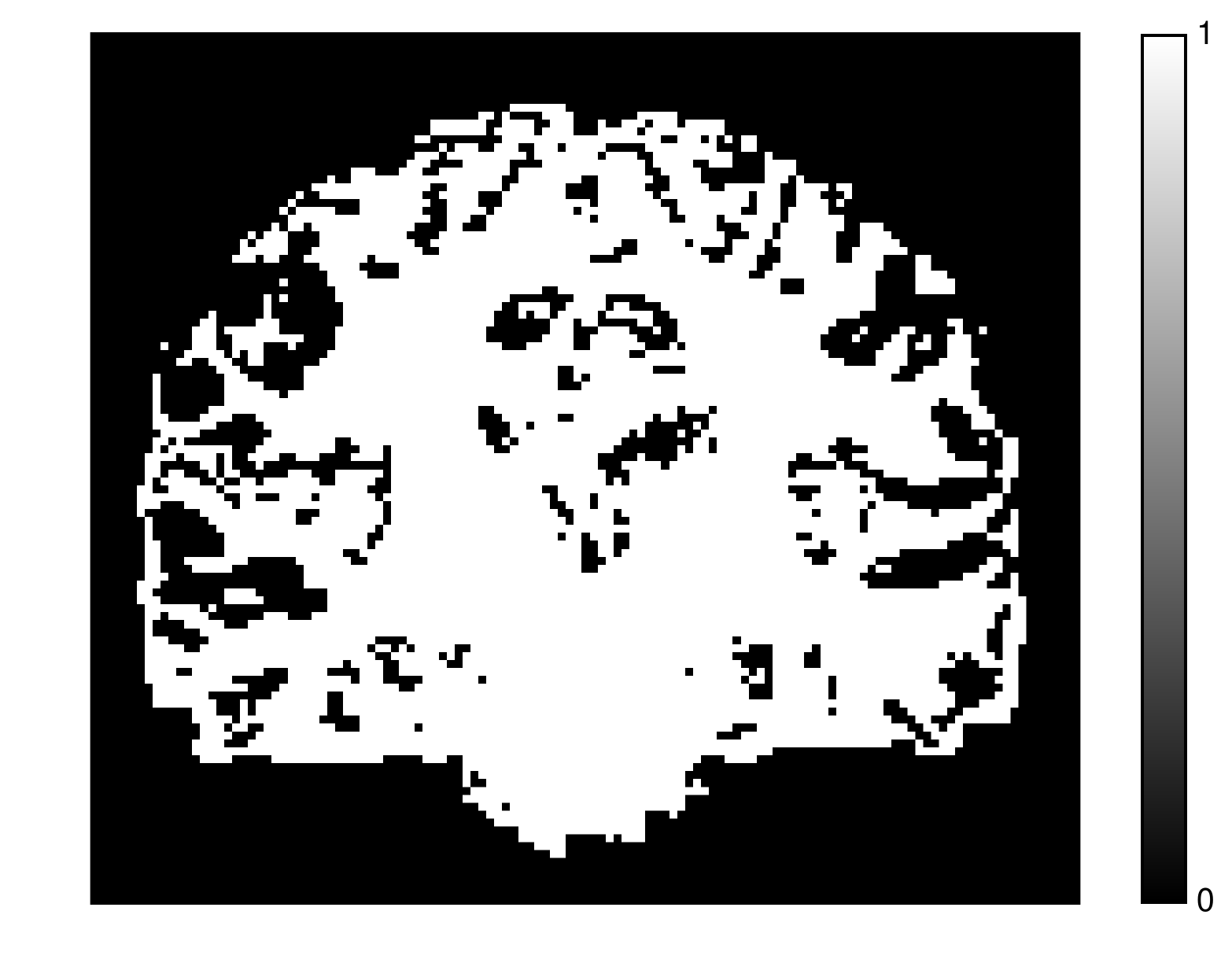} &
\includegraphics[width=0.45\columnwidth, trim={0cm 0cm 0cm 0cm}, clip]{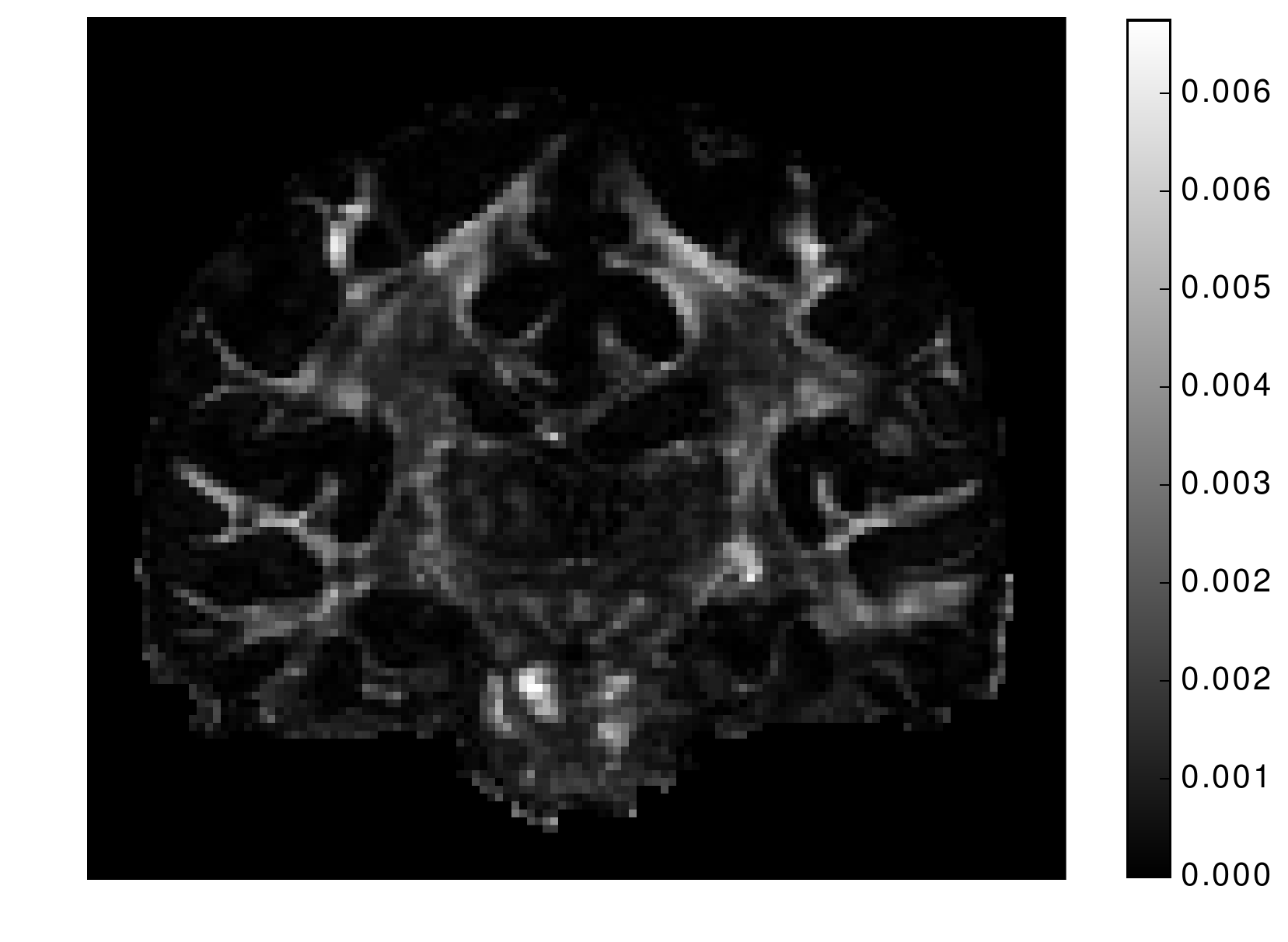} \\
  \end{tabular} }
\caption{Regions of the brain in which NODDI-SH obtain a lower MSE with respect to NODDI (left), and absolute value of the difference of the MSE (right). }
\label{fig:noddi_shadow_diff}
\end{figure}

Figure \ref{fig:noddi_shadow_diff} (left) highlights the regions of the brain in which NODDI-SH features a lower MSE than NODDI. Almost all the white matter appears to be white, which seems to indicate that NODDI-SH is able to reconstruct the diffusion signal more accurately. These results could be explained by the fact that the Watson distribution is not able to model most of the brain fiber configurations, while NODDI-SH can model any kind of fiber arrangement including fanning and crossing.
In most of the gray matter and CSF NODDI fits the diffusion signal slightly better than NODDI-SH, although the absolute value of the difference of the MSE is lower than that obtained in white matter (Figure \ref{fig:noddi_shadow_diff}, right). 

%%%%%%%%%%%%%%%%%%%%%%%%%%%%%%%%%%%%%%%%%%%%%%%%%%%%%%%%%%%%%%%%%%%%%%%%%%%%%%%%%
\begin{figure}[!t]
\centering{
\setlength{\tabcolsep}{1pt}    
\begin{tabular}{cc}
%ISOTROPIC FANNING & ANISTROPIC FANNING \\
ROI & CSD\\
\raisebox{0.5cm}{\includegraphics[width=0.45\columnwidth, trim={2cm 2cm 1cm 2cm}, clip]{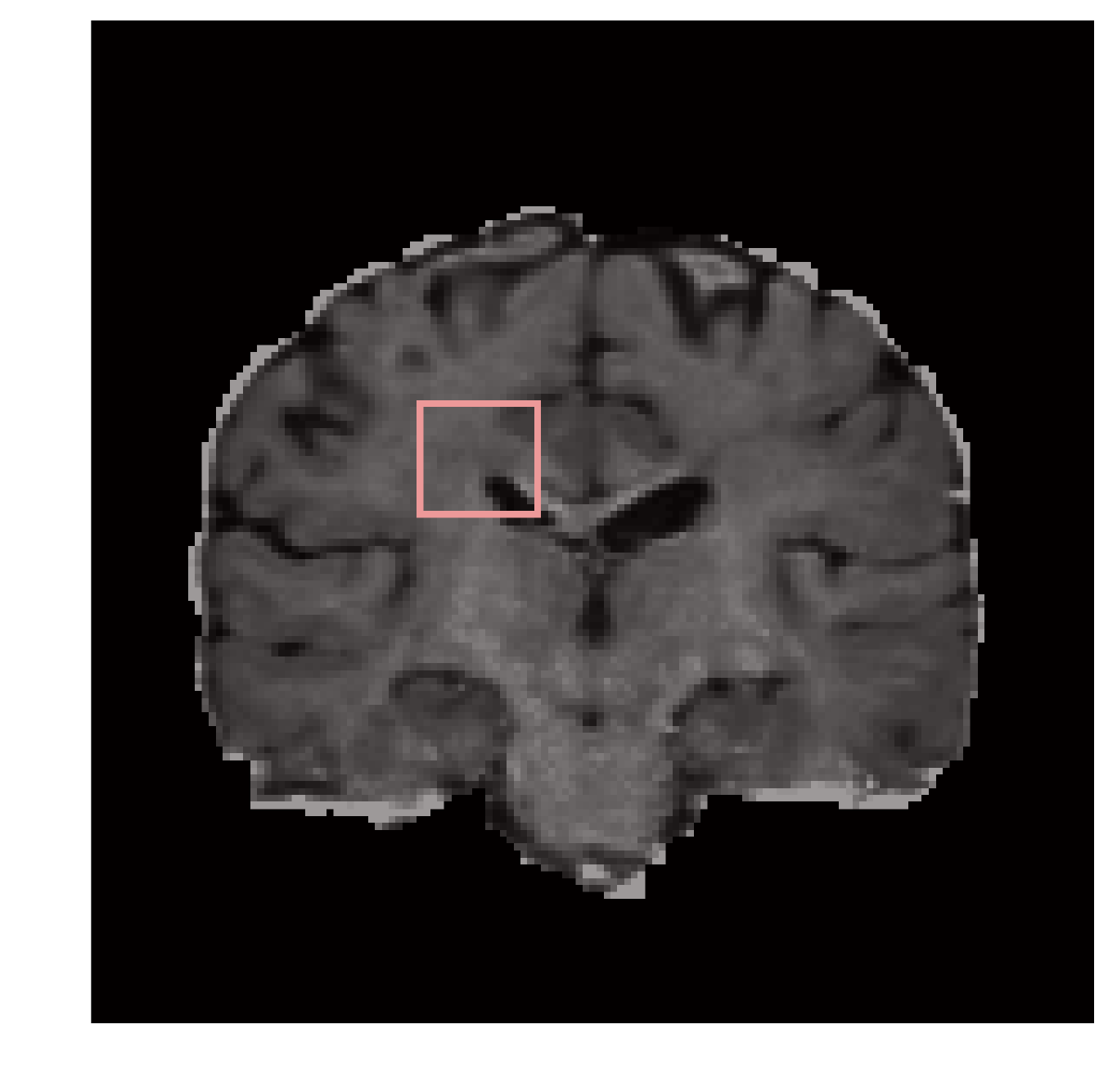}} & \includegraphics[width=0.48\columnwidth, trim={3.75cm 4cm 3.75cm 4cm}, clip]{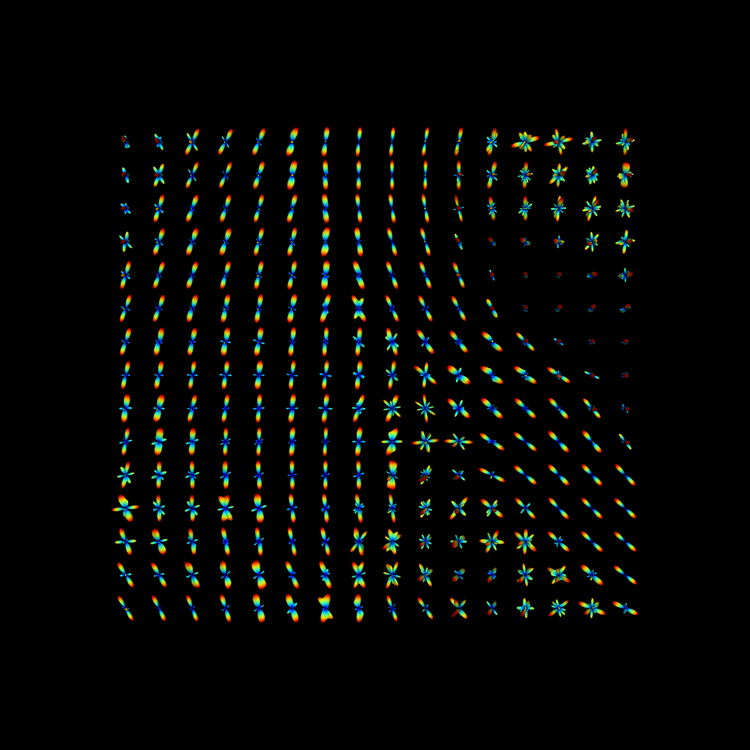} \\
FORECAST & NODDI-SH \\
\includegraphics[width=0.48\columnwidth, trim={3.75cm 4cm 3.75cm 4cm}, clip]{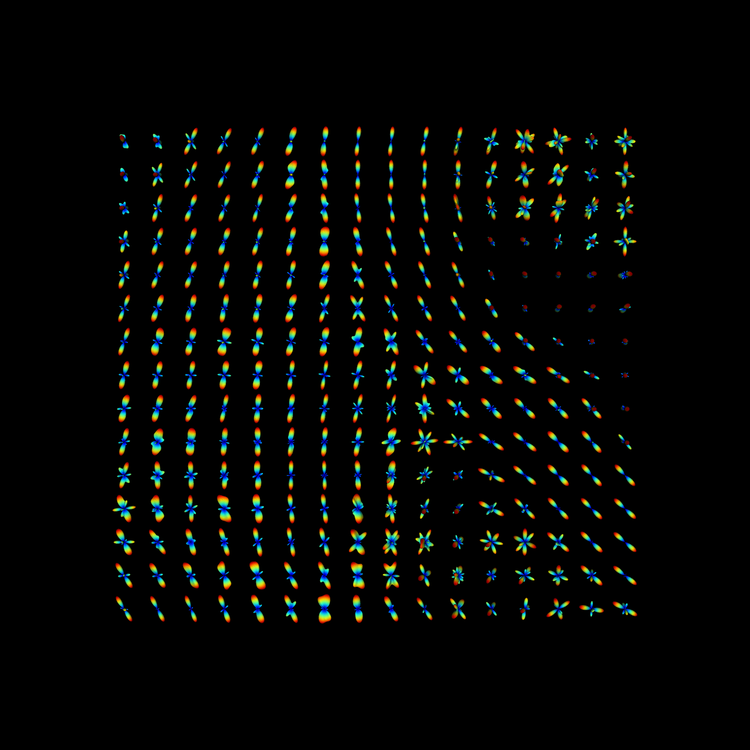} & \includegraphics[width=0.48\columnwidth, trim={3.75cm 4cm 3.75cm 4cm}, clip]{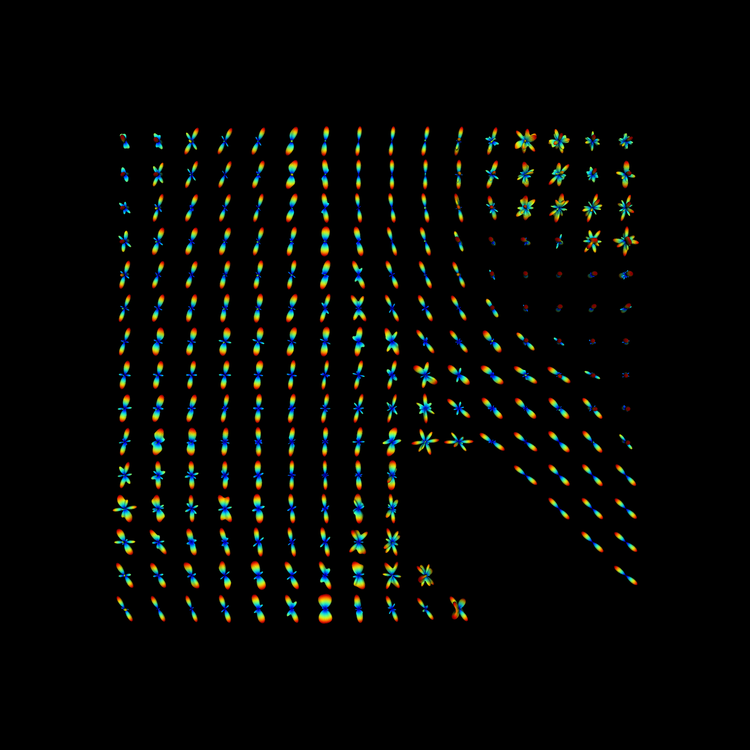} \\
  \end{tabular} }
\caption{CSD, FORECAST, and NODDI-SH fODFs, calculated on region of interest taken from a coronal slice of the HCP data.}
\label{fig:hcp_odf}
\end{figure} 
Figure \ref{fig:hcp_odf} shows the fODFs obtained by CSD, FORECAST, and NODDI-SH on a region of interest (ROI) of the HCP brain (top left) which includes crossings, fannings, a part of the corpus callosum, and the right ventricle. There are not substantial differences between the three fODFs. 
The main advantage of using NODDI-SH to calculate the fODF is that it is possible to directly filter out the fODFs which do not belong to white matter. In Figure \ref{fig:hcp_odf} (bottom right) NODDI-SH fODFs, for instance, only the voxels with $\nu_{ic} \ge 0.3$ were considered for display. In order to do the same with CSD fODFs, it would be necessary to integrate the CSD results with those obtained by a multi-compartmental model, such as NODDI.

\subsection{Robustness with respect to the number of samples}
\label{subsampling}
\begin{figure}[!ht]
\centering{
\setlength{\tabcolsep}{1pt}    
\begin{tabular}{cccc}
%ISOTROPIC FANNING & ANISTROPIC FANNING \\
& 90 directions & 60 directions & 30 directions \\
\rotatebox{90}{\hspace{0.5cm}$b = 1000 s/mm^2$}  &
\includegraphics[width=0.3\columnwidth, trim={2cm 2cm 1cm 2cm}, clip]{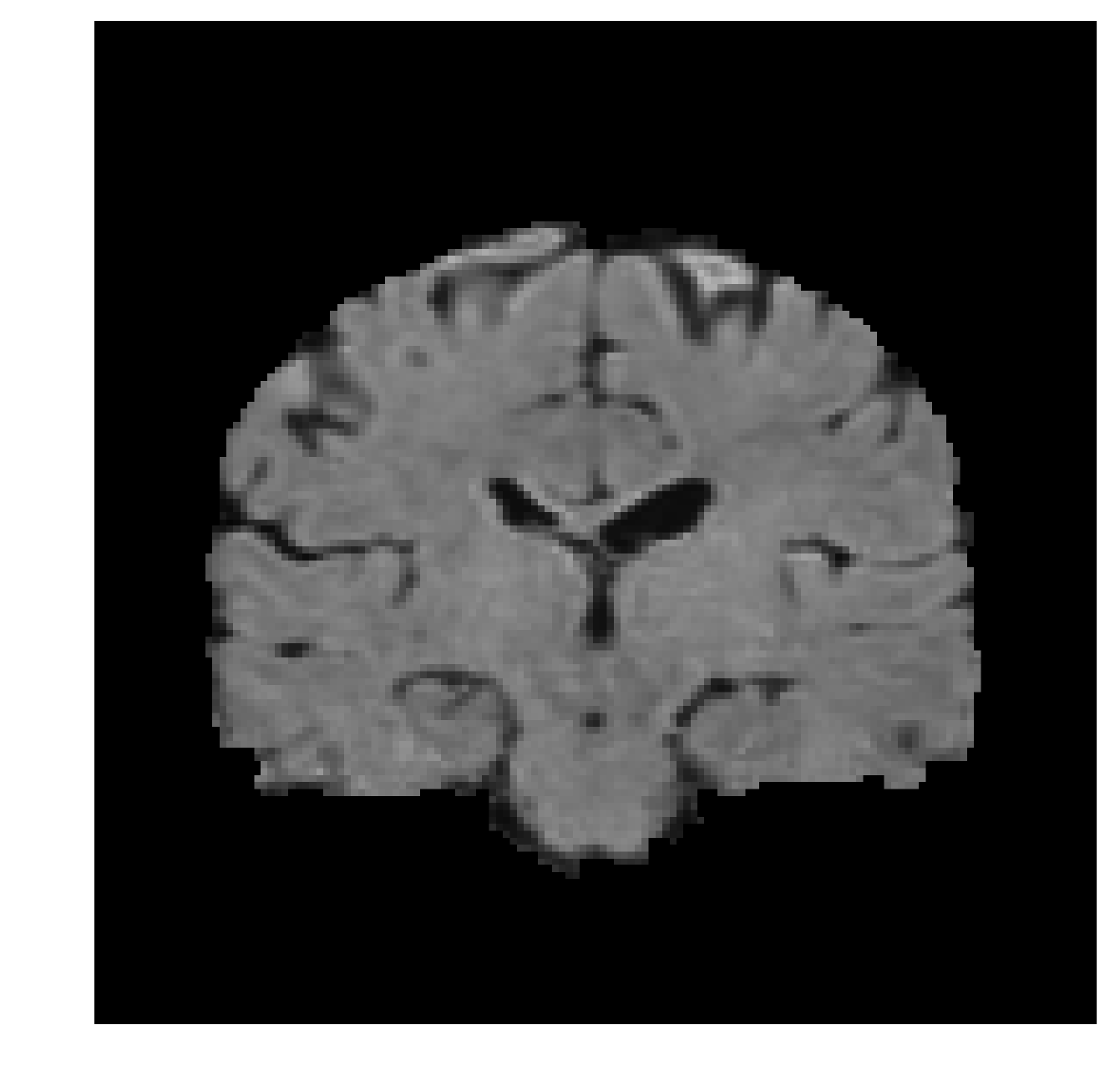}  &
\includegraphics[width=0.3\columnwidth, trim={2cm 2cm 1cm 2cm}, clip]{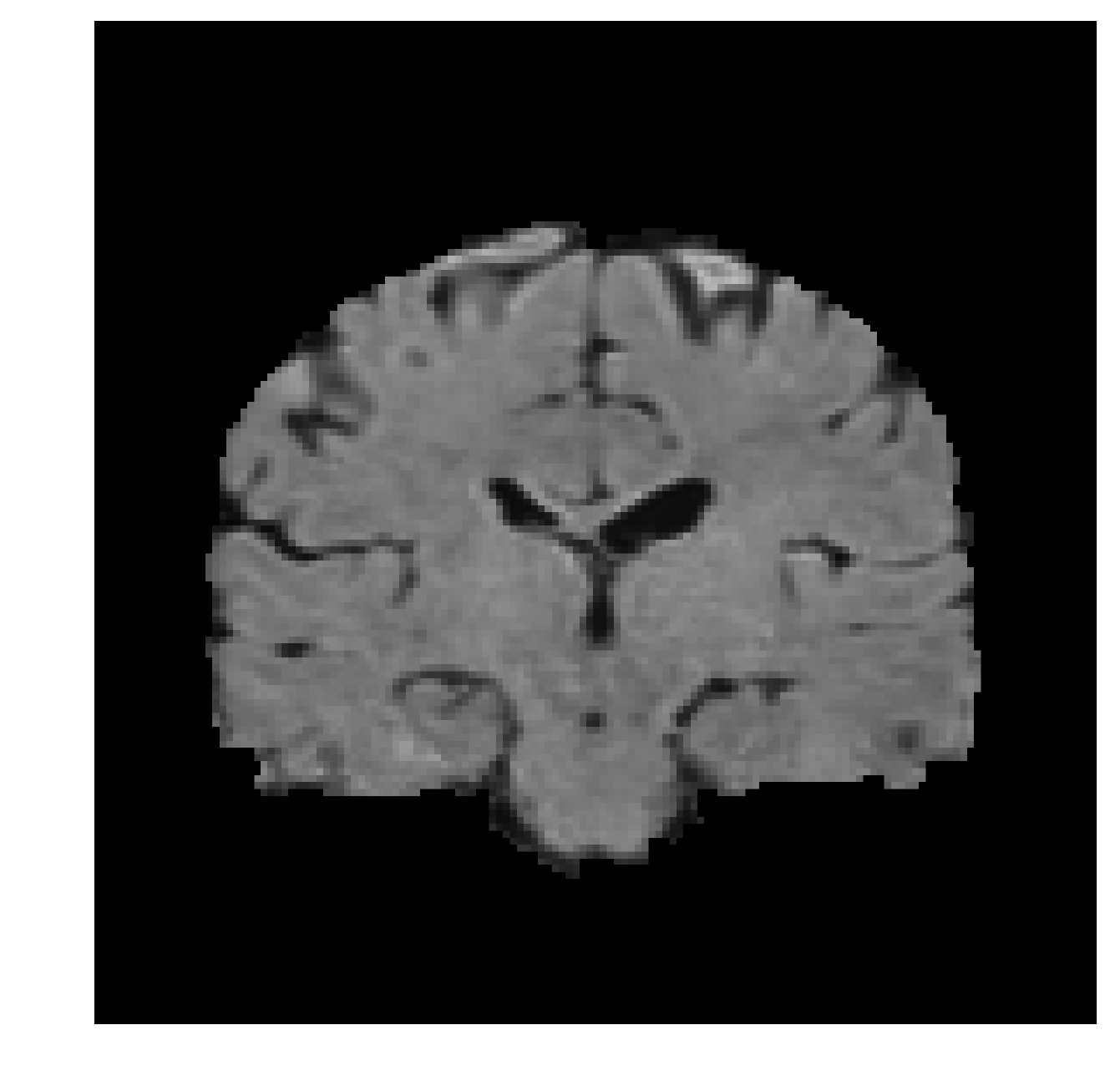}  &
\includegraphics[width=0.3\columnwidth, trim={2cm 2cm 1cm 2cm}, clip]{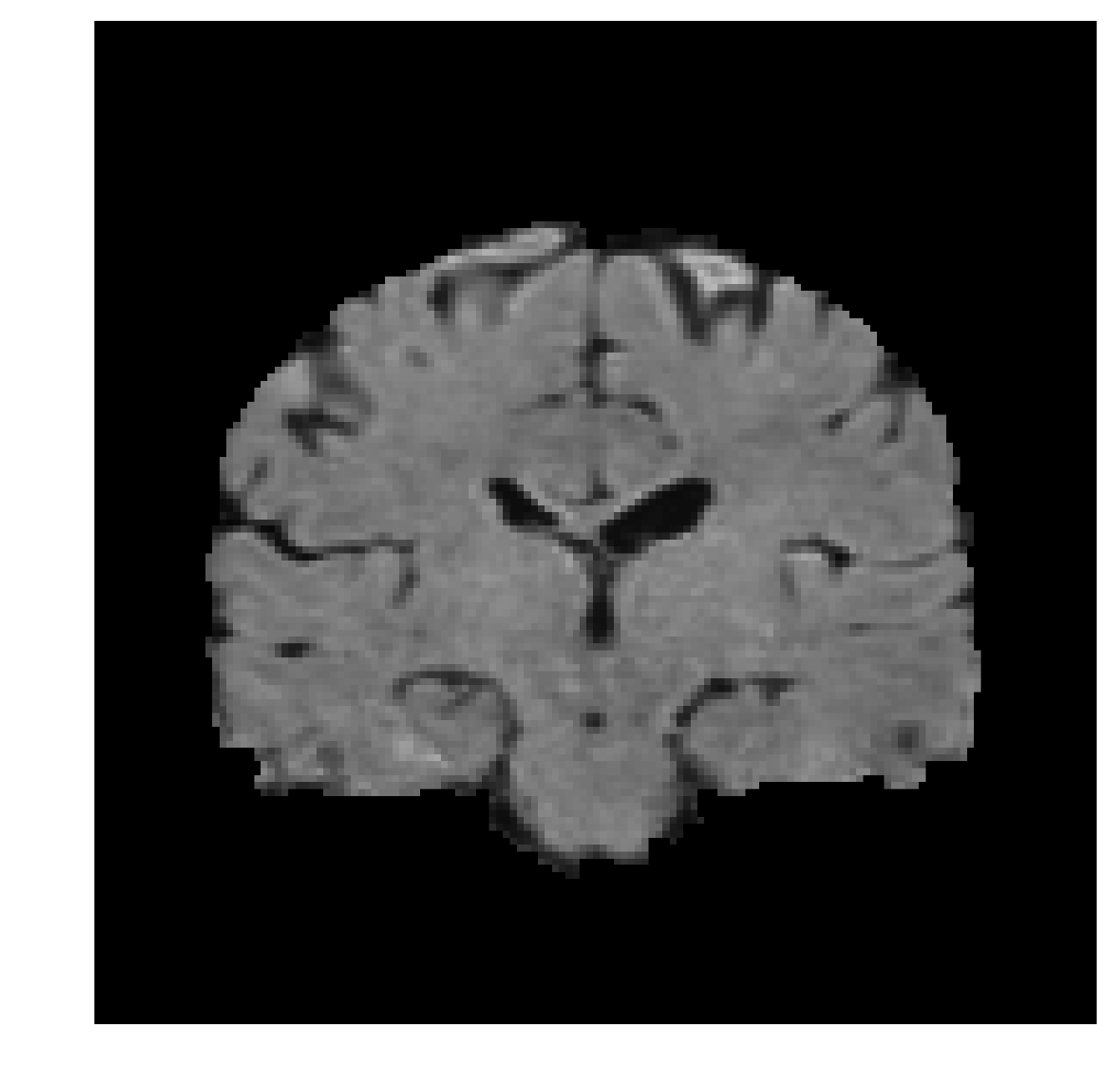}  \\
\rotatebox{90}{\hspace{0.5cm}$b = 2000 s/mm^2$}  &
\includegraphics[width=0.3\columnwidth, trim={2cm 2cm 1cm 2cm}, clip]{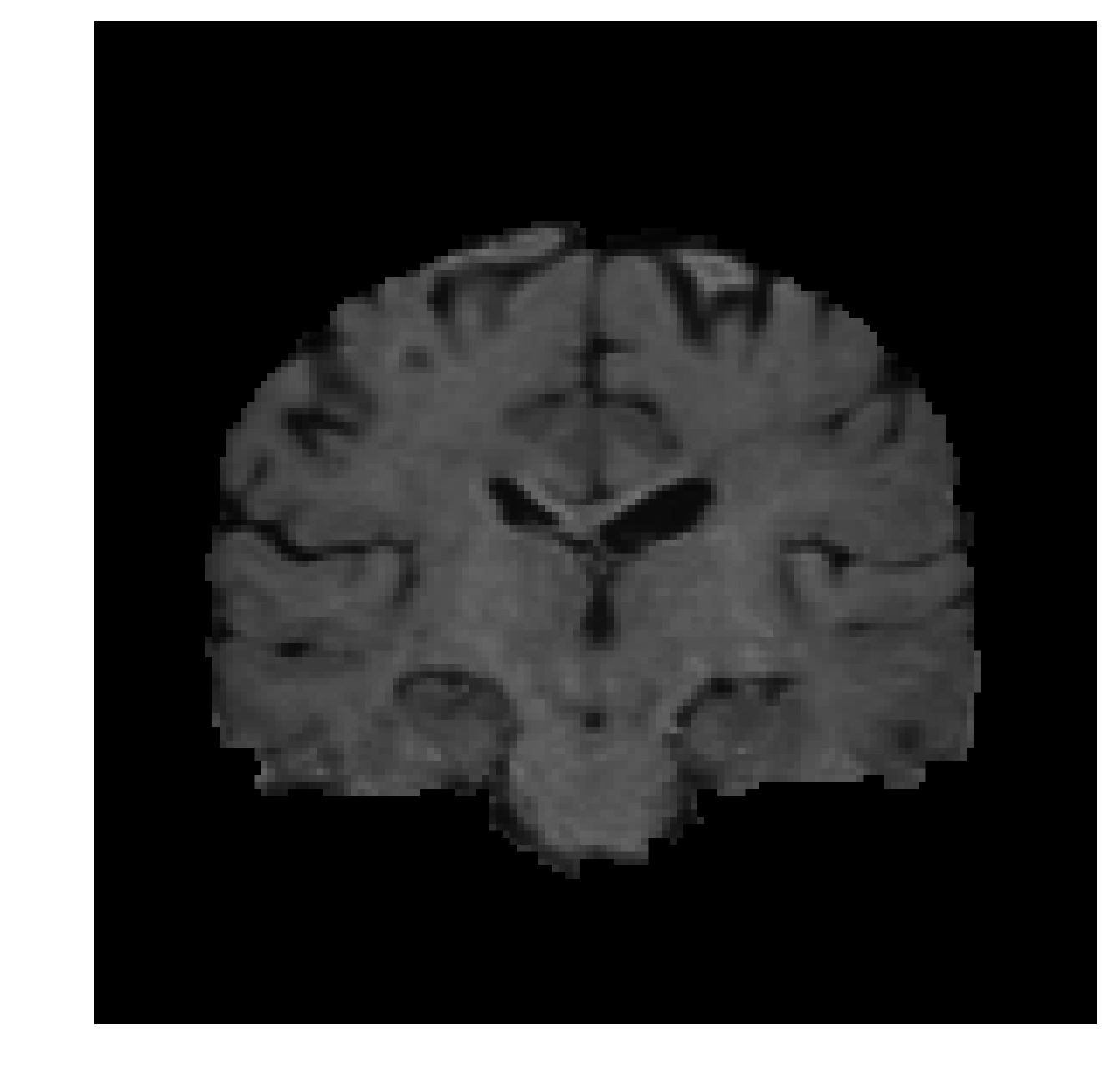}  &
\includegraphics[width=0.3\columnwidth, trim={2cm 2cm 1cm 2cm}, clip]{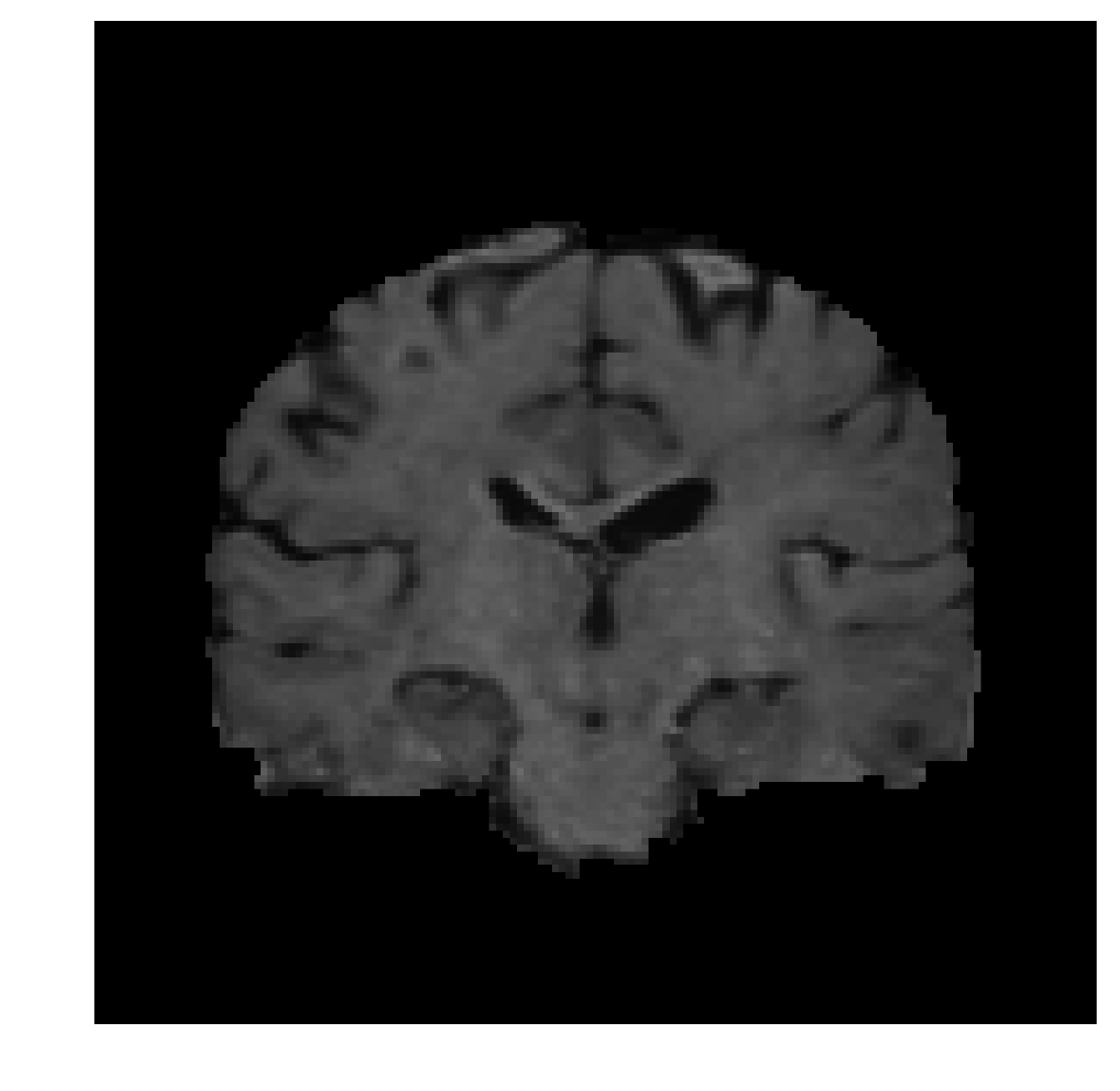}  &
\includegraphics[width=0.3\columnwidth, trim={2cm 2cm 1cm 2cm}, clip]{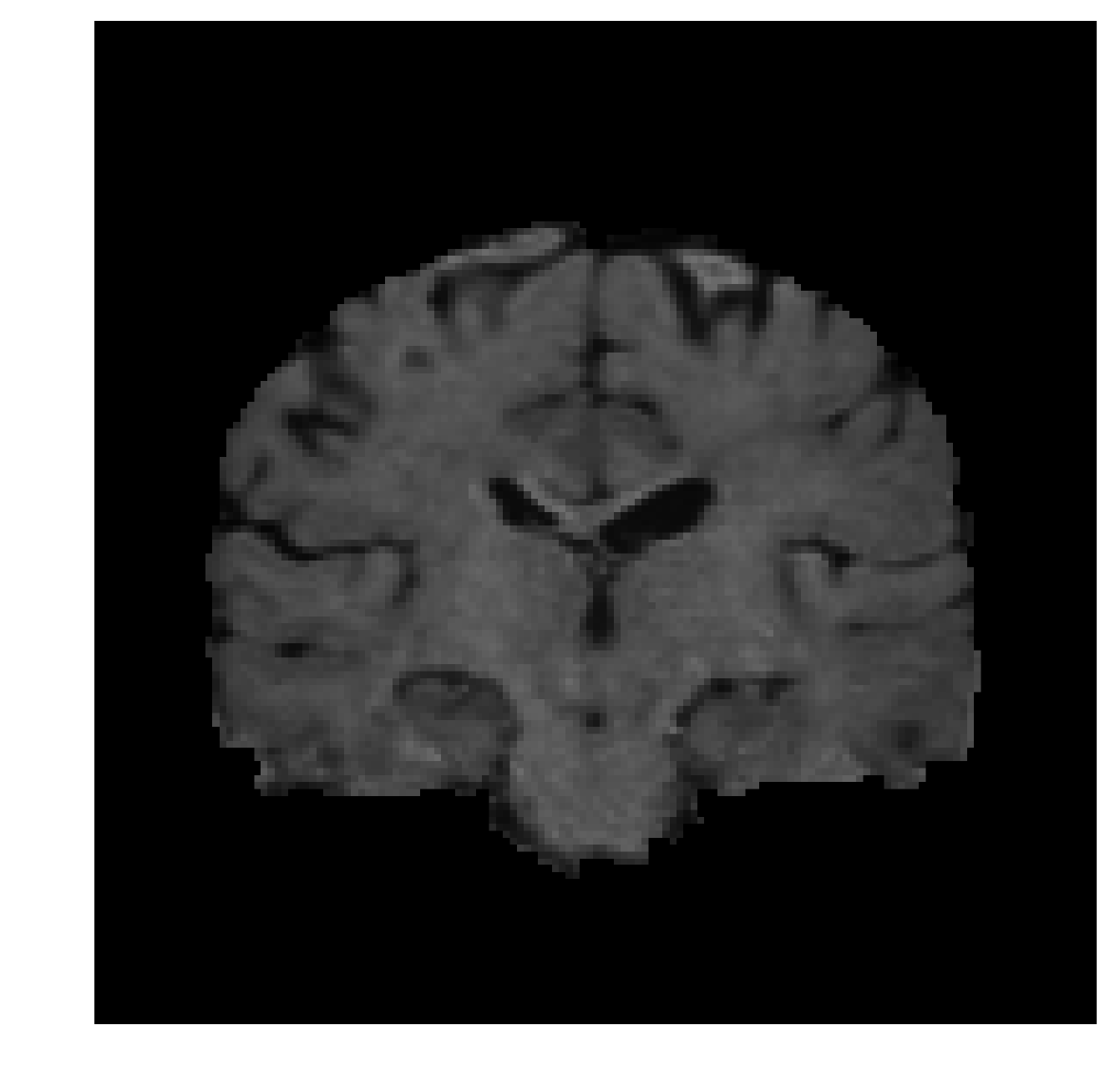}  \\
\rotatebox{90}{\hspace{0.5cm}$b = 3000 s/mm^2$}  &
\includegraphics[width=0.3\columnwidth, trim={2cm 2cm 1cm 2cm}, clip]{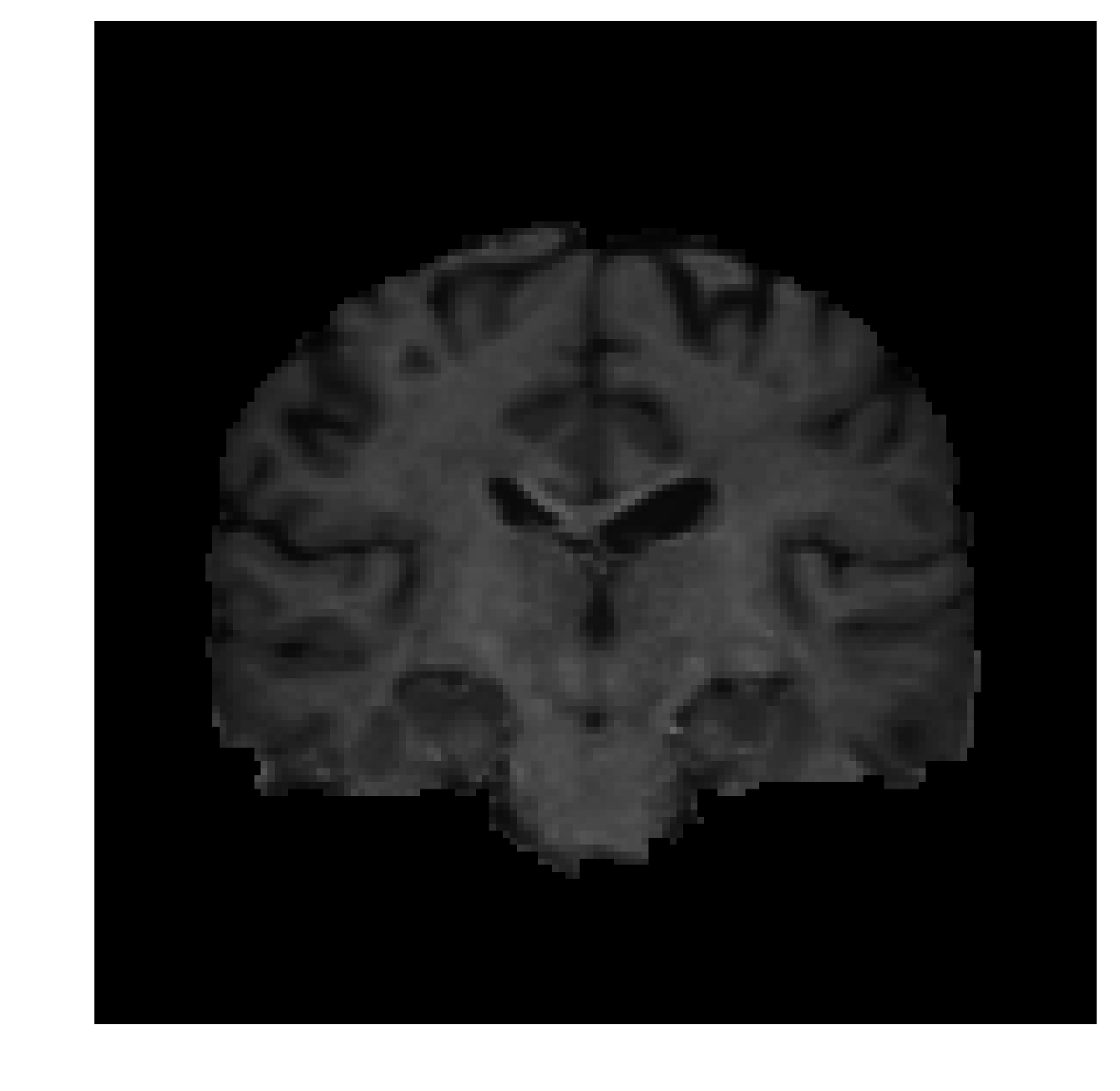}  &
\includegraphics[width=0.3\columnwidth, trim={2cm 2cm 1cm 2cm}, clip]{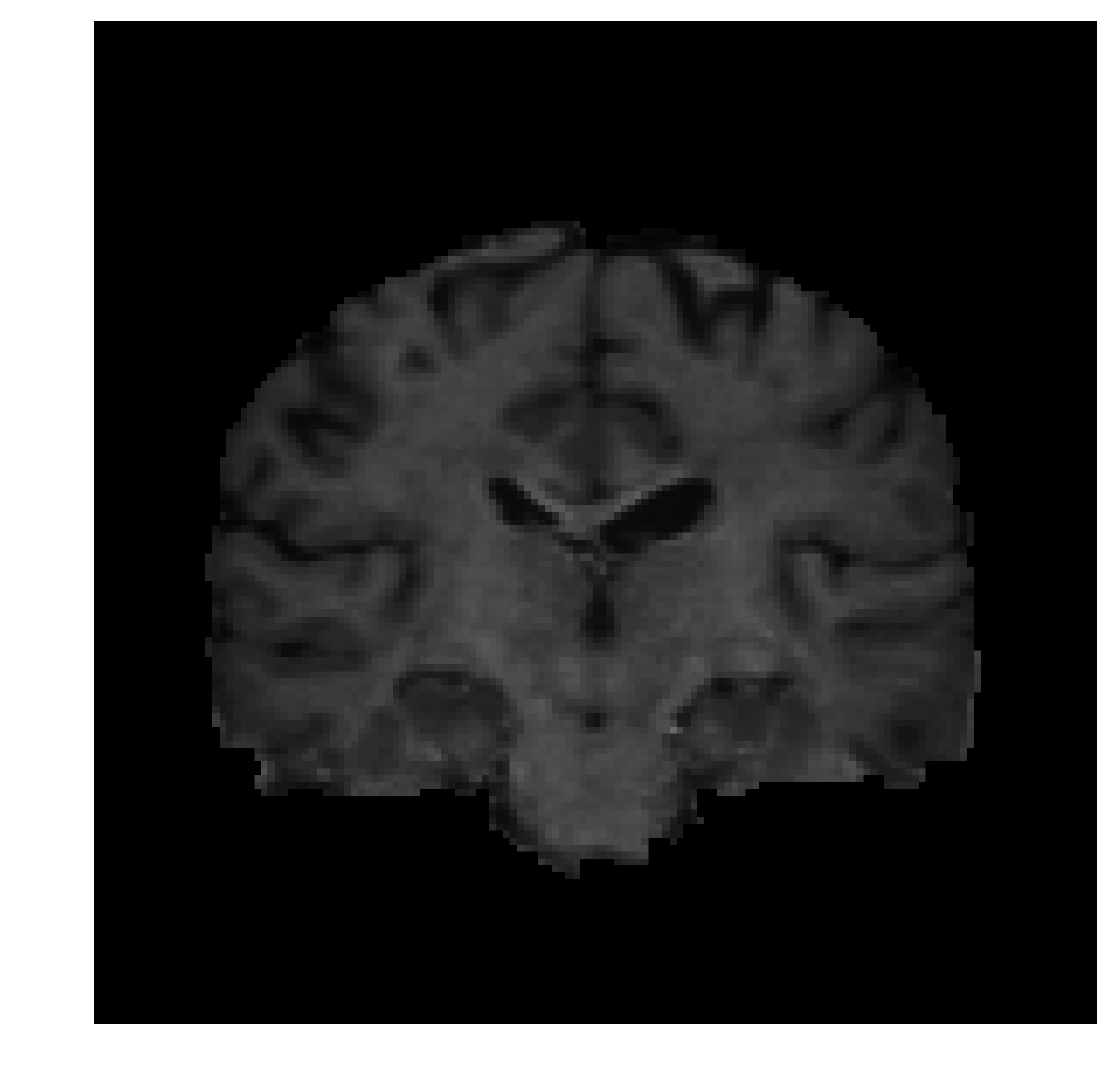}  &
\includegraphics[width=0.3\columnwidth, trim={2cm 2cm 1cm 2cm}, clip]{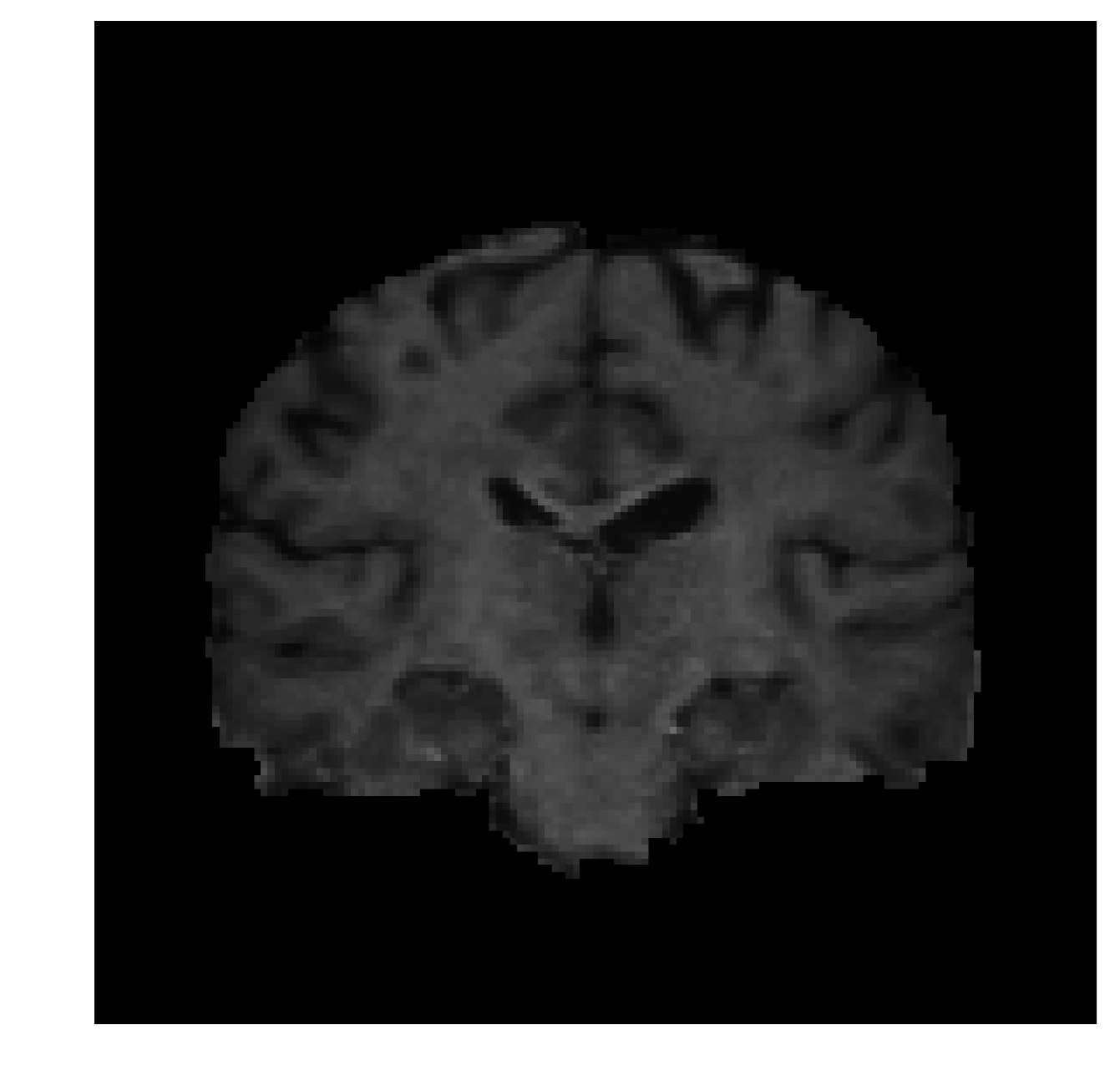}  \\
& & & \hspace{-8cm}\includegraphics[width=0.45\columnwidth, trim={0cm 0cm 0cm 11cm}, clip]{Figures/COLORBAR_VF.pdf} \\
 \end{tabular} }
\caption{Mean value of the diffusion signal for each \textit{b}-values. The mean diffusion signal is calculated using the full dataset (90 directions) or keeping only 60 and 30 directions.}
\label{fig:emean}
\end{figure}

HCP acquisition scheme consists of 270 gradients plus 18 volumes acquired at \textit{b}-value = 0 s/mm$^2$. Such acquisition is far from being considered adapt to clinical practice, which normally consists of 60 diffusion weighted images.
In order to test the ability of NODDI-SH to fit more practical acquisition schemes, we subsampled the HCP data. Figure \ref{fig:emean} shows the mean signal value of the HCP data for each \textit{b}-value, while keeping all the gradients (90 directions), 60 directions, and only 30 directions, respectively. As expected, the diffusion signal decays for increasing \textit{b}-values. In regions of fast diffusion, such as the CSF, the diffusion signal is almost zero even at \textit{b}-value = 1000 s/mm$^2$. At \textit{b}-value = 2000 s/mm$^2$ it is possible to see some contrast between white matter (high diffusion restriction) and gray matter regions, which is more pronounced at \textit{b}-value = 3000 s/mm$^2$.

At a first glance, it is impossible to distinguish between the mean value images obtained using a different number of samples. This because the mean difference in a white matter region between the mean signal values at \textit{b}-value = 1000 s/mm$^2$ derived from the 90 and the 30 samples cases is only of $1\%$. The mean signal over each shell is necessary for NODDI-SH to retrieve the volume fractions, and the robustness of the mean signal to subsampling reflects the robustness of the volume fractions estimation.

\begin{figure}[!t]
\centering{
\setlength{\tabcolsep}{1pt}    
\begin{tabular}{cccc}
%ISOTROPIC FANNING & ANISTROPIC FANNING \\
& 90 directions & 60 directions & 30 directions \\
\rotatebox{90}{\hspace{0.0cm}$b_{max} = 2000 $s/mm$^2$}  &
\includegraphics[width=0.3\columnwidth, trim={2cm 2cm 1cm 2cm}, clip]{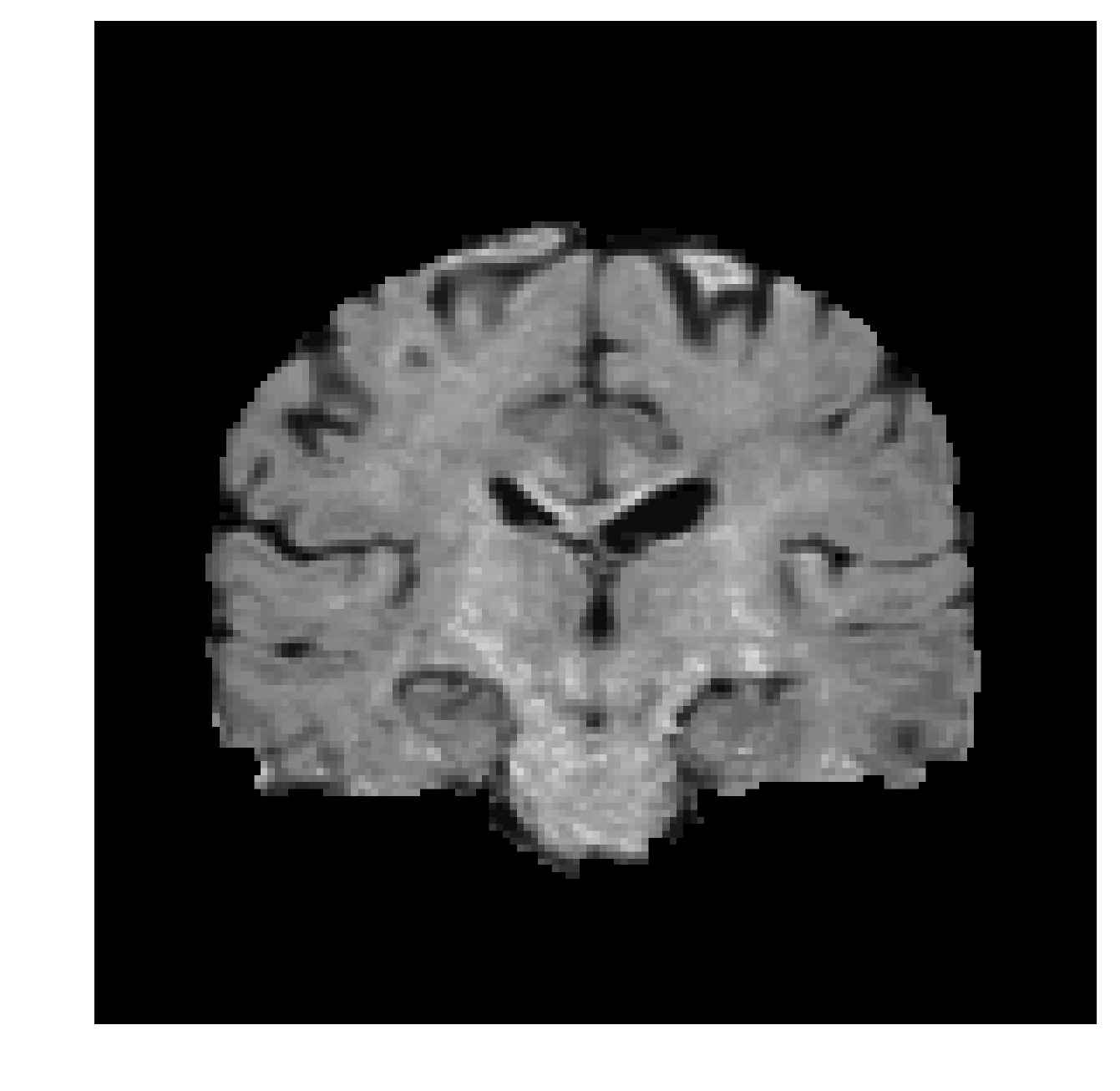}  &
\includegraphics[width=0.3\columnwidth, trim={2cm 2cm 1cm 2cm}, clip]{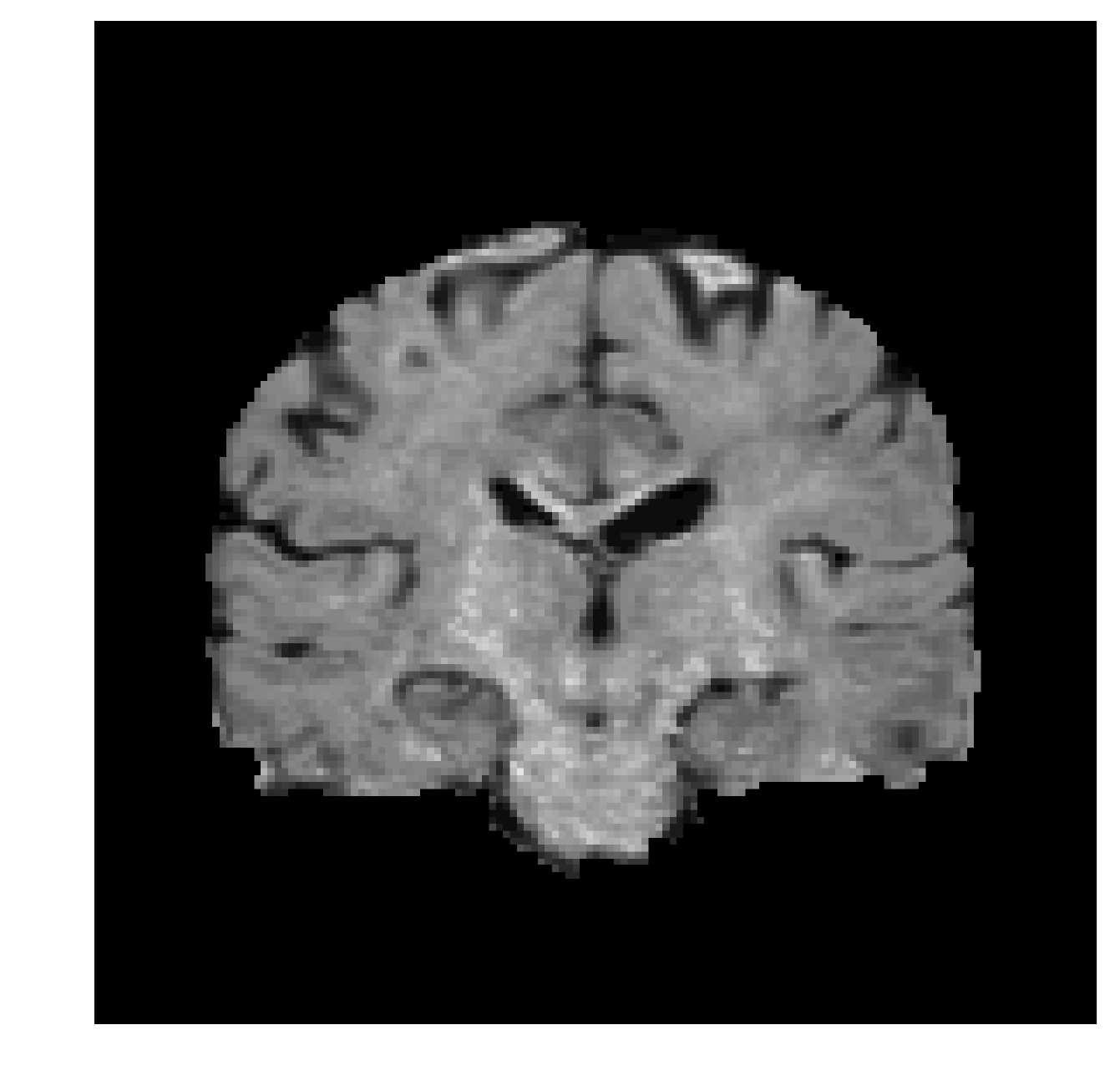}  &
\includegraphics[width=0.3\columnwidth, trim={2cm 2cm 1cm 2cm}, clip]{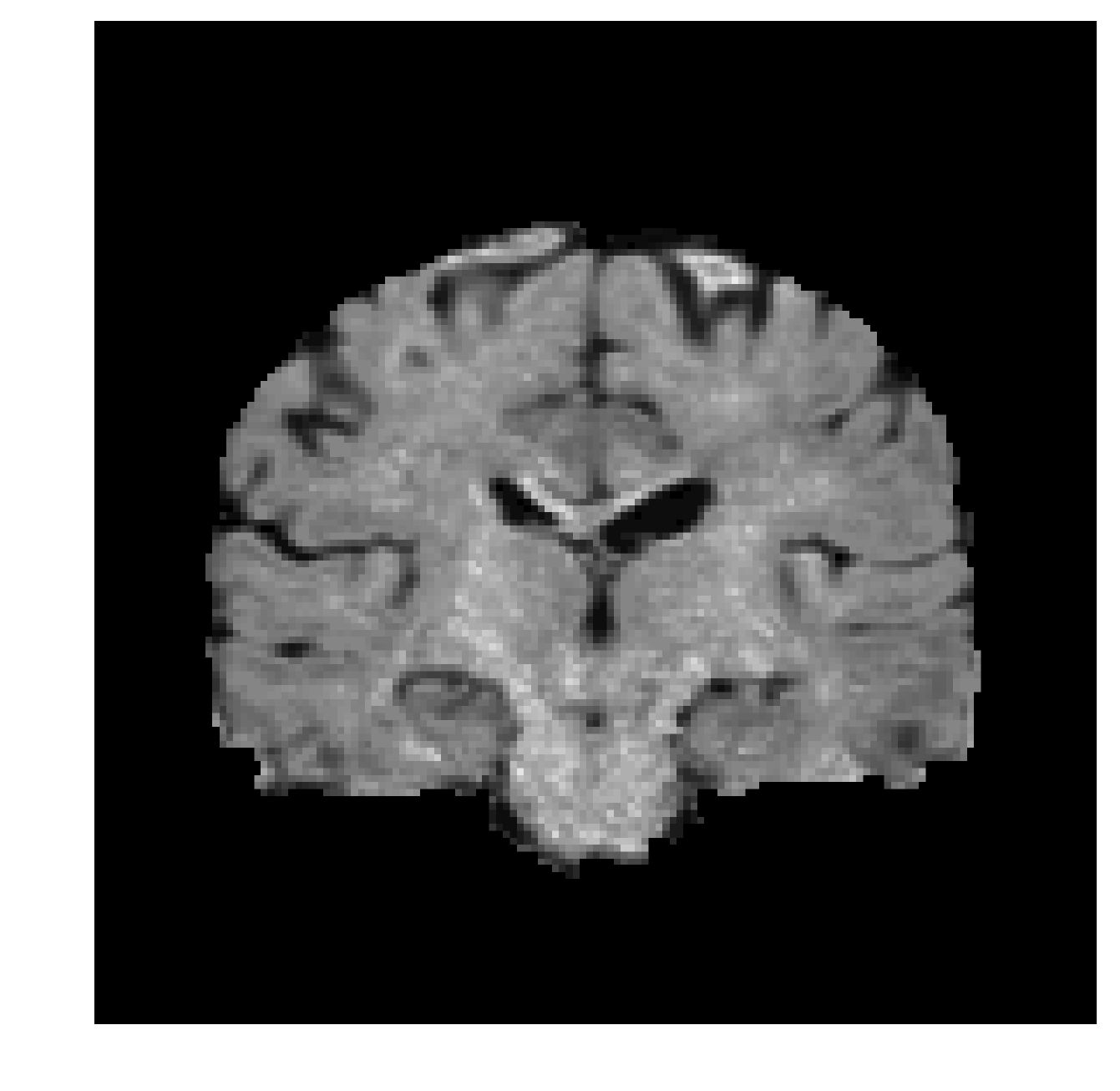}  \\
\rotatebox{90}{\hspace{0.0cm}$b_{max} = 3000 $s/mm$^2$}  &
\includegraphics[width=0.3\columnwidth, trim={2cm 2cm 1cm 2cm}, clip]{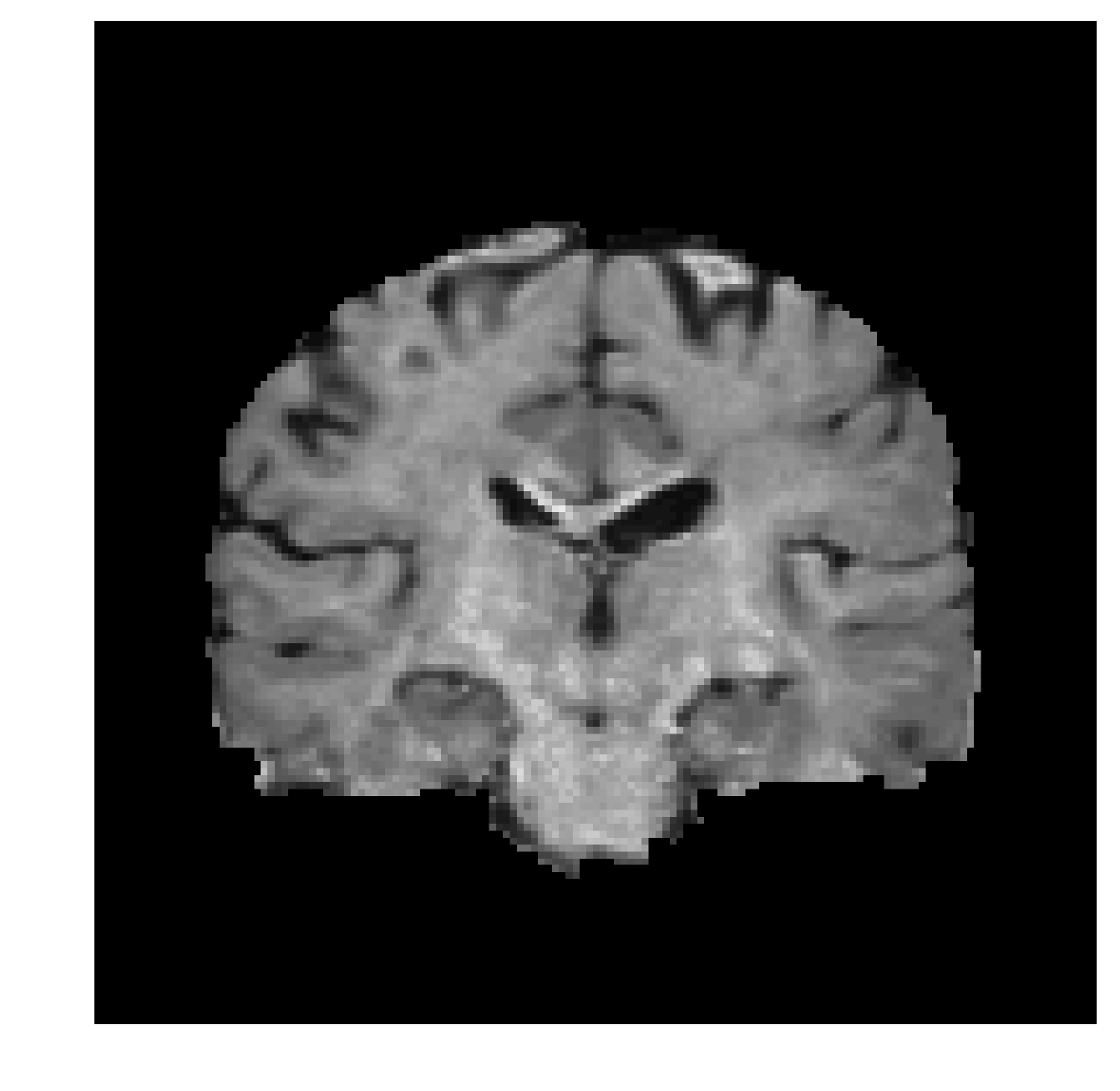}  &
\includegraphics[width=0.3\columnwidth, trim={2cm 2cm 1cm 2cm}, clip]{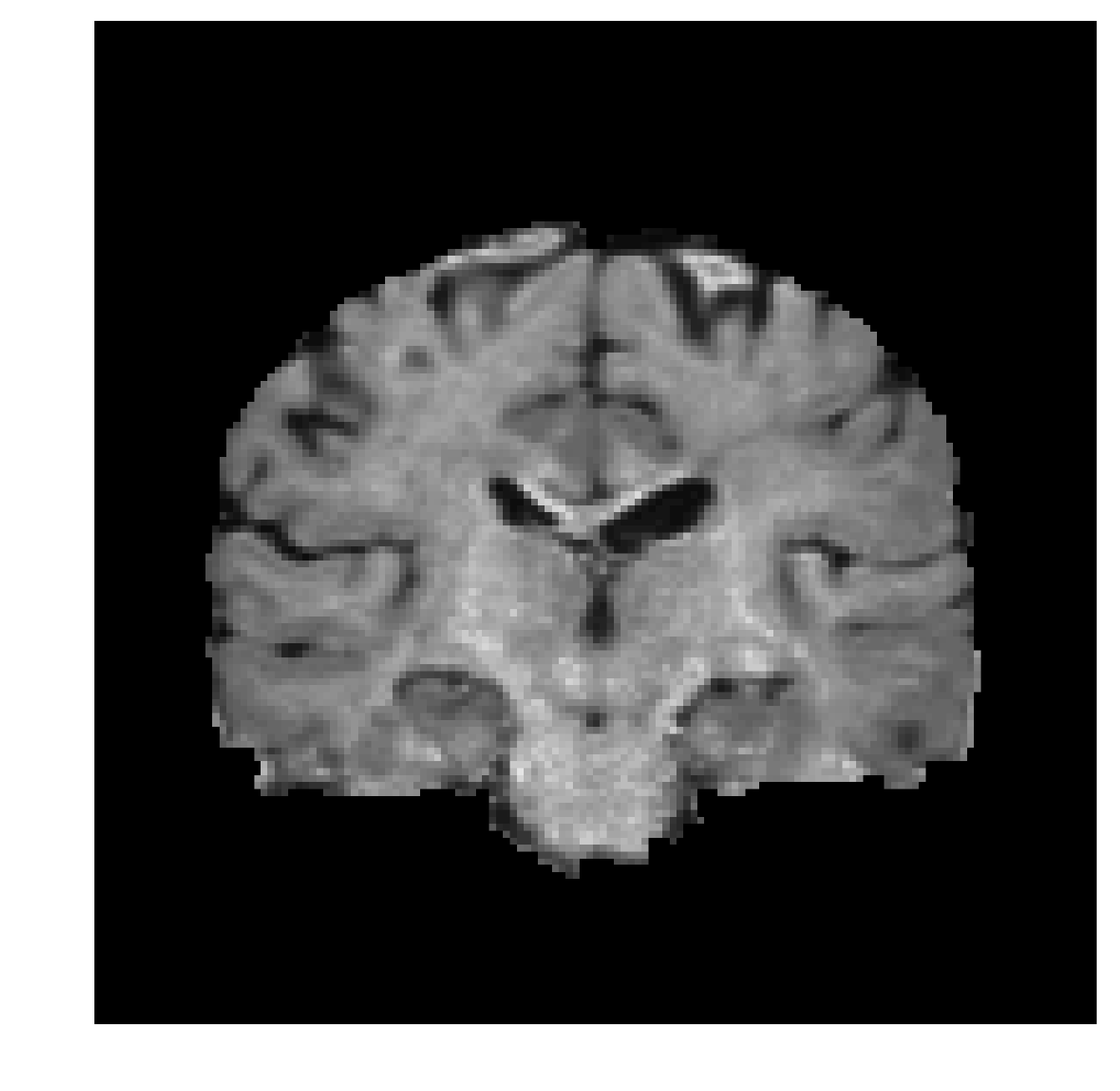}  &
\includegraphics[width=0.3\columnwidth, trim={2cm 2cm 1cm 2cm}, clip]{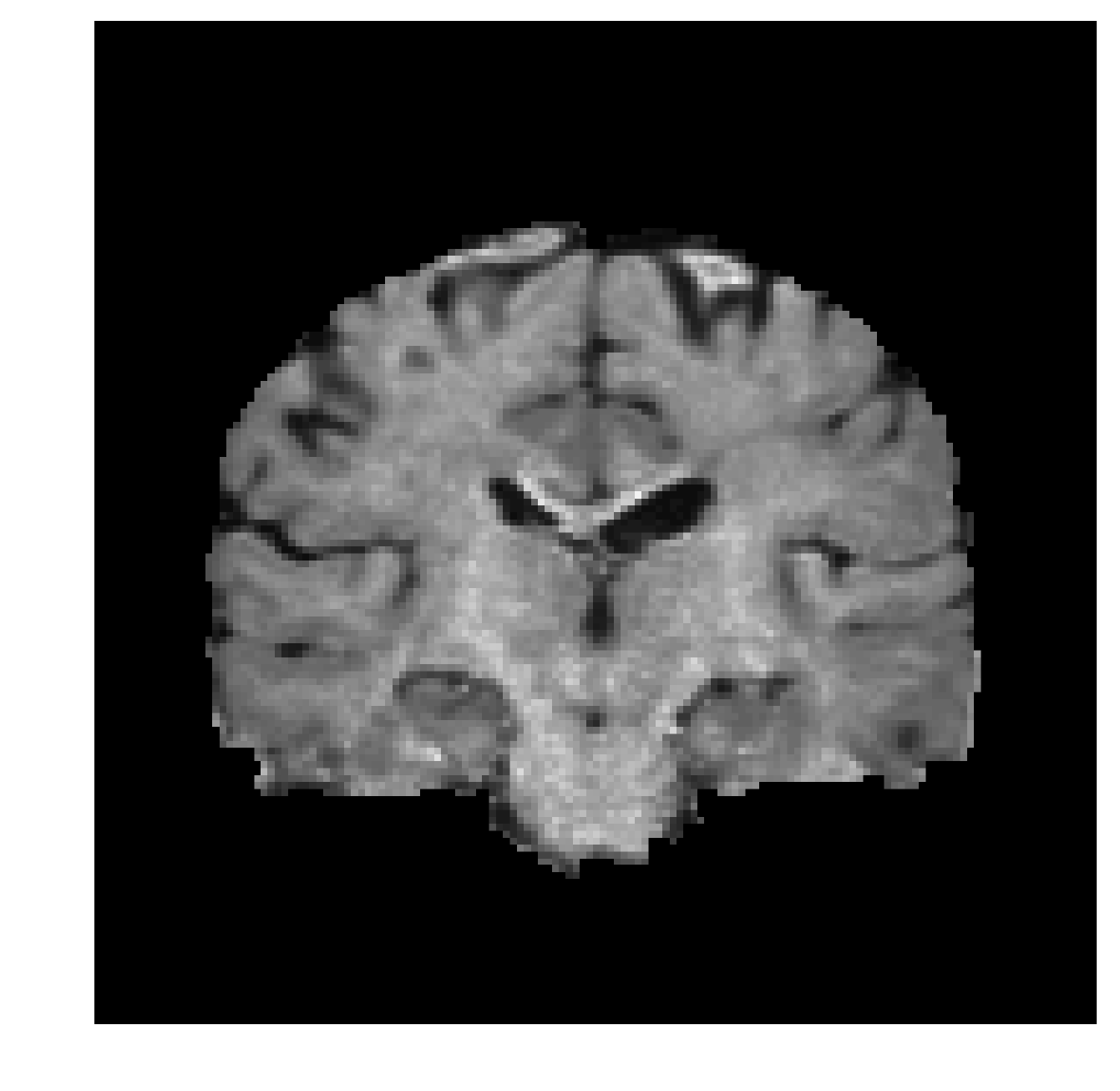}  \\
& & & \hspace{-8cm}\includegraphics[width=0.45\columnwidth, trim={0cm 0cm 0cm 11cm}, clip]{Figures/COLORBAR_VF.pdf} \\
 \end{tabular} }
\caption{Intracellular volume fraction calculated using NODDI-SH considering only a subsample of the HCP gradients for the fitting.}
\label{fig:vic_all}
\end{figure}

Figure \ref{fig:vic_all} shows the variation of the intracellular volume fraction calculated using NODDI-SH when changing the number of directions per shell and the number of shells (\textit{b}-value = [1000, 2000, 3000] s/mm$^2$ or \textit{b}-value = [1000,2000] s/mm$^2$ ). As expected, the intracellular volume fraction does not change significantly with the number of samples when all the shells are kept ($b_{max}$=3000 s/mm$^2$). On the contrary, excluding all the samples at \textit{b}-value = 3000 s/mm$^2$ affects the estimation of $\nu_{ic}$, in particular if we consider only 30 samples per shell. The $\nu_{ic}$ map in this case results more noisy. Nevertheless, the overall quality of the image does not degrade too much with respect the fully sampled image. This result is extremely promising, considering that we move from a 270 directions acquisition with $b_{max}=3000$ s/mm$^2$ to a 60 directions acquisition with $b_{max}=2000$ s/mm$^2$ (more than 76$\%$ samples less).

\begin{figure}[!t]
\centering{
\setlength{\tabcolsep}{1pt}    
\begin{tabular}{cccc}
%ISOTROPIC FANNING & ANISTROPIC FANNING \\
& 90 directions & 60 directions & 30 directions \\
\rotatebox{90}{\hspace{0.0cm}$b_{max} = 2000 $s/mm$^2$}  &
\includegraphics[width=0.3\columnwidth, trim={2cm 2cm 1cm 2cm}, clip]{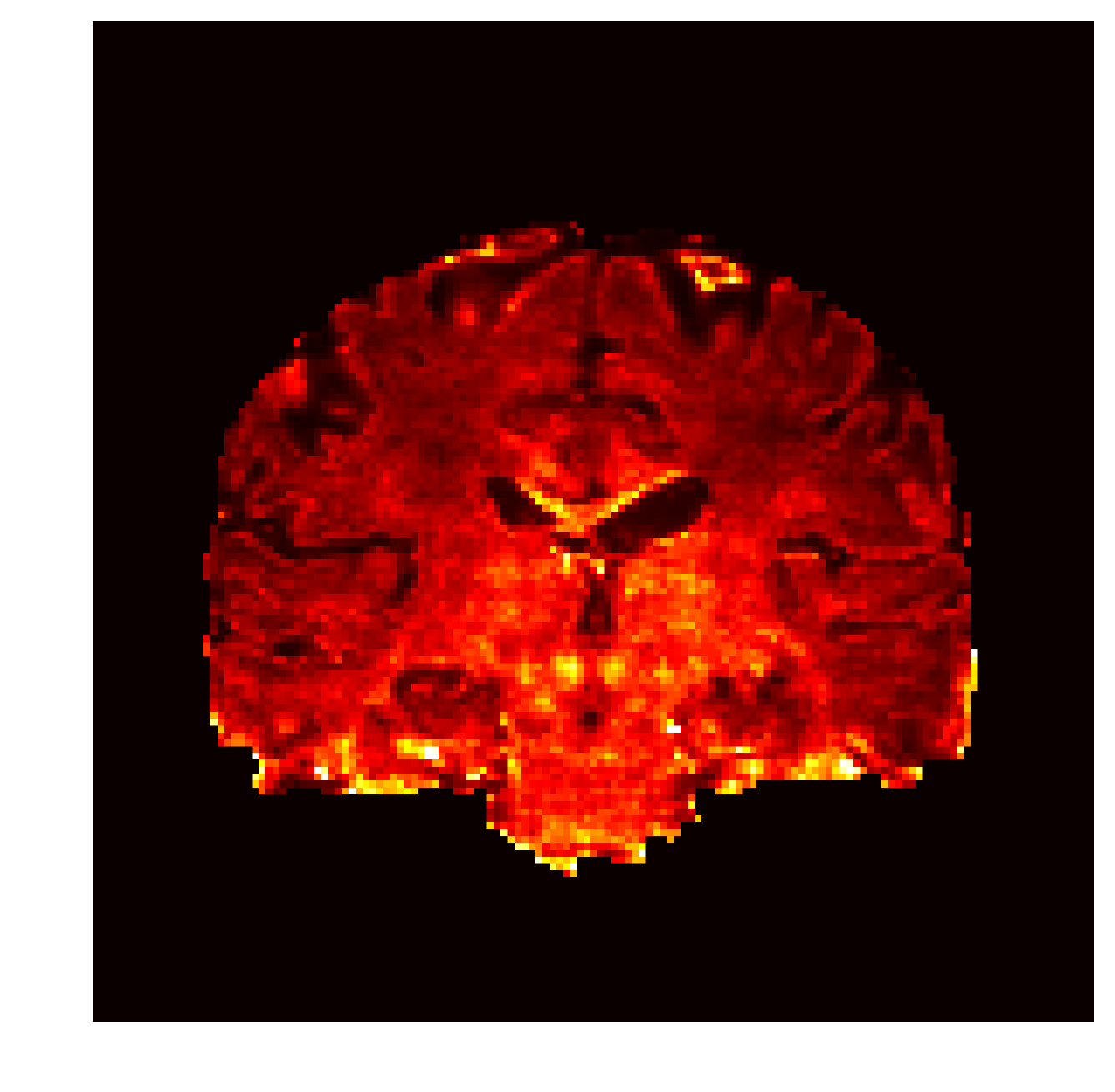}  &
\includegraphics[width=0.3\columnwidth, trim={2cm 2cm 1cm 2cm}, clip]{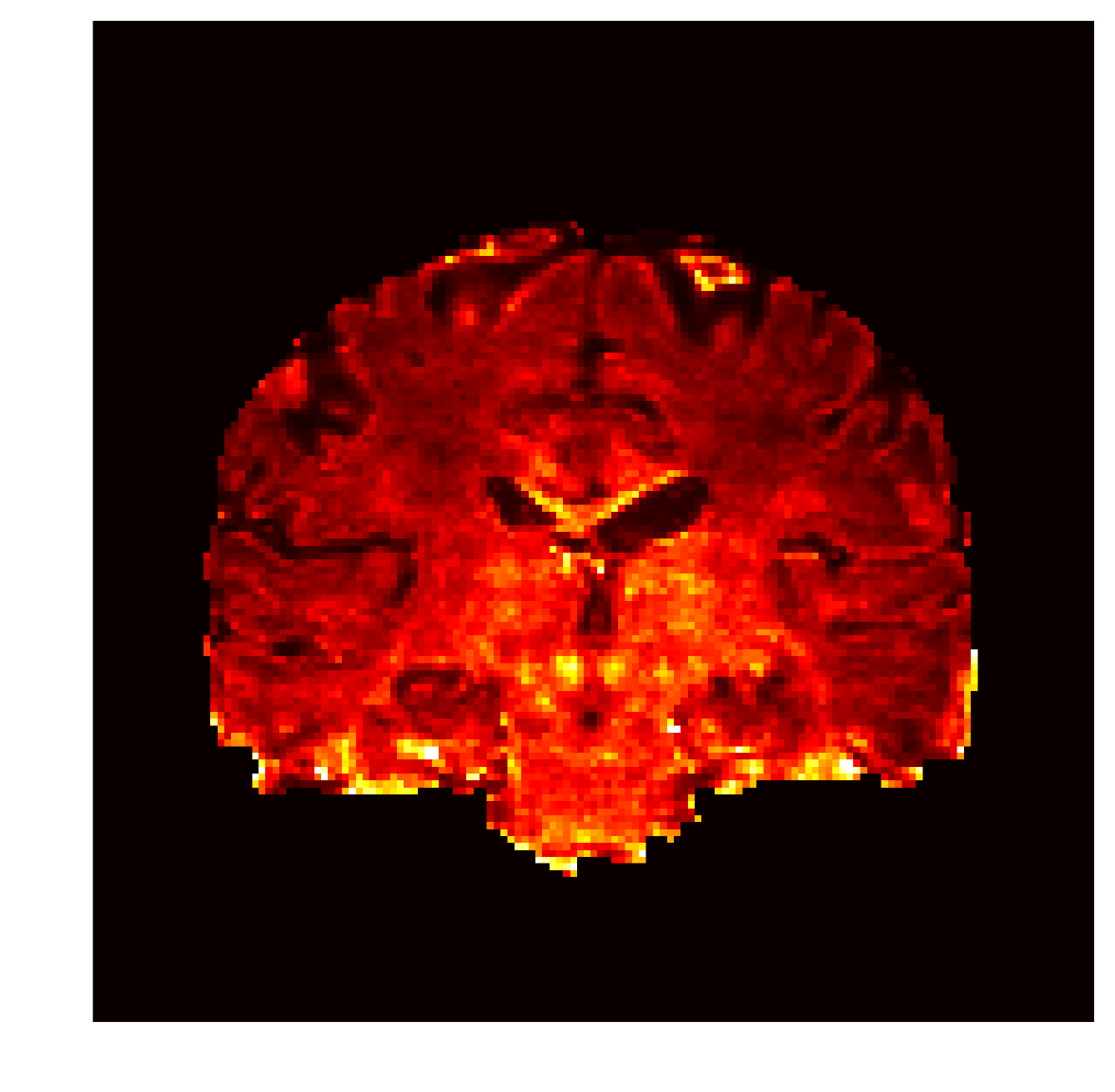}  &
\includegraphics[width=0.3\columnwidth, trim={2cm 2cm 1cm 2cm}, clip]{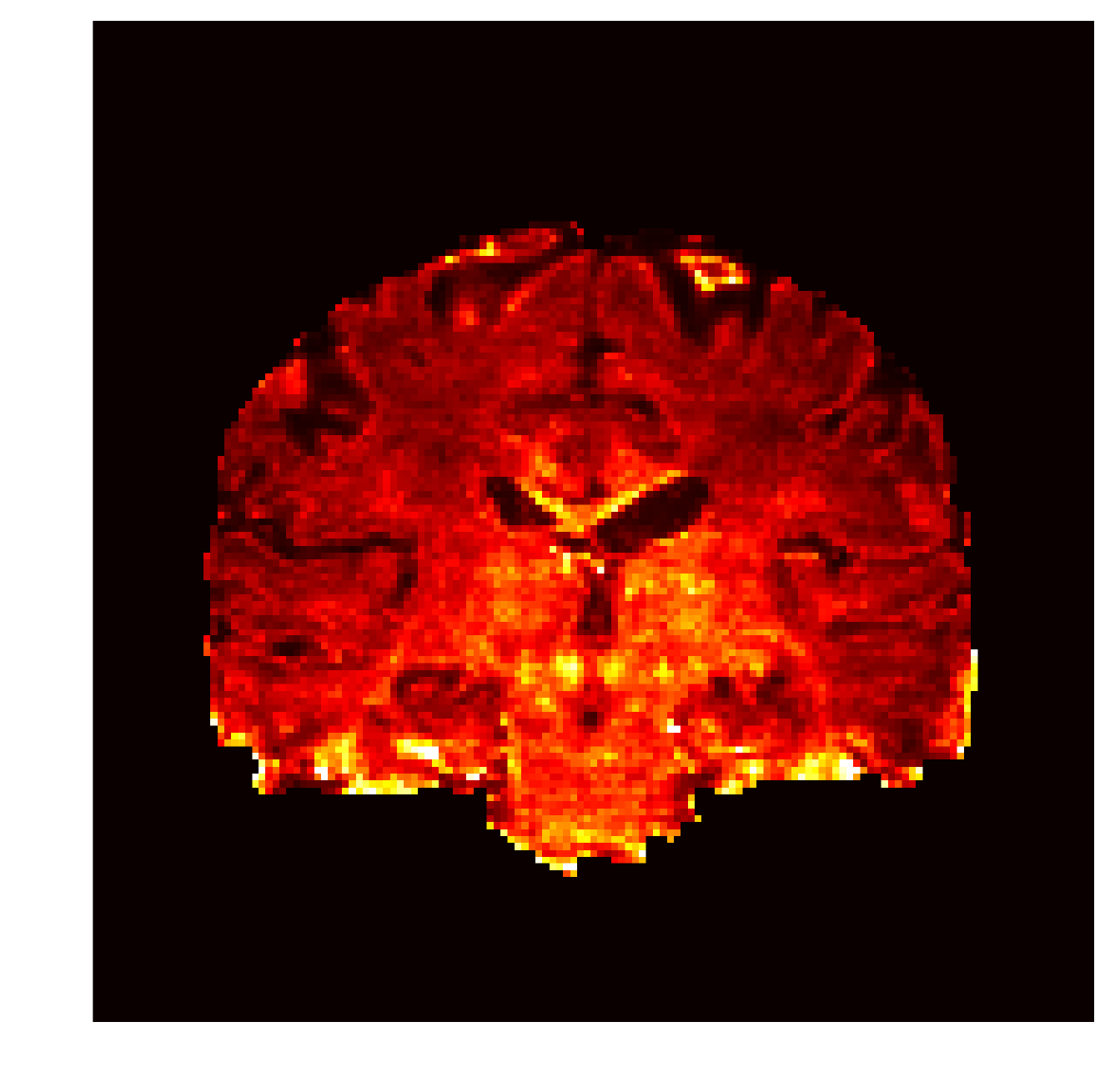}  \\
\rotatebox{90}{\hspace{0.0cm}$b_{max} = 3000 $s/mm$^2$}  &
\includegraphics[width=0.3\columnwidth, trim={2cm 2cm 1cm 2cm}, clip]{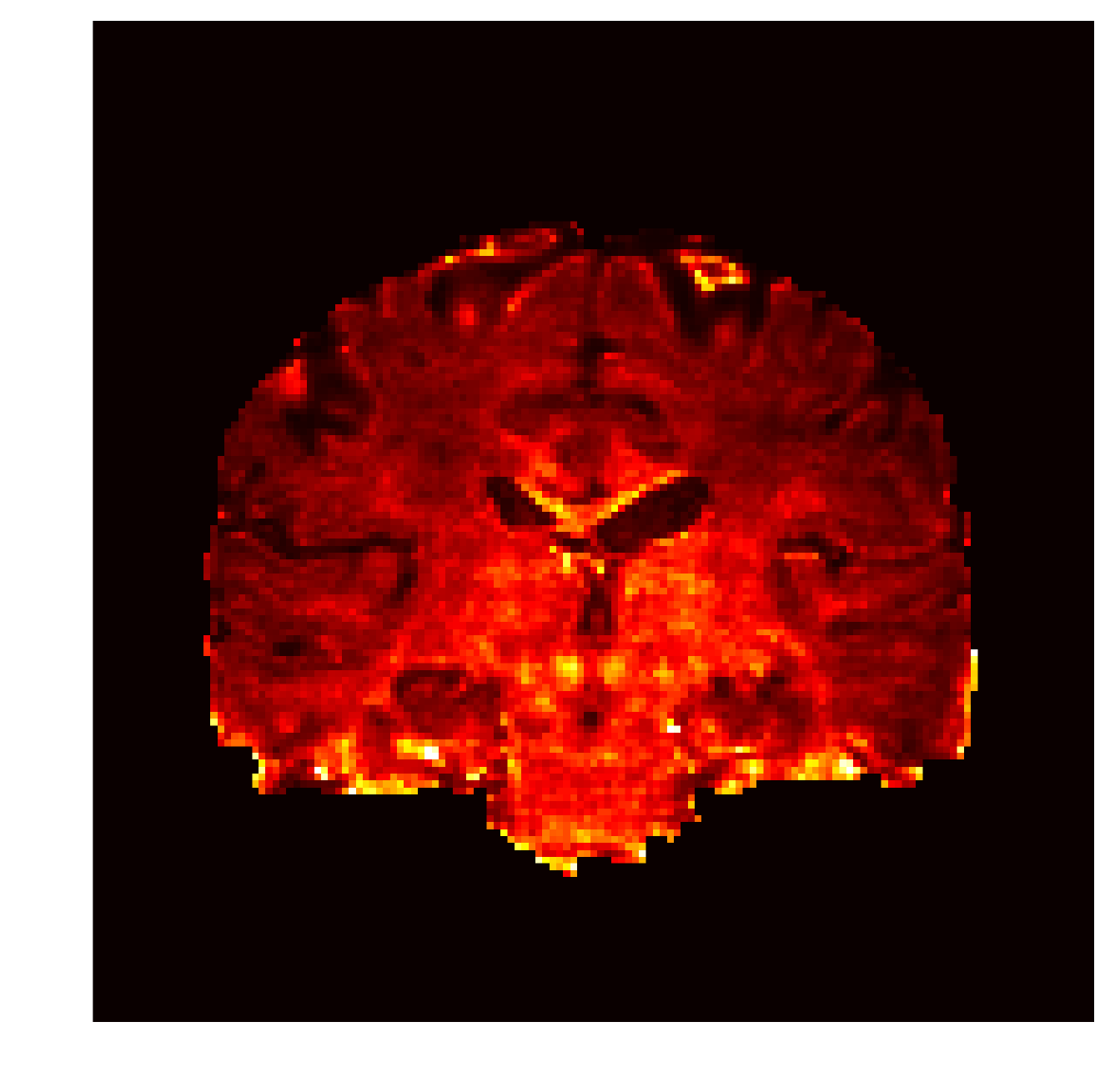}  &
\includegraphics[width=0.3\columnwidth, trim={2cm 2cm 1cm 2cm}, clip]{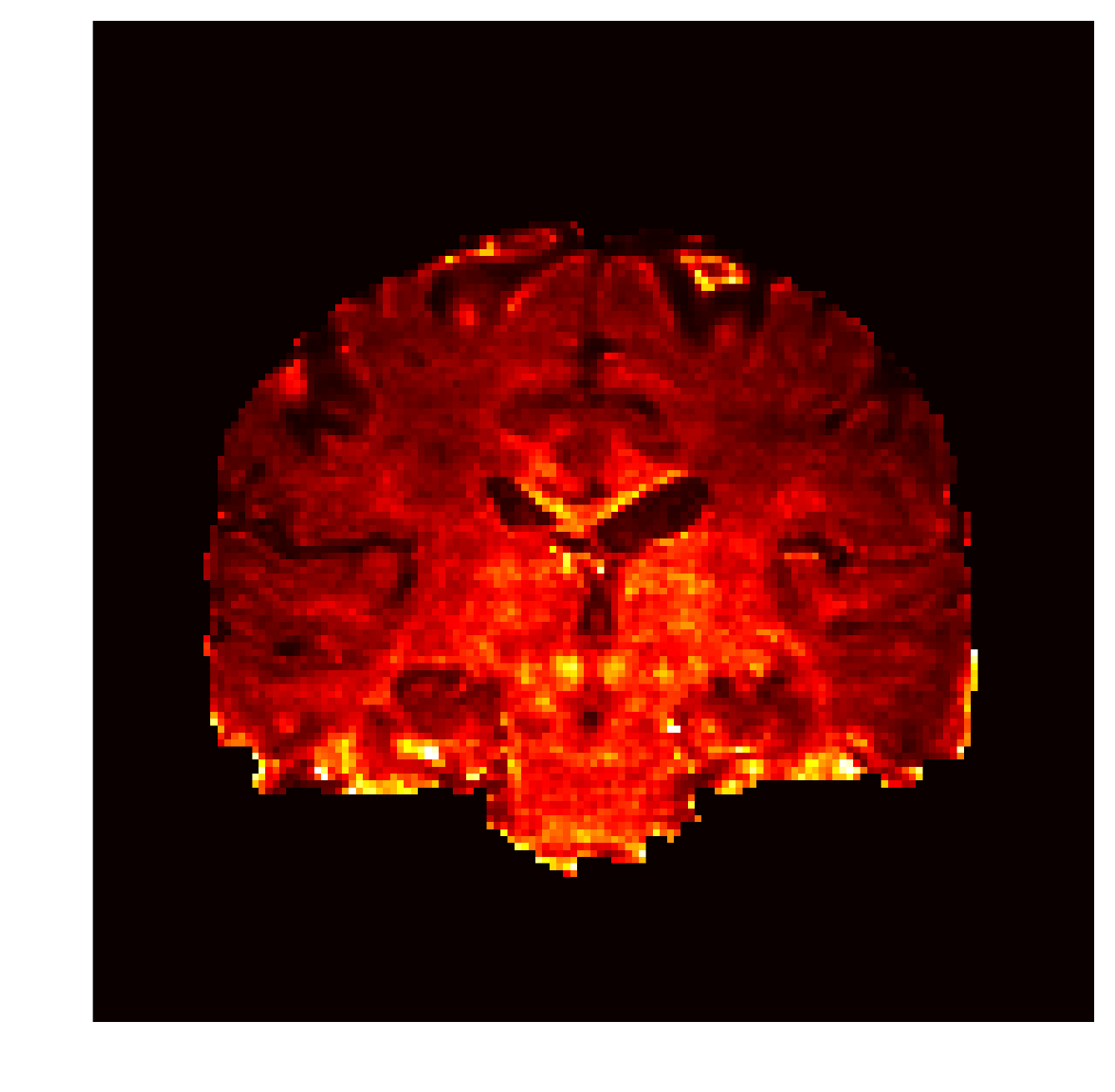}  &
\includegraphics[width=0.3\columnwidth, trim={2cm 2cm 1cm 2cm}, clip]{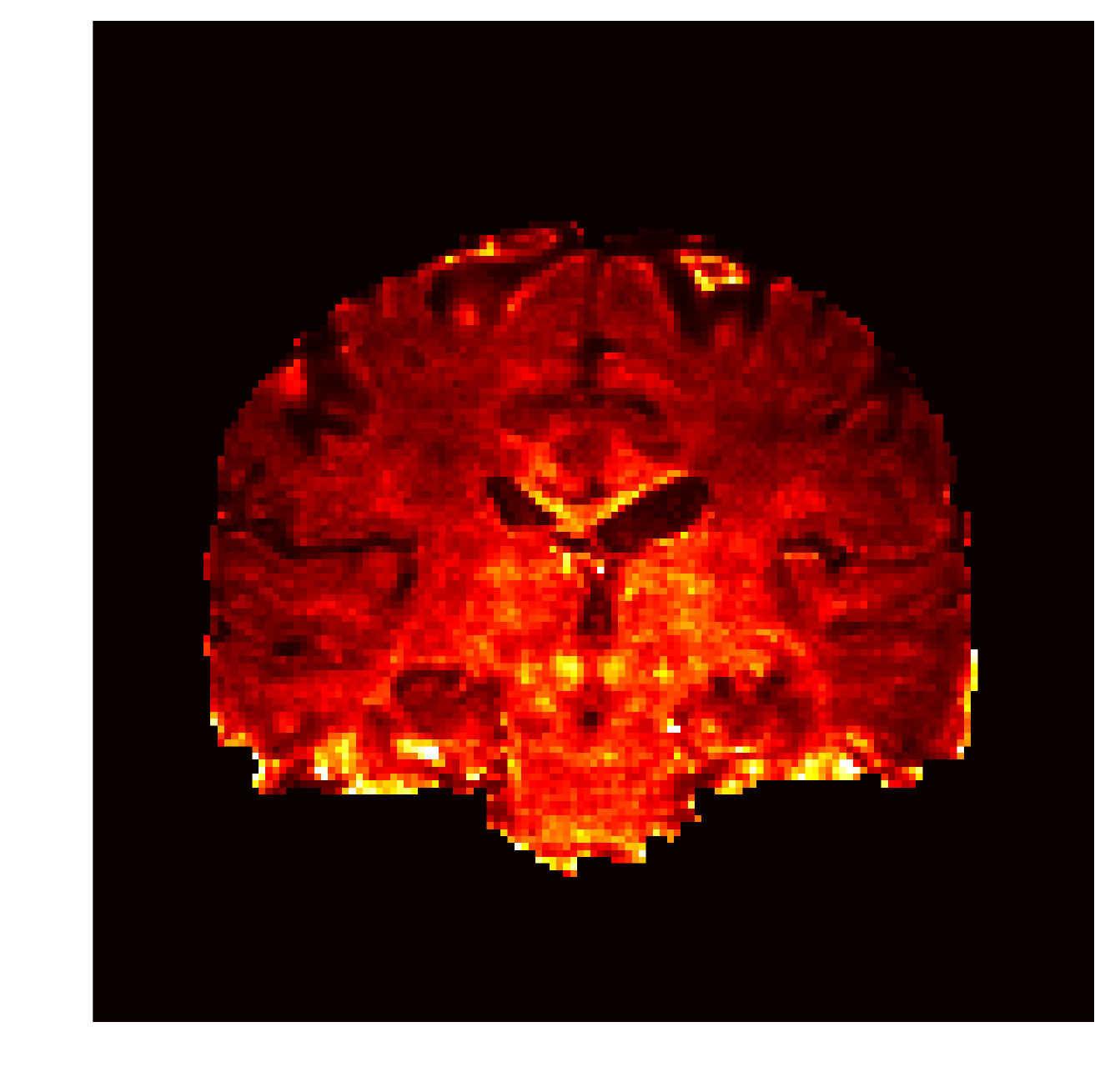}  \\
& & &\hspace{-8cm}\includegraphics[width=0.45\columnwidth, trim={0cm 0cm 0cm 11cm}, clip]{Figures/COLORBAR_MSE.pdf} \\
 \end{tabular} }
\caption{MSEs calculated using NODDI-SH considering only a subsample of the HCP gradients for the fitting.}
\label{fig:mse_all}
\end{figure}

In order to quantify the error that is introduced while decreasing the number of samples, we calculated the MSE between the reconstructed signal obtained after fitting NODDI-SH to the subsampled dataset and the full diffusion signal dataset. It is worth to mention that calculating the MSE in such a way clearly advantages the NODDI-SH model which fits the 90 directions, $b_{max}=3000$s/mm$^2$ data because the MSE is evaluated on the same points used for the fitting. On the contrary, when only 30 directions with $b_{max}=2000$s/mm$^2$ are used for the fitting, the NODDI-SH model requires to blindly estimate 210 points in order to reconstruct the complete diffusion signal. 
Figure \ref{fig:mse_all} shows the MSE for the reconstructed signal for the different sampling schemes. As for the $\nu_{ic}$ maps, it is challenging to perceive any difference. 

\begin{figure}[!t]
\centering{
\setlength{\tabcolsep}{1pt}    
\begin{tabular}{c}
%ISOTROPIC FANNING & ANISTROPIC FANNING \\
\includegraphics[width=0.45\columnwidth, trim={0cm 0cm 0cm 0cm}, clip]{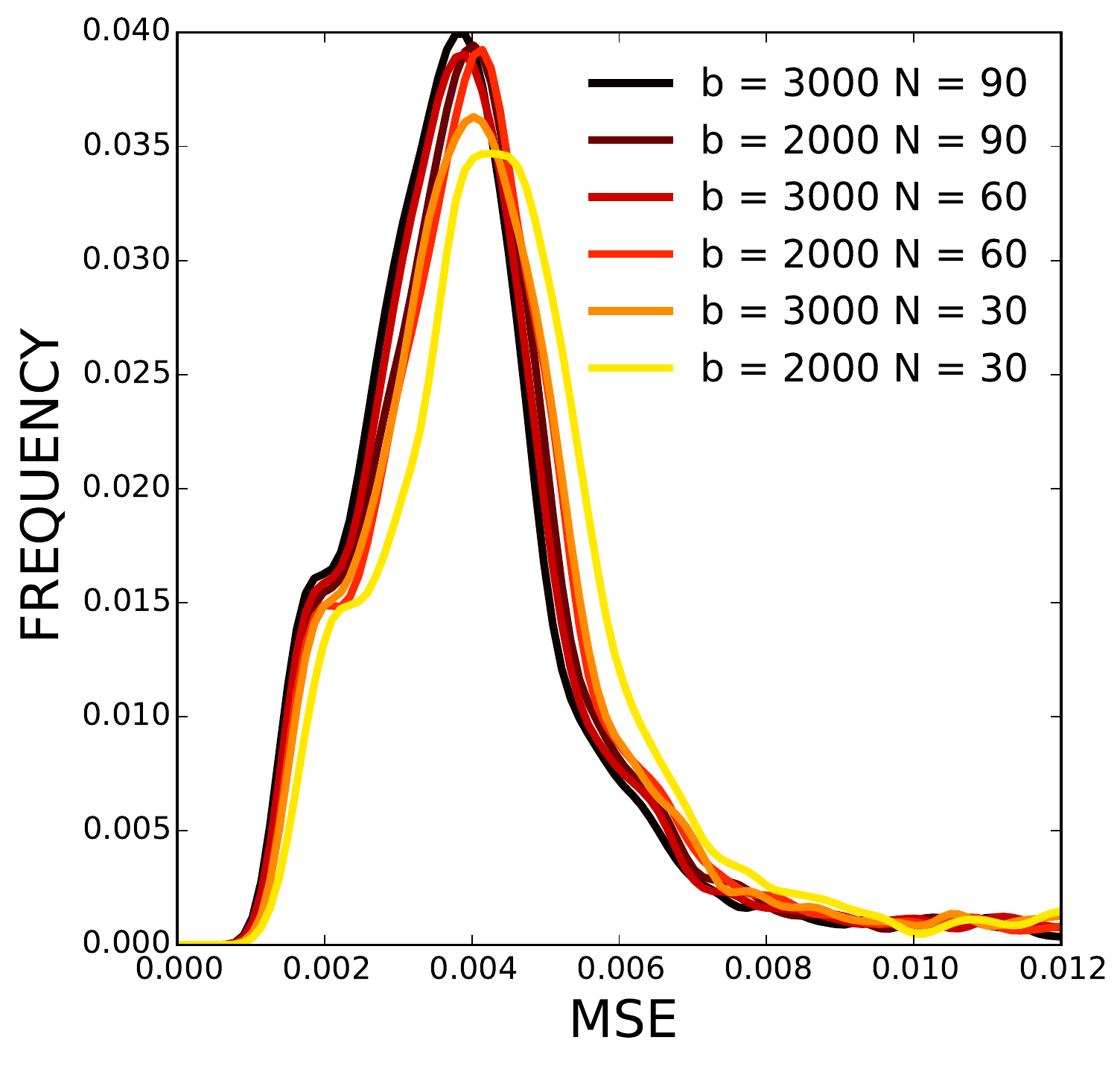} \\
 \end{tabular} }
\caption{Normalized histogram of the MSE for the different sampling schemes calculated only in voxel with $\nu_{ic} > 0.6 $.}
\label{fig:mse_his}
\end{figure}

In order to better highlight the changes, we calculated the normalized histogram of the MSE image in voxels presenting high level of intracellular volume fraction ($\nu_{ic} > 0.6 $). Figure \ref{fig:mse_his} shows the histograms of the MSEs obtained from the different sampling schemes. As it possible to see, 90 directions, $b_{max}=3000$s/mm$^2$ presents the highest number of voxels with low MSE, followed by the 60 directions, $b_{max}=3000$s/mm$^2$, and the 90 directions, $b_{max}=2000$s/mm$^2$. The worst schemes are the ones using only 30 directions per shell. The high MSE of the 30 directions scheme is probably due to the low angular resolution which makes it less suited to capture crossing fibers signal. 

\subsection{Computational time}
\label{sec:times}
NODDI-SH volume fractions estimation and fODF are substantially similar to NODDI and CSD results, respectively. However, to obtain the same information provided by NODDI-SH it is necessary to run both NODDI and CSD on the same data. Table \ref{tab:times} shows the computational time for CSD, FORECAST, NODDI, and NODDI-SH obtained on the 2970 voxels of the simulated crossing dataset. The computational time was computed using a single core on a laptop equipped with an Intel\textregistered Core\texttrademark i7-6500 CPU @2.50GH.  DIPY CSD implementation is extremely fast, followed by NODDI-SH which is six times slower. NODDI is the slowest model ($\sim$1.5 seconds per voxel). FORECAST is the slowest model for what concerns fODF estimation.
Compared to the default NODDI toolbox, the calculation of the volume fractions for NODDI-SH is 237 times faster. 

It is worth mentioning that using AMICO \cite{daducci} it is possible to fit the NODDI model extremely fast, less than 10 minutes for a full brain, according to the original paper \cite{daducci}. However, AMICO is not a model in itself but a convex optimization framework and does not solve the theoretical limitations of NODDI in modeling only isotropic fanning fibers. However, if the goal is only to estimate the volume fractions and not the fODF, only 1.30s are necessary for the NODDI-SH to fit 2970 voxels. This makes NODDI-SH even faster than NODDI implemented using AMICO. A complete comparison of the performance of the NODDI-SH with respect to AMICO is out the scope of this paper and will be the aim of future research.
\begin{table}[!t]
\center
\begin{tabular}{c|c|c}
& 2970 voxels (s) & seconds per voxel \\
\hline
CSD & 3.13  & 0.001\\
FORECAST & 61.05  & 0.021\\
NODDI & 4419.45  & 1.488\\
NODDI-SH & 18.63  & 0.007\\
NODDI-SH (no fODF) & 1.30 & 0.0004\\
\hline
\end{tabular}
\caption{Computational times of CSD, FORECAST, NODDI, and NODDI-SH.} 
\label{tab:times}
\end{table}
\section{Discussion}
\label{sec:dis}
NODDI-SH embeds both microstructural parameters estimation and fODF reconstruction in the same mathematical framework. We demonstrated that NODDI-SH was able to perform on par with NODDI for what concerns the volume fraction estimation, and provided comparable performance with respect to CSD for the fODF estimation. Considering CSD and NODDI individually, it is impossible to obtain both the volume fractions and the fODF at the same time as in the case of NODDI-SH. The possibility to obtain all these information using a single model can greatly simplify the processing pipeline for diffusion MRI data. Another important insight which can be derived from NODDI-SH is that using the SMT, the estimation of the volume fractions depends only on the mean value of the signal. Using this feature, the NODDI-SH parameters fitting is not only fast, but also requires few sampling points (60 directions spread on two shells) in order to obtain good results. NODDI-SH efficiency and speed make it suitable to be used in clinical application. We also showed that 30 samples are enough to provide a good approximation of the average signal for each shell. Acquisition schemes which aim at detecting the volume fractions should, therefore, increase the number of \textit{b}-values, using only a moderate number of samples per shell.

%\subsection{Biological plausibility}
As mentioned in Section \ref{res:real}, there is no guarantee that the volume fractions estimated using compartmental models are linked to real changes in the tissue composition in each voxel. The fitting procedure \textit{always} provides a result, regardless of the correctness of the model itself. Low MSE, as the one presented in Figure \ref{fig:noddi_shadow_mse}, can be considered only as a general indication of the goodness of the model, and not the proof that the underlying tissue architecture corresponds to the actual model parameters. 
In this work, our goal was to compare the NODDI-SH with the classical NODDI model. NODDI exploits several assumptions for reducing the number of parameters to be fit that have been shown to be wrong for the human white matter. In particular the fact that the parallel diffusivity is set to $1.7\times10^{-3}$mm$^2$/s has been shown to be wrong in \cite{jelescu}, \cite{novikov_lemonade}, \cite{reisert}, and \cite{kaden}. Another over-simplification used in NODDI is the direct dependence of the perpendicular diffusivity to the intracellular volume fraction which is generally not true \cite{novikov_lemonade}. In our implementation, NODDI-SH shares the same formulation of NODDI model and, therefore, some of its limitations. 
 More accurate results would be obtained by making $\lambda_{\parallel}$ voxel-wise adaptive as in \cite{kaden2}, \cite{reisert}, \cite{jelescu}, and \cite{novikov_lemonade}. In this case, the only modification needed is the creation of new entries in the volume fractions dictionary with different $\lambda_{\parallel}$. For what concerns the microstructural parameters, \cite{reisert} Bayesian estimation of the volume fraction and water diffusivity seems to provide more reliable results with respect to NODDI.
As it is the case of NODDI-SH, \cite{reisert} exploit the mean value of the signal, as well as other rotation invariant features, in order to estimate such parameters. This approach could be easily integrated into NODDI-SH potentially improving the resulting fODF and MSE.

In this paper, we considered only the HCP acquisition scheme and
sub-sampling from it. Some model assumptions may not be valid in the case
of different acquisitions. For example, we used a ``stick'' to model
the intracellular compartment. Stick signal presents no decay in the
perpendicular plane. While this model can be realistic in the case of
small axon diameters at low \textit{b}-values, it could not be the
best choice for larger axons (e.g. $>$ 4 $\mu m$) at higher
\textit{b}-values (in the order of at least $10000$ s/mm$^2$). Another limitation is the fact that the HCP scheme uses a single diffusion time. In the case of multiple diffusion times, the Gaussian assumption for the extracellular compartment may not be true \cite{fieremans2016}, \cite{desantis2016}, and a different model should be used instead. Thanks to the flexibility of the NODDI-SH these diffusion time dependent model could be easily integrated into its framework. %As mentioned before, the integration of different signal models in NODDI-SH is straightforward, which jointly with the fact that it provides competitive performance with state-of-the-art methods in both estimating the volume fractions and extracting the fODF makes it one of the most promising models available so far.

\section{Conclusions}
\label{sec:con}
NODDI-SH is a flexible model that allows obtaining both tissues volume
fraction parameters and the fODF at the same time. Using NODDI-SH, the signal representation is analytically derived from the convolution of a NODDI inspired single fiber response with the fiber ODF modeled using SH. In this
work, we explicitly derived a single fiber response function
containing three compartments (isotropic, intra-cellular and
extra-cellular compartments, respectively). The result
of the convolution forms a basis in the diffusion signal space. The
basis first element can be used to estimate the volume fraction of the
three-compartments model.  
The current NODDI-SH model can be viewed as an extension of generic
distribution of fibers of previously proposed compartmental models,
such as NODDI, with an efficient parameters estimation procedure that
is efficient and robust to fanning and crossing fibers. The
processing pipeline of big datasets, such as the HCP, can be
drastically improved using NODDI-SH. Results show that NODDI-SH can also
be used for dataset presenting a limited number of samples, which
makes it feasible for processing clinical data. 

%
%%The key feature of the approach is the aniso-
%%tropic spring constant or scale parameter. The anisotropically-scaled
%%basis not only improves the ability of MAP-MRI to adapt to very dif-
%%ferent signal profiles, but can reduce the technique to the widely-
%employed DTI method if only the first of the basis functions is
%employed. Consequently, the MAP-MRI technique subsumes DTI
%while also providing several novel, quantifiable parameters that cap-
%ture previously obscured intrinsic features of nervous tissue micro-
%structure. The features of the employed basis make the MAP-MRI
%framework very robust and it may also be adapted to the technical
%limitations of in vivo imaging of clinical patients. Hence, MAP-MRI
%should prove helpful for investigating a spectrum of important scien-
%tific problems regarding the functional organization of normal and
%pathologic nervous tissue. This may ultimately lead to increased
%diagnostic accuracy of diffusion-weighted MRI for patients with CNS
%disease.
 
Future works will include the extension of the NODDI-SH to other models \cite{yolcu2015}, \cite{isbi:Evren} and the estimation of the fiber dispersion from the fODF.
%\clearpage
\section{Acknowledgements}
Data were provided by the Human Connectome Project, WU-Minn Consortium (Principal Investigators: David Van Essen and Kamil Ugurbil; 1U54MH091657) funded by the 16 NIH Institutes and Centers that support the NIH Blueprint for Neuroscience Research; and by the McDonnell Center for Systems Neuroscience at Washington University.
\appendix
\section{Real Symmetric Spherical Harmonics}
\label{ap_sh}
The real symmetric SH basis $Y_l^m$ is defined as
\begin{equation}
Y_l^m=
\begin{cases}
\sqrt{2}\cdot Re(\hat{Y}_l^m),
& \text{if -l $\leq$ $m$ $<$ 0} \\
\hat{Y}_l^0, & \text{if $m$=0}\\
\sqrt{2} \cdot Img(\hat{Y}_l^m), & \text{if 0 $<$ $m$ $\leq$ $l$}
\end{cases}
\end{equation}
where $\hat{Y}_l^m$ is the general SH basis, written as
\begin{equation}
\hat{Y}_l^m(\theta,\phi) = \sqrt{\dfrac{(2l+1)(l-m)!}{4 \pi (l+m)!}} P_l^m(\cos \theta) e^{i m \phi}
\end{equation}
with $\theta,\phi$ the polar representation of $\textbf{u}$, and $P_l^m$ the associated Legendre Polynomial.
A possible way to represent $\rho$ is to consider its SH series expansion 
\begin{equation}
\rho({\textbf{v}}) = \sum_{l=0, even}^{\infty} \sum_{m=-l}^{l} c_{lm} Y_l^m({\textbf{v}})
\end{equation}
\section{Analytical derivation of the FORECAST model}
\label{sec:appendix_a}
In the FORECAST model the diffusion signal can be calculated as 
\begin{equation}
	E(b,{\textbf{u}}) = \sum_{l=0, even}^{\infty} \sum_{m=-l}^{l} c_{lm} \int_{{\textbf{v}} \in \mathcal{S}^2}    \exp\bigl(-b {\textbf{u}}^T \textbf{D}({\textbf{v}}) {\textbf{u}}\bigr)  Y_l^m({\textbf{v}}) d {\textbf{v}}
\end{equation}
with $b = \vert\textbf{b} \vert = 4 \pi^2 \tau q^2$.
The tensor $ \textbf{D}({\textbf{v}})$ can calculated as 
\begin{equation}
 \textbf{D}({\textbf{v}}) = (\lambda_{\parallel}- \lambda_{\perp}) ({\textbf{v}} {\textbf{v}}^T) + \lambda_{\perp}\textbf{I}
\end{equation}
assuming that the two perpendicular eigenvalues are equal. We can now calculate $E(\textbf{b})$ as 
\begin{equation}
\begin{split}
	& E(b,{\textbf{u}}) = \sum_{l=0, even}^{\infty} \sum_{m=-l}^{l} c_{lm} \int_{{\textbf{v}} \in \mathcal{S}^2}    \exp\bigl(-b {\textbf{u}}^T \textbf{D}({\textbf{v}}) {\textbf{u}}\bigr)  Y_l^m({\textbf{v}}) d {\textbf{v}}\\
=& \sum_{l=0, even}^{\infty} \sum_{m=-l}^{l} c_{lm} \int_{{\textbf{v}} \in \mathcal{S}^2}  \exp\bigl(-b{\textbf{u}}^T \bigl[(\lambda_{\parallel}- \lambda_{\perp}) ({\textbf{v}} {\textbf{v}}^T) + \lambda_{\perp}\textbf{I} \bigr] {\textbf{u}}\bigr)  Y_l^m({\textbf{v}}) d {\textbf{v}}\\
=& \sum_{l=0, even}^{\infty} \sum_{m=-l}^{l} c_{lm} \int_{{\textbf{v}} \in \mathcal{S}^2}  \exp\bigl(-b \bigl[ (\lambda_{\parallel}- \lambda_{\perp}){\textbf{u}}^T  ({\textbf{v}} {\textbf{v}}^T){\textbf{u}} + \lambda_{\perp} \bigr] \bigr)  Y_l^m({\textbf{v}}) d {\textbf{v}}\\
=& \sum_{l=0, even}^{\infty} \sum_{m=-l}^{l} c_{lm} \int_{{\textbf{v}} \in \mathcal{S}^2}  \exp\bigl(-b \bigl[ (\lambda_{\parallel}- \lambda_{\perp})({\textbf{u}}^T {\textbf{v}})^2 + \lambda_{\perp} \bigr] \bigr)  Y_l^m({\textbf{v}}) d {\textbf{v}}\\	
\end{split}
\end{equation}
Using the \textit{Funk-Hecke Theorem} \cite{funk}, \cite{Descoteaux2007} stating that, given a unit vector ${\textbf{u}}$, an integral on the sphere can be converted to a one dimensional integral 
\begin{equation}
\int_{\vert {\textbf{v}} \vert = 1} f({\textbf{u}}^T{\textbf{v}}) H_l({\textbf{v}}) d{\textbf{v}} = 2 \pi \int_{-1}^{1} P_l(t) f(t) dt H_l({\textbf{u}})
\end{equation}
assuming $f(t)$ continuous on $[-1,1]$ and $H_l$ any SH of order $l$, and with $P_l$ the Legendre polynomial of degree $l$.

Replacing it into FORECAST equation the problem became:
\begin{equation}
\begin{split}
	& E(b,{\textbf{u}}) = \sum_{l=0, even}^{\infty} \sum_{m=-l}^{l} c_{lm} 2 \pi  \int_{-1}^{1} P_l(t) \exp\bigl(-b \bigl[ (\lambda_{\parallel}- \lambda_{\perp})t^2 + \lambda_{\perp} \bigr] \bigr)  d t Y_l^m({\textbf{u}}) \\
=& \sum_{l=0, even}^{\infty} \sum_{m=-l}^{l} c_{lm} 2 \pi \exp\bigl(-b \lambda_{\perp} \bigr)  Y_l^m({\textbf{u}})  \int_{-1}^{1} P_l(t) \exp\bigl(-b (\lambda_{\parallel}- \lambda_{\perp})t^2  \bigr)  d t\\	
\end{split}
\end{equation}
The only step left is to solve $ \int_{-1}^{1} P_l(t) \exp\bigl(-b(\lambda_{\parallel}- \lambda_{\perp}) t^2  \bigr)  d t$. This integral does not have a general solution, but we can solve it for each single Legendre polynomial of a certain order. The first eight even Legendre polynomials are
\begin{description}
\item [$P_0(x)$] = $1$
\item [$P_2(x)$] = $\frac{1}{2}(3x^2 -1 )$
\item [$P_4(x)$] = $\frac{1}{8}(35x^4 -30x^2 +3 )$
\item [$P_6(x)$] = $\frac{1}{16}(231x^6 -315x^4 +105x^2 -5 )$
\item [$P_8(x)$] = $\frac{1}{128}(6435x^8 - 12012x^6 + 6930x^4 -1260x^2 +35 )$
%\item [$P_{10}(x)$] $\frac{1}{256}(46189x^{10} -109395x^8 + 90090x^6 -30030x^4 +3465x^2 -63 )$
%\item [$P_{12}(x)$] $\frac{1}{1024}(676039x^{12} - 1939938x^{10} +  2078505x^8 - 1021020x^6 + 225225x^4 -18018x^2 +231 )$
\end{description}
Considering $\xi= b(\lambda_{\parallel}- \lambda_{\perp})$, we can now calculate each integral $\Phi_l=\int_{-1}^{1} t^l \exp\bigl(-\xi t^2  \bigr)  d t$ necessary to integrate each Legendre polynomials
\begin{equation}
\begin{split}
& \Phi_0 = \int_{-1}^{1} \exp\bigl(-\xi t^2  \bigr) d t = \frac{\sqrt{\pi}\erf(\sqrt{\xi})}{\xi^{1/2}} \\
& \Phi_2 =  \int_{-1}^{1} t^2 \exp\bigl(-\xi t^2  \bigr) d t = \dfrac{\sqrt{\pi}\erf(\sqrt{\xi}) -2e^{-\xi}\sqrt{\xi} }{2 \xi^{3/2}} \\
& \Phi_4 =  \int_{-1}^{1} t^4 \exp\bigl(-\xi t^2  \bigr) \xi t = \dfrac{3\sqrt{\pi}\erf(\sqrt{\xi}) -2e^{-\xi}\sqrt{\xi}(2\xi+3) }{4 \xi^{5/2}}\\
& \Phi_6 =  \int_{-1}^{1} t^6 \exp\bigl(-\xi t^2  \bigr) d t = \dfrac{15\sqrt{\pi}\erf(\sqrt{\xi}) -2e^{-\xi}\sqrt{\xi}(4\xi^2+10\xi+15) }{8 \xi^{7/2}} \\
& \Phi_8 =  \int_{-1}^{1} t^8 \exp\bigl(-\xi t^2  \bigr) d t = \dfrac{105\sqrt{\pi}\erf(\sqrt{\xi}) -2e^{-\xi}\sqrt{\xi}(8\xi^3+28\xi^2+70\xi+105) }{16 \xi^{9/2}}  \\
%& \int_{-1}^{1} x^{10} \exp\bigl(-\xi t^2  \bigr) d t = \dfrac{945\sqrt{\pi}\erf(\sqrt{\xi}) -2e^{-\xi}\sqrt{\xi}(16\xi^4+ 72\xi^3+252\xi^2+630\xi+945) }{32 \xi^{11/2}}  \\
%& \int_{-1}^{1} x^{12} \exp\bigl(-\xi t^2  \bigr) \xi t = \dfrac{10395\sqrt{\pi}\erf(\sqrt{\xi}) -2e^{-\xi}\sqrt{\xi}(32\xi^5 + 176\xi^4+ 792\xi^3+2772\xi^2+6930\xi+10395) }{64 \xi^{13/2}}  
\end{split}
\end{equation}
When $\xi \to 0$ $\Phi_n(\xi)$ is indeterminate but the following limits hold true  
\begin{equation}
\begin{split}
\lim_{\xi \to 0}\Phi_0(\xi) =& 2 \\
\lim_{\xi \to 0}\Phi_2(\xi) =& \frac{2}{3} \\ 
\lim_{\xi \to 0}\Phi_4(\xi) =& \frac{2}{5} \\ 
\lim_{\xi \to 0}\Phi_6(\xi) =& \frac{2}{7} \\ 
\lim_{\xi \to 0}\Phi_8(\xi) =& \frac{2}{9} \\ 
%\lim_{b \to 0}I_{10}(b) =& \frac{2}{11} \\ 
%\lim_{b \to 0}I_{12}(b) =& \frac{2}{13} \\ 
\end{split}
\end{equation}
We can define $\Psi_l(\xi)$ as the value of the Legendre polynomials $P_l(x)$ when $x=\Phi_l(\xi)$, which for the first 8 Legendre polynomials corresponds to
 \begin{description}
\item [$\Psi_0(\xi)$] = $\Phi_0(\xi)$
\item [$\Psi_2(\xi)$] = $\frac{1}{2}\bigl(3\Phi_2(\xi)- \Phi_0(\xi)\bigr)$
\item [$\Psi_4(\xi)$] = $ \frac{1}{8}\bigl(35\Phi_4(\xi) -30\Phi_2(\xi) +3\Phi_0(\xi) \bigr)$
\item [$\Psi_6(\xi)$] = $\frac{1}{16}\bigl(231\Phi_6(\xi) -315 \Phi_4(\xi) +105\Phi_2(\xi) -5\Phi_0(\xi) \bigr)$
\item [$\Psi_8(\xi)$] = $ \frac{1}{128}\bigl(6435\Phi_8(\xi) - 12012\Phi_6(\xi) + 6930\Phi_4(\xi) -1260\Phi_2(\xi) +35\Phi_0(\xi) \bigr)$
%\item [$l=10$] $\Rightarrow$ $P_{I,10}(x) = \frac{1}{256}(46189I_{10}(x) -109395I_{8}(x) + 90090I_{6}(x) -30030I_{4}(x) +3465I_{2}(x) -63I_{0}(x) )$
%\item [$l=12$] $\Rightarrow$ $P_{I,12}(x) = \frac{1}{1024}(676039I_{12}(x) - 1939938I_{10}(x) +  2078505I_{8}(x) - 1021020I_{6}(x) + 225225I_{4}(x) -18018I_{2}(x) +231I_{0}(x) )$
\end{description}
The FORECAST signal equation can then be written as 
\begin{equation}
\begin{split}
	E(b,{\textbf{u}}) =& \sum_{l=0, even}^{N} \sum_{m=-l}^{l} c_{lm} 2 \pi \exp\bigl(-b \lambda_{\perp} \bigr) \Psi_{l}(b (\lambda_{\parallel}- \lambda_{\perp}))  Y_l^m({\textbf{u}})  \\	
\end{split}
\end{equation}
where N $\in$ [0,8].

\subsubsection{Special case b=0}
\label{sec:special}
In the $b=0$ case, the signal equation can be rewritten as 
\begin{equation}
\begin{split}
	E(0,{\textbf{u}}) =& \sum_{l=0, even}^{N} \sum_{m=-l}^{l} c_{lm} 2 \pi Y_l^m({\textbf{u}}) \Psi_{l}(0)  \\
	=& c_{00} 2 \pi  \frac{1}{2}\sqrt{\frac{1}{\pi}} 2  \\
	=& c_{00} \sqrt{4\pi} \\	
\end{split}
\end{equation}
This because $\Psi_{l}(0) = 2 $ for $l=0$ and zero otherwise, and $Y_0^0({\textbf{u}})= \frac{1}{2}\sqrt{\frac{1}{\pi}} \ \forall \ {\textbf{u}}$.
In order to have a normalized signal with $E(0)=1$, $c_{00}$ must be equal to $\frac{1}{\sqrt{4\pi}}$.
This is also the same condition which we need to verify equation \eqref{eq:property}:
\begin{equation}
\int_{{\textbf{v}}}  \sum_{l=0, even}^{N} \sum_{m=-l}^{l} c_{lm} Y_l^m({\textbf{v}}) \ d {\textbf{v}}  = 1
\end{equation}
this because
\begin{equation}
\begin{split}
& \int_{{\textbf{v}}}  \sum_{l=0, even}^{N} \sum_{m=-l}^{l} c_{lm} Y_l^m({\textbf{v}}) \ d {\textbf{v}}  = \\
=& \sum_{l=0, even}^{N} \sum_{m=-l}^{l} c_{lm} \int_{{\textbf{v}}}  Y_l^m({\textbf{v}}) \ d {\textbf{v}} \\
=& \sum_{l=0, even}^{N} \sum_{m=-l}^{l} c_{lm} \sqrt{4 \pi} \delta_{lm}^{00}\\
=&  c_{00} \sqrt{4 \pi} 
\end{split}
\end{equation}
where $\delta_{lm}^{00}$ is equal to one when $l=0$ and $m=0$ and zero elsewhere.
This equation is valid only for $c_{00} = \frac{1}{\sqrt{4\pi}}$.
\section{Analytical derivation of the NODDI-SH}
\label{sec:appendix_b}
The three-compartments NODDI-SH model can be derived equation \eqref{eq:shadow_single} into equation \eqref{eq:general}:
\begin{equation}
\begin{split}
	& E(b,{\textbf{u}}) = \sum_{l=0, even}^{\infty} \sum_{m=-l}^{l} c_{lm} \int_{{\textbf{v}} \in \mathcal{S}^2}    S(b, {\textbf{u}} | {\textbf{v}})  Y_l^m({\textbf{v}}) d {\textbf{v}}\\
=& 	\sum_{l=0, even}^{\infty} \sum_{m=-l}^{l} c_{lm} \int_{{\textbf{v}} \in \mathcal{S}^2}   \nu_{ic}S_{ic}(b, {\textbf{u}} | {\textbf{v}}) + \nu_{ec}S_{ec}(b, {\textbf{u}} | {\textbf{v}})+ \nu_{csf}S_{csf}(b)  Y_l^m({\textbf{v}}) d {\textbf{v}}\\
=& 	\sum_{l=0, even}^{\infty} \sum_{m=-l}^{l} c_{lm} \Biggl[ \int_{{\textbf{v}} \in \mathcal{S}^2}   \nu_{ic}S_{ic}(b, {\textbf{u}} | {\textbf{v}})Y_l^m({\textbf{v}}) d {\textbf{v}} +\\ 
& + \int_{{\textbf{v}} \in \mathcal{S}^2} \nu_{ec}S_{ec}(b, {\textbf{u}} | {\textbf{v}})Y_l^m({\textbf{v}}) d {\textbf{v}}+ \nu_{csf}S_{csf}(b)  \int_{{\textbf{v}} \in \mathcal{S}^2} Y_l^m({\textbf{v}}) d {\textbf{v}} \Biggr]\\
=&  \sum_{l=0, even}^{N} \sum_{m=-l}^{l} c_{lm} \Biggl[\nu_{ic} 2 \pi  Y_l^m({\textbf{u}})  \Psi_{l}(b \lambda_{\parallel})  +  \nu_{ec} 2 \pi \exp\bigl(-b \lambda_{\perp} \bigr)  Y_l^m({\textbf{u}}) \\ 
& \Psi_{l}(b (\lambda_{\parallel}- \lambda_{\perp})) + \nu_{csf}\exp\bigl(-b\ \lambda_{csf} \bigr) \sqrt{4 \pi} \delta_{lm}^{00} \Biggr] \\	
=&  c_{00}\sqrt{4 \pi} \nu_{csf}\exp\bigl(-b\ \lambda_{csf} \bigr)  +\sum_{l=0, even}^{N} \sum_{m=-l}^{l} c_{lm} 2 \pi \Bigl[\nu_{ic}  \Psi_{l}(b \lambda_{\parallel})  + \\
& + \nu_{ec} \exp\bigl(-b \lambda_{\perp} \bigr) \Psi_{l}(b (\lambda_{\parallel}- \lambda_{\perp})) \Bigr] Y_l^m({\textbf{u}}) \\ 	
\end{split}
\end{equation}
solving each integral as a single FORECAST basis (see Appendix \ref{sec:appendix_a}).
The final formulation of the NODDI-SH model can be rearranged as 
\begin{equation}
\begin{split}
	& E(q,{\textbf{u}}) =   c_{00}\sqrt{4 \pi} \nu_{csf}\exp\bigl(-4\pi^2 \tau q^2\ \lambda_{csf} \bigr)  +\sum_{l=0, even}^{N} \sum_{m=-l}^{l} c_{lm} 2 \pi \\& \Bigl[\nu_{ic}  \Psi_{l}(4\pi^2 \tau q^2 \lambda_{\parallel}) + \nu_{ec} \exp\bigl(-4\pi^2 \tau q^2 \lambda_{\perp} \bigr) \Psi_{l}(4\pi^2 \tau q^2 (\lambda_{\parallel}- \lambda_{\perp})) \Bigr] Y_l^m({\textbf{u}}) \\ 	
\end{split}
\end{equation}
\clearpage

\bibliographystyle{splncs} 
\bibliography{mybibfile}

\end{document}